\documentclass{article}
\PassOptionsToPackage{numbers}{natbib}
\usepackage[final,main]{neurips_2026}
\usepackage{graphicx}
\usepackage{adjustbox}
\usepackage{amsmath,amssymb,amsfonts}
\usepackage{url}
\usepackage{booktabs}
\usepackage{tabularx}
\usepackage{array}
\usepackage{algorithm}
\usepackage{algpseudocode}
\usepackage{newtxtext}
\usepackage{listings}
\usepackage{xcolor}
\usepackage{adjustbox}
\usepackage[table]{xcolor} 
\usepackage{xspace}
\usepackage{enumitem}
\usepackage{multirow}
\usepackage{float}
\usepackage{capt-of}
\usepackage{wrapfig}
\usepackage{tikz}
\usetikzlibrary{positioning,fit,arrows.meta,matrix,calc}
\usepackage[colorlinks=true,linkcolor=blue,citecolor=blue,urlcolor=blue]{hyperref}


\lstdefinestyle{yamlstyle}{
    basicstyle=\ttfamily\footnotesize,
    frame=single,
    breaklines=true,
    showstringspaces=false
}

\newcommand{\picid}{\textsc{Picid}\xspace}

\newcommand{\vx}{\mathbf{x}}

\newcommand{\vz}{\mathbf{z}}
\newcommand{\vW}{\mathbf{W}}
\newcommand{\vX}{\mathbf{X}}

\newcommand{\gX}{\mathcal{X}}
\newcommand{\gY}{\mathcal{Y}}
\newcommand{\gZ}{\mathcal{Z}}

\title{\picid: A Modular Evaluation Infrastructure for Reproducible PHM Across Tasks and Domains}

\author{
Lev Telyatnikov\thanks{Equal contribution.}\\
\textit{EPFL}
\And
Raffael Theiler\footnotemark[1]\\
\textit{EPFL}
\And
Leandro Von Krannichfeldt\\
\textit{EPFL}
\And
Olga Fink\\
\textit{EPFL}
}
\begin{document}
\maketitle
\begin{abstract}
  Progress in Prognostics and Health Management (PHM) is hindered by the lack of standardized and reusable evaluation practices across tasks, datasets, and application domains. Reported results are often difficult to reproduce and compare, as key protocol choices, such as data splits, preprocessing, label alignment, temporal windowing, and metrics, are often implicit or implemented ad hoc. We introduce \picid, a modular evaluation infrastructure that formalizes the PHM evaluation pipeline as an explicit, executable, and reproducible protocol. Through well-defined abstractions, \picid enforces deterministic, leakage-safe dataset construction while remaining flexible across diverse PHM settings. The framework supports fault detection, diagnostics, and prognostics through a unified interface and can be extended to new datasets and model classes without violating protocol invariants. By standardizing data contracts and evaluation boundaries, \picid  also enables fair cross-task comparisons across diagnostics (classification) and prognostics (regression), allowing identical model families to be evaluated consistently across heterogeneous settings. We demonstrate \picid through an empirical evaluation of thirteen models on twelve datasets spanning batteries, bearings, turbofan engines, hydraulics, filtration systems, and buildings. This work establishes a reusable foundation for standardized, fair and reproducible evaluation in PHM.

\end{abstract}

\section{Introduction}
\label{sec:intro}

Data-driven Prognostics and Health Management (PHM)---the application of machine learning (ML) to monitor the health of engineering assets and anticipate failures---has matured into an established research area (Appendix~\ref{app:intro_phm} provides a full introduction to the field). PHM improves the reliability, safety, and cost-efficiency in industrial and infrastructure systems through condition-based maintenance, and lifetime extension. Recent advances in deep learning have significantly advanced its core tasks: fault detection, diagnostics, and prognostics.

As the PHM field matures and model performance on widely used datasets saturates, evaluation methodology  rather than model capacity has increasingly become a primary bottleneck to further progress. Reported improvements are often  highly sensitive to protocol design choices and implementation details, yet the field still  lacks a standardized and reusable evaluation infrastructure spanning tasks, datasets, and application domains. Existing resources address isolated components of the workflow---such as public datasets, access libraries, task-specific benchmarks, or deployment-oriented prognostic tools (Section~\ref{sec:related_work})---but do not provide a unified protocol layer that makes experimental assumptions explicit, reproducible, and consistent across heterogeneous PHM settings. This gap is reflected in publication practice: prior work reports low code-and-data release rates \citep{vonhahn2023reproducibility}, and our 2022--2025 audit found publicly available code in only 8 of 329 papers (Appendix~\ref{app:phm_audit}). As a result, extending or comparing methods often requires reimplementing implicit protocol choices, which can unintentionally change the underlying prediction problem and bias empirical comparisons. Consequently, it is often unclear whether reported gains arise from improved modeling or from uncontrolled variation in preprocessing, partitioning, target construction, and evaluation. Small protocol changes can substantially alter reported performance and even model rankings. This work therefore aims not only to improve reproducibility, but also to make such protocol effects explicit and systematically analyzable, enabling reliable and falsifiable comparison across PHM methods.

Standardizing evaluation in PHM is difficult because the field does not consist of a single task, data modality, or prediction setting. Fault detection, diagnostics, and prognostics are structurally related but instantiated very differently across applications: a bearing diagnostics study may classify short vibration segments into discrete fault categories, whereas a battery prognostics study may track continuous capacity fade over hundreds of charge-discharge cycles. These settings differ in sensor modalities, label structures, temporal scales, degradation semantics, and evaluation conventions. A useful PHM evaluation infrastructure must therefore accommodate heterogeneous settings while preserving comparable execution rules. This challenge is further complicated by protocol choices, such as temporal alignment, windowing, partitioning, and target construction, that can materially affect the resulting prediction problem.

The object that must be standardized, therefore, is the execution protocol. In PHM, the protocol is not merely an implementation detail: it partially defines the prediction problem. Choices such as preprocessing, temporal windowing, target construction and alignment determine what information is available to the model, how labels are constructed, and what constitutes a valid prediction. For example, in rolling-bearing diagnostics, randomly splitting overlapping segments (or repeated measurements of the same physical bearing) across train and test can inflate reported accuracy even when the model does not generalize to new bearings \citep{hendriks2022towards,matania2024test}. When these choices remain implicit, nominally similar experiments may instantiate different prediction problems even on the same dataset, making model rankings unstable under controlled protocol changes.

Addressing these limitations requires a shift from \textbf{model-centric reporting} to \textbf{protocol-centric evaluation}. In this view, the protocol components governing comparability---including task definitions, data-flow boundaries, and evaluation invariants---are made explicit and enforced through executable software abstractions, while models, datasets, and task instantiations remain extensible research components. The goal is not only to improve reproducibility, but to make protocol effects explicit, measurable, and systematically analyzable, enabling reliable and verifiable comparison within PHM. By fixing protocol invariants while allowing extensible models and datasets, the framework enables controlled experimentation on modeling choices independently of pipeline variation.

To address these challenges, we introduce \picid{}\footnote{Code repository: \url{https://github.com/picid-research/picid}}, a modular evaluation infrastructure for PHM. \picid couples a formal multi-task specification to software that executes the protocol directly: it constructs splits, fits preprocessing on training data, aligns targets to the representation, windows trajectories into supervised samples, and computes task metrics. Diagnostics and prognostics share these protocol components, while datasets, models, and task instantiations remain extensible. By fixing the evaluation contract---the rules governing how experimental components interact---\picid reduces hidden pipeline variation and enables more reliable attribution of performance differences to modeling choices. The shared protocol and model interface further enable controlled cross-task comparison, allowing identical model families to be evaluated consistently across diagnostics and prognostics.

This design unifies diagnostics and prognostics within a single evaluation framework rather than disconnected task-specific pipelines. Our empirical study focuses on these tasks, where standardized PHM evaluation infrastructure is most lacking; fault detection is supported by the framework but omitted from the main evaluation, due to the broader availability of general time-series anomaly detection benchmarks and tooling outside PHM \citep{Wang2024DeepTS}. Built as reusable research infrastructure, \picid{} emphasizes deterministic execution, leakage-safe evaluation, modular extension, and strong automated testing. Using \picid, we further show that controlled changes to preprocessing, partitioning, and target construction can substantially alter evaluation outcomes.

We formalize our contributions as follows:

\begin{enumerate}
    \item We formalize PHM evaluation as an explicit protocol spanning preprocessing, partitioning, target construction and alignment, and metric computation across diagnostics, and prognostics, thereby clarifying the conditions under which comparisons across models, datasets, and tasks are scientifically valid (Section~\ref{sec:framework_formalization}, Appendix~\ref{app:complete_formalization}).

    \item We introduce \picid, a modular evaluation infrastructure that realizes this formalization through executable abstractions. The framework deterministic, leakage-safe evaluation and protocol invariants while remaining extensible to new datasets, models, and task instantiations (Section~\ref{sec:picid_library}, Appendix~\ref{app:library_architecture}).

    \item We instantiate \picid on a large-scale evaluation study spanning thirteen models and twelve datasets across batteries, bearings, turbofan engines, hydraulics, filtration, and buildings, and use it to analyze the sensitivity of evaluation outcomes to protocol design choices (Section~\ref{sec:empirical_validation}, Appendix~\ref{app:full_experiments}).
\end{enumerate}

\section{Related work}
\label{sec:related_work}

\paragraph{Evaluation infrastructure in ML}
Several ML domains have developed shared infrastructures for standardized, reproducible comparison (e.g., NLP, graph learning, reinforcement learning) \citep{lhoest-etal-2021-datasets,fey2019pyg,stable-baselines3}. The closest analogue outside PHM is the time-series machine learning ecosystem, with toolkits such as sktime and aeon \citep{sktimets,aeon24jmlr,gluonts,darts,olivares2022library_neuralforecast,agtimeseries}, which expose classification, forecasting, and anomaly-detection tasks through model-agnostic APIs, standardized splits, and reproducible evaluation pipelines. However, PHM is a setting where the protocol is not merely a measurement procedure: choices such as temporal alignment, target construction, preprocessing, and partitioning can change what information is available and therefore what problem is being solved. This is especially visible when feature extraction changes the temporal grid or when evaluation must respect multi-unit semantics. Consequently, general-purpose time-series benchmarking APIs do not, by themselves, enforce PHM-specific invariants (grid-aware alignment, unit-level evaluation), and ``standardized'' pipelines can still be semantically inconsistent across studies (Appendix~\ref{app:extended_related_work}).

\paragraph{PHM data access and domain software}
Within PHM, one important cluster of work focuses on data access, metadata management, and domain software rather than broad experimental execution. PyPHM and PHMD reduce friction in discovering, downloading, and preprocessing PHM datasets \citep{von2022computational,SOLISMARTIN2025102039}, while a collaborative prognostics data library emphasizes secure sharing and metadata stewardship for reusable industrial datasets \citep{sikorska2016collaborative}. ProgPy provides a richer software stack for prognostics modeling, simulation, and deployment-oriented workflows \citep{teubert2023progpy}. \citet{bieber_generic_2022} propose a generic prognostics framework integrating data preprocessing, feature extraction, and prognostic modeling with a focus on a genetic algorithm-based optimization and without a reusable implementation.
GSAP provides a modular software architecture with standardized interfaces for prognostic models and communication components interfacing with external systems in prognostics applications \citep{teubert_generic_2020}.
These efforts facilitate data access and prognostic applications, but they do not enforce a shared evaluation protocol across heterogeneous PHM tasks and datasets (Appendix~\ref{app:extended_related_work:utilities}). As a result, experimental assumptions, such as preprocessing pipelines, temporal windowing, or evaluation boundaries, remain implicit and study-specific, limiting comparability and reproducibility.

\paragraph{PHM evaluation platforms and comparative studies}
A second cluster focuses on comparative evaluation in narrower PHM settings. PHM Society challenges and their retrospective analyses helped establish shared public PHM tasks early in the field's development \citep{jia2018review}. More recently, PHM-Vibench has emerged as the strongest direct platform-level comparison point, offering a modular environment for vibration-centered PHM tasks including diagnosis, fault detection, and RUL prediction \citep{Li2026}. These efforts are important steps toward benchmarking, but their scope remains limited: they are typically tied to vibration data and do not formalize how preprocessing, target alignment, and evaluation boundaries compose into a reusable protocol layer across domains; they are also oriented toward foundation-model adaptation, whereas \picid is cross-domain and model-class agnostic (Appendix~\ref{app:extended_related_work:platforms}). More broadly, other PHM works provide strong task- or scenario-specific comparative studies, but their experimental logic typically remains embedded in a particular modality, dataset family, or evaluation setting rather than exposed as reusable infrastructure.

\paragraph{Methodology and reproducibility}
A third line of work calls for stronger methodological standardization in PHM. Early methodological work argues that PHM still lacks a sufficiently standardized research methodology \citep{uckun2008standardizing}, while modeling-ecosystem work stresses that reproducibility depends on documenting data, validation choices, uncertainty treatment, and application limits, not only the final algorithm \citep{astfalck2016modelling}. A recent PHM review of open challenge datasets likewise notes the absence of unified guidelines and systematic approaches for machine-learning-driven PHM \citep{Su2024}. These contributions sharpen the diagnosis, but stop short of providing executable infrastructure that enforces common PHM protocol rules in reusable software (Appendix~\ref{app:extended_related_work:methodology}).

\paragraph{Positioning}

Across these lines of work, the missing piece is a reusable, protocol-centric evaluation layer for PHM: existing approaches either (i) standardize data access, (ii) provide task-specific benchmarks, or (iii) articulate methodological needs, but they do not make the full execution protocol explicit, enforceable, and reusable across tasks and domains. \picid{} fills this gap by treating the evaluation contract as an executable artifact, enabling protocol-consistent comparison across fault detection, diagnostics, and prognostics and across heterogeneous domains (batteries, bearings, turbofan engines, hydraulics, filtration, buildings). Appendix~\ref{app:extended_related_work:platforms} provides detailed comparisons. The goal is not to replace specialized PHM benchmarks or software stacks, but to provide a unifying layer in which the formal evaluation protocol and the software that executes it are the same artifact.

\section{Framework formalization}
\label{sec:framework_formalization}

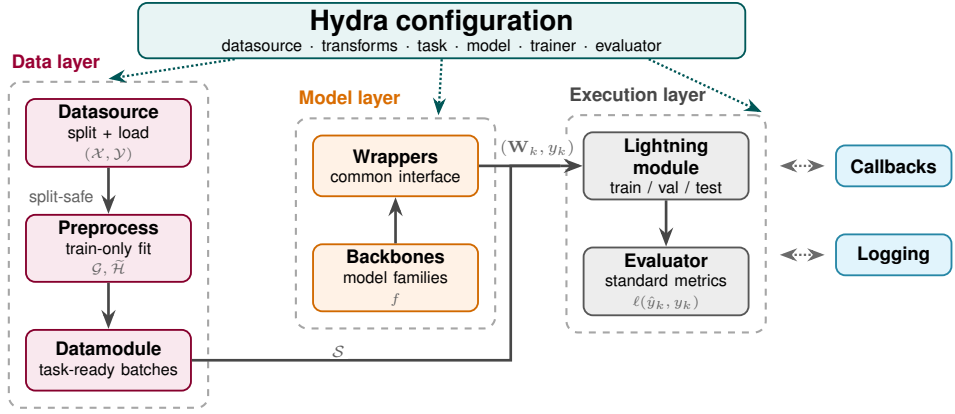
\begin{figure}[t]
\centering
\resizebox{0.9\linewidth}{!}{%
\begin{tikzpicture}[
  font=\sffamily,
  >=Stealth,
  familybox/.style={
    rounded corners=4pt,
    thick,
    minimum width=2.50cm,
    minimum height=0.92cm,
    text width=2.30cm,
    align=center,
    inner sep=2.5pt,
    font=\sffamily\footnotesize
  },
  databox/.style={familybox, draw=purple!70!black, fill=purple!9},
  modelbox/.style={familybox, draw=orange!82!black, fill=orange!10},
  execbox/.style={familybox, draw=gray!68!black, fill=gray!14},
  supportbox/.style={
    rounded corners=4pt,
    thick,
    minimum width=1.75cm,
    minimum height=0.62cm,
    text width=1.60cm,
    align=center,
    inner sep=2.0pt,
    font=\sffamily\footnotesize,
    draw=cyan!60!black,
    fill=cyan!10
  },
  cfgbox/.style={
    rounded corners=4pt,
    draw=teal!70!black,
    fill=teal!9,
    thick,
    minimum width=9.2cm,
    minimum height=0.72cm,
    text width=8.8cm,
    align=center,
    inner sep=3pt,
    font=\sffamily\small
  },
  group/.style={
    rounded corners=5pt,
    draw=black!35,
    dashed,
    line width=0.9pt,
    inner sep=7pt
  },
  flow/.style={->, line width=1.15pt, draw=black!72},
  supportline/.style={line width=1.0pt, densely dotted, draw=black!58},
  supporthead/.style={->, line width=1.0pt, draw=black!58, shorten >=0.5pt, shorten <=0.5pt},
  cfgflow/.style={->, line width=1.0pt, densely dotted, draw=teal!68!black},
  grouplabel/.style={font=\sffamily\bfseries\footnotesize, fill=white, inner sep=1pt},
  groupsub/.style={font=\sffamily\scriptsize, text=black!62},
  mathlabel/.style={font=\scriptsize, text=black!62, fill=white, inner sep=1pt}
]

\node[cfgbox] (config) at (5.05, 5.1)
  {\textbf{\large Hydra configuration}\\[-1pt]
   {\scriptsize datasource $\cdot$ transforms $\cdot$ task $\cdot$ model $\cdot$ trainer $\cdot$ evaluator}};

\node[databox] (source) at (0.0, 3.55)
  {\textbf{Datasource}\\[-1pt]
   {\scriptsize split + load}\\[-1pt]
   {\tiny\textcolor{black!62}{$(\gX,\gY)$}}};

\node[databox, below=7mm of source] (prep)
  {\textbf{Preprocess}\\[-1pt]
   {\scriptsize train-only fit}\\[-1pt]
   {\tiny\textcolor{black!62}{$\mathcal G,\widetilde{\mathcal H}$}}};

\node[databox, below=7mm of prep] (batches)
  {\textbf{Datamodule}\\[-1pt]
   {\scriptsize task-ready batches}};

\node[group, fit=(source) (prep) (batches)] (datagrp) {};
\node[grouplabel, anchor=south west, text=purple!80!black] at ($(datagrp.north west)+(0.02,0.10)$)
  {Data layer};
\node[groupsub, anchor=south west, fill=white, inner sep=1pt] at ($(prep.north west)+(0.02,0.08)$)
  {split-safe};

\node[modelbox] (wrappers) at (4.35, 3.05)
  {\textbf{Wrappers}\\[-1pt]
   {\scriptsize common interface}};

\node[modelbox, below=7mm of wrappers] (backbones)
  {\textbf{Backbones}\\[-1pt]
   {\scriptsize model families}\\[-1pt]
   {\tiny\textcolor{black!62}{$f$}}};

\node[group, fit=(wrappers) (backbones)] (modelgrp) {};
\node[grouplabel, anchor=south west, text=orange!85!black] at ($(modelgrp.north west)+(0.02,0.10)$)
  {Model layer};

\node[execbox] (pipeline) at (8.45, 3.05)
  {\textbf{Lightning module}\\[-1pt]
   {\scriptsize train / val / test}};

\node[execbox, below=7mm of pipeline] (eval)
  {\textbf{Evaluator}\\[-1pt]
   {\scriptsize standard metrics}\\[-1pt]
   {\tiny\textcolor{black!62}{$\ell(\hat y_k,y_k)$}}};

\node[group, fit=(pipeline) (eval)] (execgrp) {};
\node[grouplabel, anchor=south west, text=black!72] at ($(execgrp.north west)+(0.02,0.10)$)
  {Execution layer};

\node[supportbox] (callbacks) at (11.90, 3.05) {\textbf{Callbacks}};
\node[supportbox, below=7mm of callbacks] (logging) {\textbf{Logging}};

\draw[flow] (source) -- (prep);
\draw[flow] (prep) -- (batches);
\draw[flow] (backbones) -- (wrappers);
\coordinate (pipelinejoin) at ([xshift=-0.34cm]pipeline.west);
\draw[line width=1.15pt, draw=black!72] (wrappers.east) -- (pipelinejoin);
\coordinate (datapathturn) at ([xshift=0.22cm]modelgrp.east |- batches.east);
\draw[line width=1.15pt, draw=black!72] (batches.east) -- node[above, mathlabel, pos=0.46] {$\mathcal S$} (datapathturn) |- (pipelinejoin);
\node[mathlabel, anchor=south east] at ([xshift=-0.08cm,yshift=0.10cm]pipeline.west) {$({\vW_k},y_k)$};
\draw[flow] (pipelinejoin) -- (pipeline.west);
\draw[flow] (pipeline) -- (eval);

\coordinate (cbexecport) at ([xshift=0.18cm]execgrp.east |- callbacks.west);
\coordinate (cbboxport) at ([xshift=-0.14cm]callbacks.west);
\coordinate (logexecport) at ([xshift=0.18cm]execgrp.east |- logging.west);
\coordinate (logboxport) at ([xshift=-0.14cm]logging.west);
\draw[supportline] ([xshift=0.17cm]cbexecport) -- ([xshift=-0.17cm]cbboxport);
\draw[supporthead] ([xshift=0.17cm]cbexecport) -- (cbexecport);
\draw[supporthead] ([xshift=-0.17cm]cbboxport) -- (cbboxport);
\draw[supportline] ([xshift=0.17cm]logexecport) -- ([xshift=-0.17cm]logboxport);
\draw[supporthead] ([xshift=0.17cm]logexecport) -- (logexecport);
\draw[supporthead] ([xshift=-0.17cm]logboxport) -- (logboxport);

\draw[cfgflow] ($(config.south west)!0.16!(config.south east)$) -- ([yshift=0.03cm]datagrp.north);
\draw[cfgflow] ($(config.south west)!0.50!(config.south east)$) -- ([xshift=0.65cm,yshift=0.03cm]modelgrp.north);
\draw[cfgflow] ($(config.south west)!0.84!(config.south east)$) -- ([xshift=1.45cm,yshift=0.03cm]execgrp.north);

\end{tikzpicture}
}
\caption{Architecture of the \picid{} evaluation stack. The data layer loads raw unit trajectories, applies split-safe preprocessing, and constructs task-ready batches through the transformation pipeline $(\mathcal{G}, \widetilde{\mathcal{H}})$ and the windowing operator $\mathcal{S}$. The model layer maps heterogeneous backbones into a common wrapper interface, and the execution layer orchestrates training, validation, testing, and metric computation. Solid arrows indicate execution flow, while dotted arrows indicate configuration injection from Hydra \citep{Yadan2019Hydra}. Additional implementation details are provided in Sections~\ref{sec:framework_formalization} and \ref{sec:picid_library}.}
\label{fig:architecture}
\end{figure}

This section defines the protocol specification: task targets, deterministic dataset construction, partitioning, and evaluation semantics. Raw unit data $(\gX,\gY)$ are transformed into aligned sequences $(\gZ,\gY')$ via $\mathcal{G}$ and $\widetilde{\mathcal{H}}$, then windowed into supervised instances $(\vW_k,y_k)$ by $\mathcal{S}$. Splits are defined under an explicit leakage policy, and evaluation uses a shared predictor/metric contract. Complete derivations are provided in Appendix~\ref{app:complete_formalization}; Section~\ref{sec:picid_library} shows how the software realizes these invariants.

\subsection{PHM tasks and target definitions}
\label{subsec:tasks}

For a monitored unit, let $\gX = \{\vx(t)\}_{t=1}^{T}$ and $\gY = \{y(t)\}_{t=1}^{T}$ denote the raw sensor measurements ($\vx(t) \in \mathbb{R}^{M}$, $M$ channels) and the associated task-dependent target signal ($y(t)\in\mathcal{Y}$) over $T$ discrete time steps. \picid{} formalizes three canonical PHM task families (introduced in Section~\ref{sec:intro} and further detailed in Appendix~\ref{app:intro_phm}), each characterized by the semantics and codomain of $y(t)$:
\begin{itemize}
    \item \textbf{Prognostics.} The target is a scalar $y(t)\in\mathbb{R}_{\ge 0}$ representing Remaining Useful Life (RUL), or a health indicator $h(t)\in[0,1]$ that decreases from $1$ (healthy) toward $0$ (failed).

    \item \textbf{Diagnostics.} The target is a discrete label $y(t)\in\{0,1,\dots,K-1\}$, where $K$ denotes the number of fault/mode classes and $y(t)$ indicates the class active at time $t$.

    \item \textbf{Fault Detection.} The supervisory signal is absent, and the objective is to infer nominal behavior directly from the observations. The model produces a real-valued fault score whose magnitude reflects deviation from the learned nominal behavior; after thresholding, this score yields a binary nominal/faulty label $y(t)\in\{0,1\}$.
\end{itemize}

Supervision is attached at the window level: each extracted window receives a single associated target, and this attachment remains valid even when the feature extraction pipeline changes the temporal grid (with explicit target alignment) (Section~\ref{subsec:construction}).

\subsection{Dataset construction and supervision}
\label{subsec:construction}

Raw time-series data are converted into supervised instances by three deterministic operators: feature transformation $\mathcal{G}$, target transformation+alignment $\widetilde{\mathcal{H}}$, and windowing $\mathcal{S}$. Because PHM pipelines often change temporal resolution, target alignment is treated as an explicit protocol component rather than an implementation detail (Appendix~\ref{app:feature_transform_full}--\ref{app:windowing_full}).

\paragraph{Feature transformation ($\mathcal{G}$).}
Let the transformed feature series be
\begin{equation}
    \gZ := \mathcal{G}(\gX;\Psi)
    = \{\vz(j)\}_{j=1}^{T'}, \qquad \vz(j)\in\mathbb{R}^{F},
\end{equation}
where $\Psi$ contains the fitted preprocessing parameters, $F$ is the post-transformation feature dimension, and $T'$ is the transformed length, which may differ from $T$ under windowed or time--frequency operators such as STFT or wavelets. In that case, we equip the transformed index set with a raw-time support map $a:\{1,\dots,T'\}\to\mathcal{I}_T$, where $\mathcal{I}_T$ denotes timestamps or raw-time intervals. The map $a(\cdot)$ is induced deterministically by $\mathcal{G}$ and defines aligned supervision via the target pipeline. All parameters $\Psi$ are estimated on the training partition only and then frozen for validation and test.

\paragraph{Target transformation and alignment ($\widetilde{\mathcal{H}}$).}
The target pipeline produces one aligned target value per transformed index $j\in\{1,\dots, T'\}$, yielding the aligned target sequence
\begin{equation}
    \gY' := \{z_y(j)\}_{j=1}^{T'}, \qquad
    z_y(j) := \widetilde{\mathcal{H}}\big(\gY,a(j);\Phi\big)
    = \mathcal{A}\Big(\mathcal{H}\big(\gY;\Phi\big),a(j)\Big),
\end{equation}
where $\Phi$ contains the fitted target-side parameters, $\mathcal{H}(\cdot;\Phi)$ applies pointwise target transformations on raw time, and $\mathcal{A}$ aligns to the support $a(j)$ by pointwise sampling, when $a(j)$ is a timestamp, or interval aggregation, e.g., last, mean, or majority. This decomposition records how labels are sampled or aggregated whenever feature extraction changes the temporal grid. All fitted parameters $\Phi$ and any statistics required by $\mathcal{A}$ are estimated on the training partition only (Appendix~\ref{app:target_transform_full}). Section~\ref{subsec:transforms_datasets} describes how these constraints are enforced structurally in the software.

\paragraph{Windowing and label alignment.}
\label{subsubsec:windowing}
The operator $\mathcal{S}$ maps aligned sequences $(\gZ,\gY')$ to supervised samples $\{(\vW_k,\,y_k)\}_{k\in\mathcal{K}}$ by extracting, for each admissible start index $k\in\mathcal{K}$, the feature window
\begin{equation}
    \vW_k = \big[\vz(k), \vz(k+1), \dots, \vz(k+L_{\mathrm{seq}}-1)\big]^\top \in \mathbb{R}^{L_{\mathrm{seq}}\times F},
\end{equation}
where $\mathcal{K}$ is determined by the history length $L_{\mathrm{seq}}$, stride $\Delta$, supervision offset $\delta$, and warm-start depth $\rho$. In the default PHM benchmark setting, supervision is attached at the end of the input window or at a fixed offset, yielding the window label
\begin{equation}
    y_k = z_y\big(k + L_{\mathrm{seq}} - 1 + \delta\big).
\end{equation}
Every model thus receives identical windowed inputs, identical supervision placement, and identical split-conditioned preprocessing state. Extended mechanics---including left-padding under $\rho > 0$, multi-step supervision, and multi-unit generalization---are provided in Appendix~\ref{app:windowing_full}. The software realization of this operator is described in Section~\ref{subsec:transforms_datasets}.

\subsection{Partitioning and evaluation}
\label{subsec:partitioning}

For multi-unit datasets, PICID supports two partitioning regimes that answer different generalization questions. \textbf{Inter-unit partitioning} assigns whole units disjointly to train/validation/test, measuring generalization to unseen units. \textbf{Intra-unit temporal partitioning} instead splits each unit chronologically into train/validation/test ranges, measuring within-unit prediction along a trajectory.

In both regimes, leakage control is part of the evaluation contract: fitted pipeline state for features and targets $(\Psi,\Phi)$ and hyper-parameter selection are determined using train and validation data only, while test data are accessed exclusively for final reporting. For intra-unit partitioning, ``train-only'' refers to earlier timestamps within each unit. The full formalization and software enforcement mechanisms are detailed in Appendix~\ref{app:partitioning_full} and Section~\ref{subsec:datasources}, respectively.

\paragraph{Evaluation protocol.}
\label{subsec:evaluation}

Any model evaluated under \picid must implement a predictor $f(\vW)=\hat{y}$ that maps an input window to a valid task output: a scalar for prognostics, a class prediction for diagnostics, or a class prediction/probability/fault score for fault detection (Section~\ref{subsec:tasks}). This predictor contract is task-agnostic: the same model family can be evaluated as a classifier or regressor under the same protocol contract, with task differences captured by the output space and the corresponding metric. Test performance $\mathcal{L}_{\mathrm{test}}$ is computed by averaging a per-sample metric $\ell(\hat{y},y)$ over the held-out test partition, or (when configured) by aggregating window predictions into per-unit trajectories and scoring a unit-level metric (Appendix~\ref{app:evaluation_full}). We report standard metrics (RMSE/MAE/PHM-score; accuracy/macro-F1; AUROC/AUPRC). All compared models share identical windows, splits, and metric definitions, so performance differences cannot be attributed to hidden dataset construction changes. The framework's extensibility interfaces are described in Section~\ref{subsec:extensibility}.

\section{The \picid{} library: architecture overview}
\label{sec:picid_library}

This section maps the formal protocol in Section~\ref{sec:framework_formalization} to executable software. In \picid, each experiment is defined by a structured YAML configuration (datasource, transforms, task definition, model wrapper, training settings), and the execution layer composes it into a deterministic run graph. Figure~\ref{fig:architecture} summarizes the module organization, and Appendix~\ref{app:library_architecture} lists the full module inventory and extension interfaces. Each subsection below states which formal operator it implements and what the corresponding software component does.\footnote{The framework also supports forecasting tasks through the same software abstractions, but forecasting is not a benchmarked claim in the present submission.}

\begin{table}[H]
\centering
\footnotesize
\caption{Module--protocol mapping. Each row links a library module to its formal role in Section~\ref{sec:framework_formalization}.}
\label{tab:module_protocol_map}
\renewcommand{\arraystretch}{1.15}
\begin{tabular}{@{}lll@{}}
\toprule
\textbf{Module} & \textbf{Protocol operator} & \textbf{Data flow} \\
\midrule
Datasources (\ref{subsec:datasources}) & Ingestion, split policy & Raw files $\to$ $\gX,\gY,\;\mathcal{D}^{\mathrm{train/val/test}}$ \\
Transform pipeline (\ref{subsec:transforms_datasets}) & $\mathcal{G}(\cdot;\Psi),\; \widetilde{\mathcal{H}}(\cdot;\Phi)$ & $\gX,\gY,\;\mathcal{D}$ $\to$ $\gZ,\;\gY'$ \\
Datasets \& sequencers (\ref{subsec:transforms_datasets}) & Windowing $\mathcal{S}$ & $\gZ,\gY'$ $\to$ $\{(\vW_k,y_k)\}_{k\in\mathcal{K}}$ \\
Wrappers + evaluator (\ref{subsec:evaluation_reporting}) & $f,\;\ell(\cdot,\cdot),\;\mathcal{L}_{\mathrm{test}}$ & $(f,\,\vW_k)$ $\to$ task metrics \\
\bottomrule
\end{tabular}
\end{table}

\subsection{Data ingestion and split-safe transport}
\label{subsec:datasources}

\emph{Protocol role.}\; $\text{Raw files} \xrightarrow{\text{load, split}} \gX,\gY,\;\mathcal{D}^{\mathrm{train}},\mathcal{D}^{\mathrm{val}},\mathcal{D}^{\mathrm{test}}$ \;(Section~\ref{subsec:tasks} and Section~\ref{subsec:partitioning}).

Execution begins at the datasource layer, which loads raw files, organizes units, and applies the split policy before any task-specific formatting occurs. \picid{} supports single-source, multi-source, and predefined-split datasources, allowing the same protocol to span batteries, bearings, turbofan engines, and other PHM domains. The codebase currently integrates more than twelve datasources, while the present paper benchmarks the subset documented in Appendix~\ref{sec:dataset_descriptions}. Split-aware typed containers then transport the resulting per-partition payloads across the pipeline while preserving metadata, unit identity, and structural consistency.

This separation is deliberate. Datasources handle only data-centric concerns---file I/O, unit grouping, and split semantics---while target alignment and window construction are deferred to downstream stages. The split policy, whether inter-unit or intra-unit temporal (Section~\ref{subsec:partitioning}), is fixed at the datasource boundary, turning partitioning and leakage control into library invariants rather than model-level implementation choices.

\subsection{Transformation and dataset construction}
\label{subsec:transforms_datasets}

\emph{Protocol role.}\; $\gX,\gY,\;\mathcal{D} \xrightarrow{\mathcal{G},\;\widetilde{\mathcal{H}},\;\mathcal{S}} \{(\vW_k,\,y_k)\}_{k\in\mathcal{K}}$ \;(Sections~\ref{subsec:construction}).

The transform system realizes the operators $\mathcal{G}$ and $\widetilde{\mathcal{H}}$. It reads raw feature and target keys from the typed container, fits any required preprocessing statistics on the training partition only, and writes the transformed and aligned representation back into pipeline state. Each transform is a composable unit with a logic block (the algorithm and its hyperparameters) and a metadata block controlling routing, fitting policy, and key assignment. The explicit \texttt{fit\_on} constraint restricts parameter estimation to the training partition, turning the train-only fitting rule from Section~\ref{subsec:construction} into a software-level invariant rather than a user convention. Because transforms are instantiated from declarative configuration, experiment definitions can be replayed exactly and audited at the level of routing, fitting policy, and parameterization. Appendix~\ref{app:yaml_example} provides configuration examples, a transform inventory, and preprocessing pseudocode.

In PHM, preprocessing often changes the effective prediction problem (e.g., when windowed or time--frequency feature extraction changes the temporal grid), so \picid{} treats transforms and alignment as part of the protocol rather than hidden implementation detail. Appendix~\ref{app:yaml_example} and Appendix~\ref{app:library_architecture} provide concrete transform examples and the full module inventory.

Datasets and sequencers then realize the windowing operator $\mathcal{S}$, mapping the aligned sequences $(\gZ,\gY')$ to model-ready samples $\{(\vW_k,\,y_k)\}_{k\in\mathcal{K}}$ while preserving the protocol semantics from Section~\ref{subsubsec:windowing} (window length, stride, supervision placement, and target interpretation). This task-centric layer allows the same preprocessed trajectory to support multiple formulations: for example, from the same window representation one can regress end-of-window RUL for prognostics or predict a nominal/faulty label (or fault score) for fault detection, without redefining the upstream transformation contract. Details of ragged multi-unit handling are given in Appendix~\ref{app:library_architecture}.

\subsection{Model execution and evaluation}
\label{subsec:evaluation_reporting}

\emph{Protocol role.}\; $\vW_k \xrightarrow{f} \hat{y}_k \xrightarrow{\ell} \mathcal{L}_{\mathrm{test}}$ \;(Section~\ref{subsec:evaluation}).

Downstream of dataset construction, \picid{} unifies model execution through wrapper classes that provide a common prediction interface for heterogeneous backbones. Feed-forward wrappers support batchwise deep learning models (LSTM, CNN1D, PatchTST, Crossformer, TiDE), while fit-predict wrappers support scikit-learn-compatible estimators (XGBoost, TabPFN, AutoGluon, Isolation Forest) and other non-gradient baselines. All wrappers emit a standardized output dictionary containing predictions $\hat{y}_k$ alongside the aligned targets $y_k$. This is what allows sequence models, tree ensembles, tabular foundation models, and fault detectors to be compared under identical preprocessing, identical windows, and identical reporting logic.
The same wrapper interface is used for both diagnostics and prognostics, so model families are compared across tasks under identical data contracts and evaluation boundaries.

The evaluator subsystem computes task-specific metrics on standardized $(\hat{y}_k, y_k)$ pairs, optionally applying inverse scaling so that metrics are reported in physical units. Specialized evaluators cover regression (RMSE, MAE), classification (accuracy, macro-F1), and fault detection (AUROC, AUPRC). A pluggable hook system extends the evaluator with actions such as saving predictions, logging trend plots, or exporting visualizations without modifying the core metric logic.

\subsection{Protocol-preserving extensibility}
\label{subsec:extensibility}

\picid{} separates \emph{invariant} protocol components---task definitions, transformation operators, window construction, partitioning rules, and evaluation metrics---from \emph{extensible} research components such as new datasets, models, and transforms. New components can be added, but they must respect the same interfaces and leakage constraints formalized in Section~\ref{sec:framework_formalization}. In practice, this means that adding a new datasource requires only defining the raw data contract; adding a new model requires only implementing the wrapper interface; and adding a new transform requires only specifying its routing and fitting metadata. Public extension interfaces are documented in Appendix~\ref{app:lib_interfaces}.\footnote{An optional tabular representation adapter, including a protocol for tabular foundation models and in-context learning under the same leakage constraints, is provided in Appendix~\ref{app:tabularization_full}.}

Beyond modularity, \picid{} is engineered for trustworthy reuse: split-aware typed containers preserve unit identity and split membership across the pipeline, and each run records the experiment configuration and code version. Deterministic seeding makes runs stable, and automated tests guard against regressions across the protocol stack. Full reproducibility methodology is detailed in Appendix~\ref{app:reproducibility}.
\section{Empirical validation}
\label{sec:empirical_validation}

\subsection{Experimental setup}
\label{subsec:exp_setup}

To demonstrate that \picid{} produces consistent, comparable results across heterogeneous PHM tasks, we evaluate a diverse set of models on a representative subset of the integrated datasources. All experiments are executed through the unified protocol described in Section~\ref{sec:framework_formalization}, using identical preprocessing, windowing, and evaluation logic for every model--dataset pair. Each configuration is run with five independent seeds; Appendix~\ref{app:reproducibility} provides full reproducibility details.

\paragraph{Datasets.}
We select datasources spanning two PHM task categories formalized in this paper: (i)~\textit{prognostics}---N-CMAPSS-DS02 and N-CMAPSS-P~\citep{arias2021aircraft, frederick2007user} (jet-engine RUL), NB14~\citep{bole2014adaptation} and Unibo~\citep{univbo_dataset} (battery ah-RUL), PHME20~\citep{PHME20-GTU} (industrial filtration RUL), and XJTU-SY~\citep{yaguo2019xjtu} (bearing RUL); (ii)~\textit{diagnostics}---N-CMAPSS Multi-source concepts~\citep{arias2021aircraft, frederick2007user} (concept classification), HSF15~\citep{hsf15_helwig} (four hydraulic component tasks: accumulator, cooler, pump, valve), and MZVAV~\citep{Granderson2020} (building-system fault classification). Full dataset descriptions and split are described in Appendix~\ref{sec:dataset_descriptions}.

\paragraph{Models.}
We incorporate and evaluate models from different categories and domains: (i)~\textit{simple baselines}---linear regression (Linear), exponential regression (Exp), and MLP; (ii)~\textit{deep sequence models}---LSTM, CNN-1D, and TiDE; (iii)~\textit{transformers}---Timeseries Transformer (TST), Spacetimeformer (STF), Crossformer (CF), and PatchTST (PTST); (iv)~\textit{tabular models}---XGBoost; (v)~\textit{tabular foundation models}---TabPFN and TabDPT. Please refer to Appendix~\ref{app:models} for more details.
Each model family is instantiated in both task types (diagnostics classifier, prognostics regressor) and evaluated under the same preprocessing, windowing, and splitting, enabling cross-task comparisons.

\subsection{Results}
\label{subsec:results}

Table~\ref{tab:results_main_combined} reports performance across all 12 datasets and 13 models, with prognostics evaluated by normalized MAE ($\times 100$; top floor, $\downarrow$) and diagnostics by F1 ($\times 100$; bottom floor, $\uparrow$). All 150 model--dataset pairs (13~models~$\times$~6~prognostics~datasets and 12~models~$\times$~6~diagnostics~datasets;) ran end-to-end under the unified protocol of Section~\ref{sec:framework_formalization}, demonstrating that the framework supports the heterogeneous PHM models and datasets domains. Per-task rankings vary substantially across families---no single architecture dominates every column---while learned methods consistently improve on the Linear, Exp and MLP baselines and all five model categories rank in the top half on at least one dataset, showing that the unified abstractions accommodate diverse architectures on equal footing. Per-metric variants---AUROC and Accuracy for diagnostics; MSE, denormalized errors, and PHM/NASA scores for prognostics---are reported in Appendix~\ref{app:full_experiments}.

A secondary purpose of the unified benchmark is to illustrate that comparable evaluation conditions can surface cross-architecture patterns that are invisible in isolated per-paper studies. We highlight two such patterns here not as scientific conclusions of this work---a detailed analysis is outside the scope of this infrastructure paper---but as evidence that the framework produces meaningful signal when models are evaluated on equal footing. First, tabular foundation models (TabDPT and TabPFN) achieve the top two combined ranks across both task categories, outperforming all deep sequence models and transformers on average; this pattern emerges specifically because identical preprocessing, splits, and evaluation logic remove the confounds that typically make cross-architecture comparison unreliable in PHM. Because the same model families are evaluated as both classifiers and regressors under an identical protocol, these patterns reflect cross-task generalization behavior rather than differences in task-specific evaluation pipelines. Second, the transformer family (Crossformer, TST, STF, PatchTST) exhibits a consistent and pronounced asymmetry: these models perform competitively on several prognostics tasks, yet collapse to near-random performance on most diagnostics tasks (F1 of 17--40 on NC-D and MZVAV), a contrast that only becomes apparent when the same models are evaluated across both task families under a common protocol. Whether these patterns generalize beyond the datasets benchmarked here is a question for domain-focused follow-up work; the point for present purposes is that the infrastructure makes such comparisons structurally sound and reproducible.

\begin{table}[t]
\centering
\scriptsize
\setlength{\tabcolsep}{4pt}
\renewcommand{\arraystretch}{0.92}
\caption{Main evaluation results. \textbf{Top block:} prognostics, measured by MAE in the normalized target space ($\times 100$) ($\downarrow$). \textbf{Bottom block:} diagnostics, measured by F1 score ($\times 100$) ($\uparrow$). Models are grouped by family: simple baselines, deep sequence models, transformers, tabular models, and tabular foundation models. \textbf{Bold}/\underline{underline} denote the best/second-best result.}
\label{tab:results_main_combined}
\begin{adjustbox}{max width=\textwidth}
\begin{tabular}{lrrrrrrr}
\toprule
Model & \multicolumn{1}{c}{NC-DS02$\downarrow$} & \multicolumn{1}{c}{NC-P$\downarrow$} & \multicolumn{1}{c}{NB14$\downarrow$} & \multicolumn{1}{c}{PHME20$\downarrow$} & \multicolumn{1}{c}{Unibo$\downarrow$} & \multicolumn{1}{c}{XJTU-SY$\downarrow$} & \multicolumn{1}{c}{Avg rank} \\
\midrule
Linear & 10.13 ± 0.14 & 16.11 ± 0.60 & 41.69 ± 12.02 & 12.19 ± 0.36 & 27.59 ± 14.36 & 76.80 ± 60.41 & 12.50 \\
Exp & 5.35 ± 0.06 & 10.96 ± 0.09 & 30.47 ± 47.76 & 8.82 ± 0.52 & 12.19 ± 0.31 & 27.22 ± 4.06 & 9.67 \\
MLP & 6.37 ± 0.23 & 13.17 ± 0.78 & 14.38 ± 9.77 & 4.62 ± 1.15 & 12.50 ± 0.76 & 30.64 ± 2.67 & 10.33 \\
LSTM & \underline{4.93 ± 0.13} & 7.56 ± 0.31 & 3.80 ± 0.22 & 3.73 ± 0.98 & 6.50 ± 0.16 & \cellcolor[gray]{0.85}\textbf{21.89 ± 0.40} & 3.67 \\
CNN-1D & 5.33 ± 0.37 & 7.53 ± 0.22 & 8.89 ± 1.70 & 5.35 ± 3.71 & 12.41 ± 1.15 & 31.02 ± 8.25 & 8.67 \\
TiDE & 5.29 ± 0.22 & 7.62 ± 0.20 & \cellcolor[gray]{0.85}\textbf{3.44 ± 0.17} & 4.20 ± 0.66 & 6.46 ± 0.78 & 25.11 ± 2.38 & 5.17 \\
TST & 5.31 ± 0.13 & \underline{7.02 ± 0.17} & 6.28 ± 0.25 & 4.11 ± 0.84 & 7.23 ± 0.39 & 33.30 ± 7.72 & 7.00 \\
STF & \cellcolor[gray]{0.85}\textbf{4.89 ± 0.10} & 7.35 ± 1.16 & 10.67 ± 3.16 & 3.91 ± 1.00 & 8.89 ± 0.81 & 28.49 ± 4.01 & 6.17 \\
CF & 5.76 ± 0.51 & 9.98 ± 0.57 & \underline{3.57 ± 0.07} & 3.87 ± 0.85 & 5.58 ± 1.08 & \underline{22.09 ± 1.06} & 5.00 \\
PTST & 16.62 ± 0.04 & 21.55 ± 0.03 & 5.22 ± 0.10 & 15.09 ± 1.13 & 11.18 ± 1.11 & 25.42 ± 1.48 & 10.33 \\
XGBoost & 8.52 ± 0.00 & 15.24 ± 0.00 & 4.48 ± 0.00 & 2.68 ± 0.00 & 4.06 ± 0.00 & 24.59 ± 0.00 & 6.50 \\
TabPFN & 4.96 ± 0.04 & 7.79 ± 0.04 & 3.91 ± 0.03 & \cellcolor[gray]{0.85}\textbf{1.95 ± 0.03} & \cellcolor[gray]{0.85}\textbf{3.72 ± 0.06} & 22.27 ± 0.35 & 3.33 \\
TabDPT & 5.07 ± 0.06 & \cellcolor[gray]{0.85}\textbf{6.85 ± 0.02} & 3.63 ± 0.04 & \underline{2.19 ± 0.01} & \underline{3.94 ± 0.05} & 23.24 ± 0.45 & 2.67 \\
\midrule
Model & \multicolumn{1}{c}{NC-D$\uparrow$} & \multicolumn{1}{c}{HSF15-A$\uparrow$} & \multicolumn{1}{c}{HSF15-C$\uparrow$} & \multicolumn{1}{c}{HSF15-P$\uparrow$} & \multicolumn{1}{c}{HSF15-V$\uparrow$} & \multicolumn{1}{c}{MZVAV$\uparrow$} & \multicolumn{1}{c}{Avg rank} \\
\midrule
Linear & 72.34 ± 3.04 & 58.75 ± 1.81 & 98.35 ± 0.82 & 54.40 ± 11.97 & 32.55 ± 2.36 & 39.89 ± 8.77 & 7.33 \\
MLP & 79.49 ± 1.80 & 91.02 ± 2.25 & \underline{99.91 ± 0.14} & 97.32 ± 0.60 & 80.99 ± 29.38 & 60.10 ± 6.39 & 5.00 \\
LSTM & \cellcolor[gray]{0.85}\textbf{88.84 ± 0.73} & 94.59 ± 0.97 & \cellcolor[gray]{0.85}\textbf{100.00 ± 0.00} & 95.94 ± 2.83 & 97.35 ± 3.32 & 51.31 ± 6.01 & 3.83 \\
CNN-1D & \underline{87.53 ± 2.83} & 94.03 ± 1.98 & \cellcolor[gray]{0.85}\textbf{100.00 ± 0.00} & 98.73 ± 0.52 & 97.92 ± 0.83 & \underline{66.11 ± 5.74} & 3.00 \\
TiDE & 32.57 ± 4.65 & 42.90 ± 5.07 & 61.57 ± 10.41 & 59.37 ± 7.68 & 42.11 ± 14.36 & 25.19 ± 5.29 & 8.17 \\
TST & 26.34 ± 3.96 & 37.37 ± 4.38 & 59.18 ± 10.29 & 46.07 ± 4.13 & 35.40 ± 5.77 & 24.93 ± 4.38 & 10.00 \\
STF & 24.55 ± 3.60 & 40.01 ± 5.75 & 65.94 ± 20.61 & 50.39 ± 11.54 & 37.34 ± 5.68 & 38.01 ± 5.56 & 8.67 \\
CF & 23.74 ± 0.93 & 25.04 ± 5.28 & 59.19 ± 10.05 & 29.46 ± 2.61 & 23.98 ± 2.95 & 17.34 ± 4.24 & 11.50 \\
PTST & 19.57 ± 0.26 & 31.56 ± 4.04 & 41.57 ± 5.18 & 41.22 ± 6.63 & 26.20 ± 2.44 & 25.81 ± 4.15 & 11.00 \\
XGBoost & 48.13 ± 0.00 & \underline{98.07 ± 0.00} & \cellcolor[gray]{0.85}\textbf{100.00 ± 0.00} & \underline{99.66 ± 0.00} & \underline{99.65 ± 0.00} & 57.08 ± 0.00 & 3.17 \\
TabPFN & 67.15 ± 1.46 & \cellcolor[gray]{0.85}\textbf{99.47 ± 0.00} & \cellcolor[gray]{0.85}\textbf{100.00 ± 0.00} & \cellcolor[gray]{0.85}\textbf{100.00 ± 0.00} & \cellcolor[gray]{0.85}\textbf{100.00 ± 0.00} & 58.32 ± 2.44 & 2.33 \\
TabDPT & 85.21 ± 0.16 & 96.66 ± 1.03 & \cellcolor[gray]{0.85}\textbf{100.00 ± 0.00} & 99.06 ± 0.25 & 98.92 ± 0.35 & \cellcolor[gray]{0.85}\textbf{71.29 ± 0.48} & 2.33 \\
\bottomrule
\end{tabular}
\end{adjustbox}
\end{table}

\section{Conclusion and outlook}
\label{sec:conclusions}

Progress in PHM is increasingly constrained not only by datasets and models, but by the execution protocols  governing preprocessing, target alignment, partitioning, and evaluation. \picid~addresses this challenge by coupling a formal multi-task specification with executable protocol abstractions that enforce benchmark invariants directly in the experimental stack. We demonstrated this approach across diagnostics and prognostics on twelve datasets spanning batteries, bearings, turbofan engines, hydraulics, filtration, and buildings, and across thirteen models ranging from classical baselines to deep sequence, transformer, tabular, and tabular foundation models. Across this heterogeneous collection of datasets, tasks, and models, a single unified protocol executes the evaluation pipeline end-to-end while enabling controlled and comparable cross-family evaluation. These results support protocol-centric benchmarking as a foundation for more reproducible, auditable, and cumulative PHM research. By standardizing evaluation contracts across tasks, \picid~also enables fair cross-task comparison, helping disentangle architecture effects from hidden protocol variation.

The present work also has important limitations. First, while \picid~standardizes execution protocols, it does not resolve fundamental challenges related to dataset quality, label noise, domain shift, or inconsistent failure definitions across PHM benchmarks. As a result, protocol consistency alone cannot guarantee scientifically meaningful comparison when the underlying datasets are themselves heterogeneous or weakly standardized. Second, the current benchmark suite, although diverse, remains limited in scale relative to modern ML evaluation efforts and does not yet capture many industrial settings, sensing regimes, or long-horizon deployment conditions encountered in practice. Third, the framework currently focuses on supervised diagnostics and prognostics under fixed offline evaluation protocols; continual learning, online adaptation, uncertainty-aware evaluation, and streaming settings remain outside the present scope. Finally, although \picid~makes evaluation protocols explicit and executable, some protocol choices in PHM remain inherently application-dependent, including failure definitions, labeling strategies, and operational evaluation criteria. Achieving broader comparability across PHM domains will therefore require not only shared infrastructure, but also stronger community-level standardization of task semantics and evaluation practice.

\newpage
\bibliographystyle{unsrtnat}
\bibliography{references.bib}

@inproceedings{lhoest-etal-2021-datasets,
    title = "Datasets: A Community Library for Natural Language Processing",
    author = "Lhoest, Quentin  and
      Villanova del Moral, Albert  and
      Jernite, Yacine  and
      Thakur, Abhishek  and
      von Platen, Patrick  and
      Patil, Suraj  and
      Chaumond, Julien  and
      Drame, Mariama  and
      Plu, Julien  and
      Tunstall, Lewis  and
      Davison, Joe  and
      {\v{S}}a{\v{s}}ko, Mario  and
      Chhablani, Gunjan  and
      Malik, Bhavitvya  and
      Brandeis, Simon  and
      Le Scao, Teven  and
      Sanh, Victor  and
      Xu, Canwen  and
      Patry, Nicolas  and
      McMillan-Major, Angelina  and
      Schmid, Philipp  and
      Gugger, Sylvain  and
      Delangue, Cl{\'e}ment  and
      Matussi{\`e}re, Th{\'e}o  and
      Debut, Lysandre  and
      Bekman, Stas  and
      Cistac, Pierric  and
      Goehringer, Thibault  and
      Mustar, Victor  and
      Lagunas, Fran{\c{c}}ois  and
      Rush, Alexander  and
      Wolf, Thomas",
    booktitle = "Proceedings of the 2021 Conference on Empirical Methods in Natural Language Processing: System Demonstrations",
    month = nov,
    year = "2021",
    address = "Online and Punta Cana, Dominican Republic",
    publisher = "Association for Computational Linguistics",
    url = "https://aclanthology.org/2021.emnlp-demo.21",
    pages = "175--184",
    eprint={2109.02846},
    archivePrefix={arXiv},
    primaryClass={cs.CL},
}

@inproceedings{fey2019pyg,
  title={Fast Graph Representation Learning with {PyTorch Geometric}},
  author={Fey, Matthias and Lenssen, Jan E.},
  booktitle={ICLR Workshop on Representation Learning on Graphs and Manifolds},
  year={2019},
}

@article{stable-baselines3,
  author  = {Antonin Raffin and Ashley Hill and Adam Gleave and Anssi Kanervisto and Maximilian Ernestus and Noah Dormann},
  title   = {Stable-Baselines3: Reliable Reinforcement Learning Implementations},
  journal = {Journal of Machine Learning Research},
  year    = {2021},
  volume  = {22},
  number  = {268},
  pages   = {1-8},
  url     = {http://jmlr.org/papers/v22/20-1364.html}
}

@article{yaguo2019xjtu,
  title={XJTU-SY rolling element bearing accelerated life test datasets: A tutorial},
  author={Yaguo, LEI and Tianyu, HAN and Biao, WANG and Naipeng, LI and Tao, YAN and Jun, YANG},
  journal={Journal of Mechanical Engineering},
  volume={55},
  number={16},
  pages={1--6},
  year={2019}
}

@article{vonhahn2023reproducibility,
title={Computational Reproducibility Within Prognostics and Health Management},
volume={2}, 
url={https://ojs.istp-press.com/dmd/article/view/141}, 
DOI={10.37965.jdmd.2023.141},
number={1}, 
journal={Journal of Dynamics, Monitoring and Diagnostics}, 
author={von Hahn, Tim and Mechefske, Chris K.}, year={2023}, 
month={Feb.}, 
pages={42–50} 
}

@inproceedings{bole2014adaptation,
  title={Adaptation of an electrochemistry-based li-ion battery model to account for deterioration observed under randomized use},
  author={Bole, Brian and Kulkarni, Chetan S and Daigle, Matthew},
  booktitle={Annual conference of the PHM society},
  volume={6},
  year={2014}
}

@article{bosello2023charge,
  title={To charge or to sell? EV pack useful life estimation via LSTMs, CNNs, and autoencoders},
  author={Bosello, Michael and Falcomer, Carlo and Rossi, Claudio and Pau, Giovanni},
  journal={Energies},
  volume={16},
  number={6},
  pages={2837},
  year={2023},
  publisher={MDPI}
}

@inproceedings{univbo_dataset,
author = {Wong, Kei Long and Bosello, Michael and Tse, Rita and Falcomer, Carlo and Rossi, Claudio and Pau, Giovanni},
title = {Li-Ion Batteries State-of-Charge Estimation Using Deep LSTM at Various Battery Specifications and Discharge Cycles},
year = {2021},
isbn = {9781450384780},
publisher = {Association for Computing Machinery},
address = {New York, NY, USA},
url = {https://doi.org/10.1145/3462203.3475878},
doi = {10.1145/3462203.3475878},
booktitle = {Proceedings of the Conference on Information Technology for Social Good},
pages = {85–90},
numpages = {6},
location = {Roma, Italy},
series = {GoodIT '21}
}

@inproceedings{nectoux2012pronostia,
  title={PRONOSTIA: An experimental platform for bearings accelerated degradation tests.},
  author={Nectoux, Patrick and Gouriveau, Rafael and Medjaher, Kamal and Ramasso, Emmanuel and Chebel-Morello, Brigitte and Zerhouni, Noureddine and Varnier, Christophe},
  booktitle={IEEE International Conference on Prognostics and Health Management, PHM'12.},
  pages={1--8},
  year={2012},
  organization={IEEE Catalog Number: CPF12PHM-CDR}
}

@inproceedings{hollmanntabpfn,
  title={TabPFN: A Transformer That Solves Small Tabular Classification Problems in a Second},
  author={Hollmann, Noah and M{\"u}ller, Samuel and Eggensperger, Katharina and Hutter, Frank},
  booktitle={The Eleventh International Conference on Learning Representations},
    year={2023} 
}

@article{hollmann2025accurate,
  title={Accurate predictions on small data with a tabular foundation model},
  author={Hollmann, Noah and M{\"u}ller, Samuel and Purucker, Lennart and Krishnakumar, Arjun and K{\"o}rfer, Max and Hoo, Shi Bin and Schirrmeister, Robin Tibor and Hutter, Frank},
  journal={Nature},
  volume={637},
  number={8045},
  pages={319--326},
  year={2025},
  publisher={Nature Publishing Group UK London}
}

@article{ma2024tabdpt,
  title={Tabdpt: Scaling tabular foundation models},
  author={Ma, Junwei and Thomas, Valentin and Hosseinzadeh, Rasa and Kamkari, Hamidreza and Labach, Alex and Cresswell, Jesse C and Golestan, Keyvan and Yu, Guangwei and Volkovs, Maksims and Caterini, Anthony L},
  journal={arXiv preprint arXiv:2410.18164},
  year={2024}
}

@inproceedings{zhang2023crossformer,
  title={Crossformer: Transformer utilizing cross-dimension dependency for multivariate time series forecasting},
  author={Zhang, Yunhao and Yan, Junchi},
  booktitle={The eleventh international conference on learning representations},
  year={2023}
}

@misc{grigsby2021longrange,
      title={Long-Range Transformers for Dynamic Spatiotemporal Forecasting}, 
      author={Jake Grigsby and Zhe Wang and Yanjun Qi},
      year={2021},
      eprint={2109.12218},
      archivePrefix={arXiv},
      primaryClass={cs.LG}
}

@inproceedings{Yuqietal-2023-PatchTST,
  title     = {A Time Series is Worth 64 Words: Long-term Forecasting with Transformers},
  author    = {Nie, Yuqi and
               H. Nguyen, Nam and
               Sinthong, Phanwadee and 
               Kalagnanam, Jayant},
  booktitle = {International Conference on Learning Representations},
  year      = {2023}
}

@article{chen2016xgboost,
  title={XGBoost: A Scalable Tree Boosting System},
  author={Chen, Tianqi},
  journal={Cornell University},
  year={2016}
}

@article{lecunDeepLearning2015,
  author  = {LeCun, Yann and Bengio, Yoshua and Hinton, Geoffrey},
  title   = {Deep Learning},
  journal = {Nature},
  volume  = {521},
  number  = {7553},
  pages   = {436--444},
  year    = {2015},
  doi     = {10.1038/nature14539}
}

@article{daslong,
  title={Long-term Forecasting with TiDE: Time-series Dense Encoder},
  author={Das, Abhimanyu and Kong, Weihao and Leach, Andrew and Mathur, Shaan K and Sen, Rajat and Yu, Rose},
  journal={Transactions on Machine Learning Research},
year={2023}
}

@article{siami-naminiPerformanceLSTMBiLSTM2019,
  author  = {Siami-Namini, Sima and Tavakoli, Neda and Siami Namin, Akbar},
  title   = {The Performance of LSTM and BiLSTM in Forecasting Time Series},
  journal = {IEEE Access},
  volume  = {7},
  pages   = {92485--92494},
  year    = {2019},
  doi     = {10.1109/ACCESS.2019.2923712}
}

@article{zhao2020deep,
  title   = {Deep learning algorithms for rotating machinery intelligent diagnosis: An open source benchmark study},
  author  = {Zhao, Zhibin and Li, Tianfu and Wu, Jingyao and Sun, Chuang and Wang, Shibin and Yan, Ruqiang and Chen, Xuefeng},
  journal = {ISA Transactions},
  volume  = {107},
  pages   = {224--255},
  year    = {2020},
  doi     = {10.1016/j.isatra.2020.08.010}
}

@article{von2022computational,
  title={Computational reproducibility within prognostics and health management},
  author={von Hahn, Tim and Mechefske, Chris K},
  journal={arXiv preprint arXiv:2205.15489},
  year={2022}
}

@article{SOLISMARTIN2025102039,
	title = {PHMD: An easy data access tool for prognosis and health management datasets},
	journal = {SoftwareX},
	volume = {29},
	pages = {102039},
	year = {2025},
	issn = {2352-7110},
	doi = {https://doi.org/10.1016/j.softx.2025.102039},
	url = {https://www.sciencedirect.com/science/article/pii/S2352711025000068},
	author = {David Solís-Martín and Juan Galán-Páez and Joaquín Borrego-Díaz},
}

@inproceedings{sikorska2016collaborative,
  title={A collaborative data library for testing prognostic models},
  author={Sikorska, Joanna and Hodkiewicz, Melinda and D'Cruz, Ashwin and Astfalck, Lachlan and Keating, Adrian},
  booktitle={PHM Society European Conference},
  volume={3},
  number={1},
  year={2016}
}

@article{teubert2023progpy,
  title={ProgPy: Python packages for prognostics and health management of engineering systems},
  author={Teubert, Christopher and Jarvis, Katelyn and Corbetta, Matteo and Kulkarni, Chetan and Daigle, Matthew},
  journal={Journal of Open Source Software},
  volume={8},
  number={87},
  pages={5099},
  year={2023}
}

@inproceedings{jia2018review,
  title={A review of PHM Data Competitions from 2008 to 2017: Methodologies and Analytics},
  author={Jia, Xiaodong and Huang, Bin and Feng, Jianshe and Cai, Haoshu and Lee, Jay},
  booktitle={Proceedings of the Annual Conference of the Prognostics and Health Management Society},
  pages={1--10},
  year={2018}
}

@article{Li2026,
  title = {PHM-Vibench: A unified,  extensible,  and reproducible vibration benchmarking framework for prognostics and health management},
  volume = {5},
  ISSN = {2994-7219},
  url = {http://dx.doi.org/10.36001/phmap.2025.v5i1.4303},
  DOI = {10.36001/phmap.2025.v5i1.4303},
  number = {1},
  journal = {PHM Society Asia-Pacific Conference},
  publisher = {PHM Society},
  author = {Li,  Qi and Chen,  Bojian and Li,  Xuan and Chen,  Qitong and Chen,  Liang and Shen,  Changqing and Lu,  Lu and Qin,  Zhaoye and Chu,  Fulei},
  year = {2026},
  month = jan 
}

@article{zhao2024domain,
  title={Domain generalization for cross-domain fault diagnosis: An application-oriented perspective and a benchmark study},
  author={Zhao, Chao and Zio, Enrico and Shen, Weiming},
  journal={Reliability Engineering \& System Safety},
  volume={245},
  pages={109964},
  year={2024},
  publisher={Elsevier}
}

@article{hendriks2022towards,
  title={Towards better benchmarking using the CWRU bearing fault dataset},
  author={Hendriks, Jacob and Dumond, Patrick and Knox, DA},
  journal={Mechanical Systems and Signal Processing},
  volume={169},
  pages={108732},
  year={2022},
  publisher={Elsevier}
}

@inproceedings{matania2024test,
  title={Test-training leakage in evaluation of machine learning algorithms for condition-based maintenance},
  author={Matania, Omri and Cohen, Roee and Bechhoefer, Eric and Bortman, Jacob},
  booktitle={PHM Society European Conference},
  volume={8},
  number={1},
  pages={13--13},
  year={2024}
}

@inproceedings{uckun2008standardizing,
  title={Standardizing research methods for prognostics},
  author={Uckun, Serdar and Goebel, Kai and Lucas, Peter JF},
  booktitle={2008 international conference on prognostics and health management},
  pages={1--10},
  year={2008},
  organization={IEEE}
}

@inproceedings{astfalck2016modelling,
  title={A modelling ecosystem for prognostics},
  author={Astfalck, Lachlan and Hodkiewicz, Melinda and Keating, Adrian and Cripps, Edward and Pecht, Michael},
  booktitle={Annual Conference of the PHM Society},
  volume={8},
  number={1},
  year={2016}
}

@article{Su2024,
  title = {Machine Learning Approaches for Diagnostics and Prognostics of Industrial Systems Using Open Source Data from PHM Data Challenges: A Review},
  volume = {15},
  ISSN = {2153-2648},
  url = {http://dx.doi.org/10.36001/ijphm.2024.v15i2.3993},
  DOI = {10.36001/ijphm.2024.v15i2.3993},
  number = {2},
  journal = {International Journal of Prognostics and Health Management},
  publisher = {PHM Society},
  author = {Su,  Hanqi and Lee,  Jay},
  year = {2024},
  month = sep 
}

@article{hagmeyer2021creation,
  title={Creation of publicly available data sets for prognostics and diagnostics addressing data scenarios relevant to industrial applications},
  author={Hagmeyer, Simon and Mauthe, Fabian and Zeiler, Peter},
  journal={International Journal of Prognostics and Health Management},
  volume={12},
  number={2},
  year={2021}
}

@inproceedings{mauthe2025overview,
  title={Overview and analysis of publicly available degradation data sets for tasks within prognostics and health management},
  author={Mauthe, Fabian and Steinmann, Luca and Neu, Moritz and Zeiler, Peter},
  booktitle={35th european safety and reliability conference.(accepted). Research Publishing},
  year={2025}
}

@inproceedings{bishay2023design,
  title={Design and Implementation of a Model Selection Pipeline for Prognostics and Health Management in the Operational Environment},
  author={Bishay, Peter and Sarah, Lukens and Damon, Rousis and Nathan, Danneman},
  booktitle={Annual Conference of the PHM Society},
  volume={15},
  number={1},
  year={2023}
}

@article{jardine2006review,
  title={A review on machinery diagnostics and prognostics implementing condition-based maintenance},
  author={Jardine, Andrew KS and Lin, Daming and Banjevic, Dragan},
  journal={Mechanical Systems and Signal Processing},
  volume={20},
  number={7},
  pages={1483--1510},
  year={2006},
  publisher={Elsevier},
  doi={10.1016/j.ymssp.2005.09.012}
}

@article{lee2014prognostics,
  title={Prognostics and health management design for rotary machinery systems---Reviews, methodology and applications},
  author={Lee, Jay and Wu, Fangji and Zhao, Wenyu and Ghaffari, Masoud and Liao, Linxia and Siegel, David},
  journal={Mechanical Systems and Signal Processing},
  volume={42},
  number={1-2},
  pages={314--334},
  year={2014},
  publisher={Elsevier},
  doi={10.1016/j.ymssp.2013.06.004}
}

@article{zonta2020predictive,
  title={Predictive maintenance in the Industry 4.0: A systematic literature review},
  author={Zonta, Tiago and Da Costa, Cristiano Andr{\'e} and da Rosa Righi, Rodrigo and De Lima, Miromar Jos{\'e} and Da Trindade, Eduardo Silveira and Li, Guann Pyng},
  journal={Computers \& Industrial Engineering},
  volume={150},
  pages={106889},
  year={2020},
  publisher={Elsevier},
  doi={10.1016/j.cie.2020.106889}
}

@article{li2024small,
  title={Small data challenges for intelligent prognostics and health management: a review},
  author={Li, Chuanjiang and Li, Shaobo and Feng, Yixiong and Gryllias, Konstantinos and Gu, Fengshou and Pecht, Michael},
  journal={Artificial Intelligence Review},
  volume={57},
  number={8},
  pages={214},
  year={2024},
  publisher={Springer},
  doi={10.1007/s10462-024-10820-4}
}

@article{sim2020tutorial,
  title={A tutorial for feature engineering in the prognostics and health management of gears and bearings},
  author={Sim, Jinwoo and Kim, Seokgoo and Park, Hyung Jun and Choi, Joo-Ho},
  journal={Applied Sciences},
  volume={10},
  number={16},
  pages={5639},
  year={2020},
  publisher={MDPI},
  doi={10.3390/app10165639}
}

@article{zhao2019deep,
  title={Deep learning and its applications to machine health monitoring},
  author={Zhao, Rui and Yan, Ruqiang and Chen, Zhenghua and Mao, Kezhi and Wang, Peng and Gao, Robert X},
  journal={Mechanical Systems and Signal Processing},
  volume={115},
  pages={213--237},
  year={2019},
  publisher={Elsevier},
  doi={10.1016/j.ymssp.2018.05.050}
}

@inproceedings{helwig2015condition,
  title={Condition monitoring of a complex hydraulic system using multivariate statistics},
  author={Helwig, Nikolai and Pignanelli, Eliseo and Sch{\"u}tze, Andreas},
  booktitle={2015 IEEE International Instrumentation and Measurement Technology Conference (I2MTC) Proceedings},
  pages={210--215},
  year={2015},
  organization={IEEE},
  doi={10.1109/I2MTC.2015.7151267}
}

@article{severson2019data,
  title={Data-driven prediction of battery cycle life before capacity degradation},
  author={Severson, Kristen A and Attia, Peter M and Jin, Norman and Perkins, Nicholas and Jiang, Benben and Yang, Zi and Chen, Michael H and Aykol, Muratahan and Herring, Patrick K and Fraggedakis, Dimitrios and others},
  journal={Nature Energy},
  volume={4},
  number={5},
  pages={383--391},
  year={2019},
  publisher={Nature Publishing Group},
  doi={10.1038/s41560-019-0356-8}
}

@article{gluonts,
  author  = {Alexander Alexandrov and Konstantinos Benidis and Michael Bohlke-Schneider and Valentin Flunkert and Jan Gasthaus and Tim Januschowski and Danielle C. Maddix and Syama Rangapuram and David Salinas and Jasper Schulz and Lorenzo Stella and Ali Caner TÃ¼rkmen and Yuyang Wang},
  title   = {GluonTS: Probabilistic and Neural Time Series Modeling in Python},
  journal = {Journal of Machine Learning Research},
  year    = {2020},
  volume  = {21},
  number  = {116},
  pages   = {1--6},
  url     = {http://jmlr.org/papers/v21/19-820.html}
}

@article{darts,
  author  = {Julien Herzen and Francesco Lassig and Samuele Giuliano Piazzetta and Thomas Neuer and Lao Tafti and Guillaume Raille and Tomas Van Pottelbergh and Marek Pasieka and Andrzej Skrodzki and Nicolas Huguenin and Maxime Dumonal and Jan Kocisz and Dennis Bader and Fradarick Gusset and Mounir Benheddi and Camila Williamson and Michal Kosinski and Matej Petrik and Ga Grosch},
  title   = {Darts: User-Friendly Modern Machine Learning for Time Series},
  journal = {Journal of Machine Learning Research},
  year    = {2022},
  volume  = {23},
  number  = {124},
  pages   = {1--6},
  url     = {http://jmlr.org/papers/v23/21-1177.html}
}

@misc{sktimets,
  doi = {10.48550/ARXIV.1909.07872},
  url = {https://arxiv.org/abs/1909.07872},
  author = {L\"{o}ning,  Markus and Bagnall,  Anthony and Ganesh,  Sajaysurya and Kazakov,  Viktor and Lines,  Jason and Király,  Franz J.},
  title = {sktime: A Unified Interface for Machine Learning with Time Series},
  publisher = {arXiv},
  year = {2019},
  copyright = {arXiv.org perpetual,  non-exclusive license}
}

@article{Wu_2021, 
    title={Current Time Series Anomaly Detection Benchmarks are Flawed and are Creating the Illusion of Progress}, 
    ISSN={2326-3865}, 
    url={http://dx.doi.org/10.1109/tkde.2021.3112126}, 
    DOI={10.1109/tkde.2021.3112126}, 
    journal={IEEE Transactions on Knowledge and Data Engineering}, 
    publisher={Institute of Electrical and Electronics Engineers (IEEE)}, 
    author={Wu, Renjie and Keogh, Eamonn}, 
    year={2021}, 
    pages={1â€“1} 
}

@inproceedings{Liu_2024, 
    series={NeurIPS 2024}, 
    title={The Elephant in the Room: Towards A Reliable Time-Series Anomaly Detection Benchmark}, 
    url={http://dx.doi.org/10.52202/079017-3437}, 
    DOI={10.52202/079017-3437}, 
    booktitle={Advances in Neural Information Processing Systems 37}, 
    publisher={Neural Information Processing Systems Foundation, Inc. (NeurIPS)}, 
    author={Liu, Qinghua and Paparrizos, John}, 
    year={2024}, 
    pages={108231â€“108261}, 
    collection={NeurIPS 2024} 
}

@article{aeon24jmlr,
  author  = {Matthew Middlehurst and Ali Ismail-Fawaz and Antoine Guillaume and Christopher Holder and David Guijo-Rubio and Guzal Bulatova and Leonidas Tsaprounis and Lukasz Mentel and Martin Walter and Patrick Sch{{\"a}}fer and Anthony Bagnall},
  title   = {aeon: a Python Toolkit for Learning from Time Series},
  journal = {Journal of Machine Learning Research},
  year    = {2024},
  volume  = {25},
  number  = {289},
  pages   = {1--10},
  url     = {http://jmlr.org/papers/v25/23-1444.html}
}

@misc{olivares2022library_neuralforecast,
    author={Kin G. Olivares and
            Cristian Challú and
            Azul Garza and
            Max Mergenthaler Canseco and
            Artur Dubrawski},
    title = {{NeuralForecast}: User friendly state-of-the-art neural forecasting models.},
    year={2022},
    howpublished={{PyCon} Salt Lake City, Utah, US 2022},
    url={https://github.com/Nixtla/neuralforecast}
}

@inproceedings{agtimeseries,
  title={{AutoGluon-TimeSeries}: {AutoML} for Probabilistic Time Series Forecasting},
  author={Shchur, Oleksandr and Turkmen, Caner and Erickson, Nick and Shen, Huibin and Shirkov, Alexander and Hu, Tony and Wang, Yuyang},
  booktitle={International Conference on Automated Machine Learning},
  year={2023}
}

@article{arias2021aircraft,
  title={Aircraft engine run-to-failure dataset under real flight conditions for prognostics and diagnostics},
  author={Arias Chao, Manuel and Kulkarni, Chetan and Goebel, Kai and Fink, Olga},
  journal={Data},
  volume={6},
  number={1},
  pages={5},
  year={2021},
  publisher={MDPI}
}

@techreport{frederick2007user,
  title={User's guide for the commercial modular aero-propulsion system simulation (C-MAPSS)},
  author={Frederick, Dean K and DeCastro, Jonathan A and Litt, Jonathan S},
  year={2007}
}

@article{Granderson2020,
author = "Jessica Granderson and Guanjing Lin and Ari Harding and Piljae Im and Yan Chen",
title = "{Dataset for building fault detection and diagnostics algorithm creation and performance testing}",
year = "2020",
month = "1",
url = "https://figshare.com/articles/dataset/LBNLDataSynthesisInventory_pdf/11752740",
doi = "10.6084/m9.figshare.11752740.v3"
}

@InProceedings{PHME20-GTU,
    author = {Kürşat İnce and Engin Sirkeci and Yakup Genç},
    title = {Remaining Useful Life Prediction for Experimental Filtration System: A Data Challenge},
    booktitle = {Proceedings of the European Conference of the PHM Society 2020},
    month = July,
    year = 2020,
    publisher = {PHM Society},
    editor = {Anibal Bregon and Kamal Medjaher},
    note = {Available at https://phmpapers.org/index.php/phme/article/view/1317}
}

@INPROCEEDINGS{hsf15_helwig,
  author={Helwig, Nikolai and Pignanelli, Eliseo and Schütze, Andreas},
  booktitle={2015 IEEE International Instrumentation and Measurement Technology Conference (I2MTC) Proceedings}, 
  title={Condition monitoring of a complex hydraulic system using multivariate statistics}, 
  year={2015},
  volume={},
  number={},
  pages={210-215},
  doi={10.1109/I2MTC.2015.7151267}}

@article{Wang2024DeepTS,
  title={Deep Time Series Models: A Comprehensive Survey and Benchmark},
  author={Yuxuan Wang and Haixu Wu and Jiaxiang Dong and Yong Liu and Mingsheng Long and Jianmin Wang},
  journal={ArXiv},
  year={2024},
  volume={abs/2407.13278},
  url={https://api.semanticscholar.org/CorpusID:271270683}
}

@article{teubert_generic_2020,
	title = {A {Generic} {Software} {Architecture} for {Prognostics} ({GSAP})},
	volume = {8},
	issn = {2153-2648, 2153-2648},
	url = {https://papers.phmsociety.org/index.php/ijphm/article/view/2618},
	doi = {10.36001/ijphm.2017.v8i2.2618},
	number = {2},
	urldate = {2026-04-30},
	journal = {International Journal of Prognostics and Health Management},
	author = {Teubert, Christopher and J. Daigle, Matthew and Sankararaman, Shankar and Goebel, Kai and Watkins, Jason},
	month = nov,
	year = {2020},
}

@article{bieber_generic_2022,
	title = {A {Generic} {Framework} for {Prognostics} of {Complex} {Systems}},
	volume = {9},
	issn = {2226-4310},
	url = {https://www.mdpi.com/2226-4310/9/12/839},
	doi = {10.3390/aerospace9120839},
	language = {en},
	number = {12},
	urldate = {2026-04-30},
	journal = {Aerospace},
	author = {Bieber, Marie and Verhagen, Wim J. C.},
	month = dec,
	year = {2022},
	pages = {839},
}

@Misc{Yadan2019Hydra,
  author =       {Omry Yadan},
  title =        {Hydra - A framework for elegantly configuring complex applications},
  howpublished = {Github},
  year =         {2019},
  url =          {https://github.com/facebookresearch/hydra}
}

\newpage
\appendix

\section*{Appendix Contents}
\addcontentsline{toc}{section}{Appendix Contents}
\vspace{1em}
\begingroup
\makeatletter
\setlength{\parindent}{0pt}
\setlength{\parskip}{0pt}
\footnotesize
\hypersetup{linkcolor=black}

\newcommand{\appcontentsection}[2]{%
  \@dottedtocline{1}{0em}{2.4em}{\hyperref[#1]{\textbf{#2}}}{\pageref{#1}}%
}
\newcommand{\appcontentsubsection}[2]{%
  \@dottedtocline{2}{2.4em}{2.4em}{\hyperref[#1]{\ref*{#1}\hspace{0.6em}#2}}{\pageref{#1}}%
}

\appcontentsection{app:intro_phm}{A.\hspace{0.4em}Introduction to PHM}
\appcontentsubsection{app:intro_phm:context}{PHM in Context}
\appcontentsubsection{app:intro_phm:core_tasks}{Core PHM Tasks}
\appcontentsubsection{app:intro_phm:data_labels}{Data, Labels, and Domain Heterogeneity}
\appcontentsubsection{app:intro_phm:representations_targets}{Representations and task targets}
\appcontentsubsection{app:intro_phm:benchmarking}{Why PHM Is Difficult to Standardize and Benchmark}
\vspace{0.2\baselineskip}

\appcontentsection{app:phm_audit}{B.\hspace{0.4em}Audit of Reproducibility Artifact Availability in PHM Society Conference and IJPHM}
\appcontentsubsection{app:phm_audit:scope}{Scope and corpus construction}
\appcontentsubsection{app:phm_audit:protocol}{Audit protocol}
\appcontentsubsection{app:phm_audit:main_results}{Main results}
\appcontentsubsection{app:phm_audit:nuance}{Nuance beyond strict release}
\appcontentsubsection{app:phm_audit:validation}{Validation and limitations}
\vspace{0.2\baselineskip}

\appcontentsection{app:extended_related_work}{C.\hspace{0.4em}Extended Related Work}
\appcontentsubsection{app:extended_related_work:ml}{Evaluation infrastructures in machine learning}
\appcontentsubsection{app:extended_related_work:tsml}{The time-series ML ecosystem}
\appcontentsubsection{app:extended_related_work:utilities}{PHM data access and prognostic software utilities}
\appcontentsubsection{app:extended_related_work:platforms}{PHM evaluation platforms}
\appcontentsubsection{app:extended_related_work:critiques}{Benchmark-construction critiques}
\appcontentsubsection{app:extended_related_work:methodology}{Methodology, reproducibility, and workflow frameworks}
\vspace{0.2\baselineskip}

\appcontentsection{app:complete_formalization}{D.\hspace{0.4em}Complete Framework Formalization}
\appcontentsubsection{app:tasks_full}{PHM tasks and target definitions}
\appcontentsubsection{app:feature_transform_full}{Feature transformation}
\appcontentsubsection{app:target_transform_full}{Target transformation and alignment}
\appcontentsubsection{app:windowing_full}{Windowing and label alignment}
\appcontentsubsection{app:partitioning_full}{Partitioning and leakage control}
\appcontentsubsection{app:evaluation_full}{Evaluation protocol}
\appcontentsubsection{app:training_objectives}{Model families and fitting conventions}
\appcontentsubsection{app:tabularization_full}{Tabularization and dual representation}
\vspace{0.2\baselineskip}

\appcontentsection{app:library_architecture}{E.\hspace{0.4em}Complete PICID Library Architecture and Extension Interfaces}
\appcontentsubsection{app:lib_datasources}{Datasources}
\appcontentsubsection{app:lib_typed_containers}{Typed data containers}
\appcontentsubsection{app:lib_transforms}{Transform pipeline and preprocessing}
\appcontentsubsection{app:lib_datasets}{Datasets and sequencers}
\appcontentsubsection{app:lib_datamodule}{Datamodule and batching}
\appcontentsubsection{app:lib_wrappers}{Model wrappers}
\appcontentsubsection{app:lib_evaluation}{Evaluation and reporting}
\appcontentsubsection{app:lib_interfaces}{Public extension interfaces}
\vspace{0.2\baselineskip}

\appcontentsection{app:yaml_example}{F.\hspace{0.4em}Transform System: Configuration, Inventory, and Pipeline}
\appcontentsubsection{app:transform_walkthrough}{Transform configuration walkthrough}
\appcontentsubsection{app:transform_inventory}{Transform inventory}
\appcontentsubsection{app:preprocessing_pipeline}{Preprocessing pipeline}
\vspace{0.2\baselineskip}

\appcontentsection{sec:dataset_descriptions}{G.\hspace{0.4em}Dataset Descriptions}
\appcontentsubsection{sec:battery_datasets}{Battery datasets}
\appcontentsubsection{sec:bearing_datasets}{Bearing datasets}
\appcontentsubsection{sec:ncmapss_families}{N-CMAPSS families}
\appcontentsubsection{sec:hsf15_dataset}{Hydraulic diagnostics (HSF15)}
\appcontentsubsection{sec:mzvav_dataset}{MZVAV}
\appcontentsubsection{sec:phme20_dataset}{PHME20}
\vspace{0.2\baselineskip}

\appcontentsection{app:models}{H.\hspace{0.4em}Model inventory}
\vspace{0.2\baselineskip}

\appcontentsection{app:full_experiments}{I.\hspace{0.4em}Full Experimental Results}
\appcontentsubsection{app:exp_setup_full}{Experimental setup}
\appcontentsubsection{app:transform_schemas}{Transformation schemas}
\appcontentsubsection{app:hyperparameter_search}{Hyperparameter search}
\appcontentsubsection{app:reading_guide}{Reading the result tables}
\appcontentsubsection{app:full_diagnostics_results}{Diagnostics results}
\appcontentsubsection{app:full_prognostics_results}{Prognostics results}
\vspace{0.2\baselineskip}

\appcontentsection{app:reproducibility}{J.\hspace{0.4em}Reproducibility}
\appcontentsubsection{app:reproducibility:details}{Details}
\appcontentsubsection{app:reproducibility:experiment_generator}{Experiment generator}
\vspace{0.2\baselineskip}

\appcontentsection{app:data_code_access}{K.\hspace{0.4em}Data and code availability}
\vspace{0.2\baselineskip}

\appcontentsection{sec:ethical_considerations}{L.\hspace{0.4em}Ethical considerations}

\hypersetup{linkcolor=blue}
\makeatother
\endgroup
\newpage

\section{Introduction to PHM}
\label{app:intro_phm}

\subsection{PHM in Context}
\label{app:intro_phm:context}

Maintenance strategies are often described along a progression from \emph{reactive} maintenance (intervene after failure) to \emph{preventive} maintenance (intervene on a fixed schedule), and then to \emph{condition-based maintenance} (CBM), where decisions are driven by observed asset condition rather than elapsed time alone \citep{jardine2006review,lee2014prognostics,zonta2020predictive}. \emph{Predictive maintenance} (PdM) extends CBM by using monitoring data and predictive models to anticipate degradation or failure before functional loss becomes operationally critical.

Within this landscape, \emph{prognostics and health management} (PHM) focuses on converting condition-monitoring signals into health-relevant statements over time \citep{lee2014prognostics,zonta2020predictive}. Concretely, PHM couples (i)~\emph{continuous condition monitoring}---the collection of measurements such as vibration, temperature, pressure, current, or voltage during operation---with (ii)~\emph{analytical inference}, which maps those measurements to outputs such as fault scores, fault labels, or remaining useful life estimates.

\picid{} targets this monitoring-and-inference layer rather than downstream maintenance optimization. In this layer, PHM methods are commonly categorized as physics-model-based, data-driven, or hybrid \citep{lee2014prognostics,li2024small}; \picid{} focuses on the data-driven setting and provides benchmarking infrastructure for systematic comparison under explicit, leakage-safe experimental protocols.

\subsection{Core PHM Tasks}
\label{app:intro_phm:core_tasks}

\begin{figure}[t]
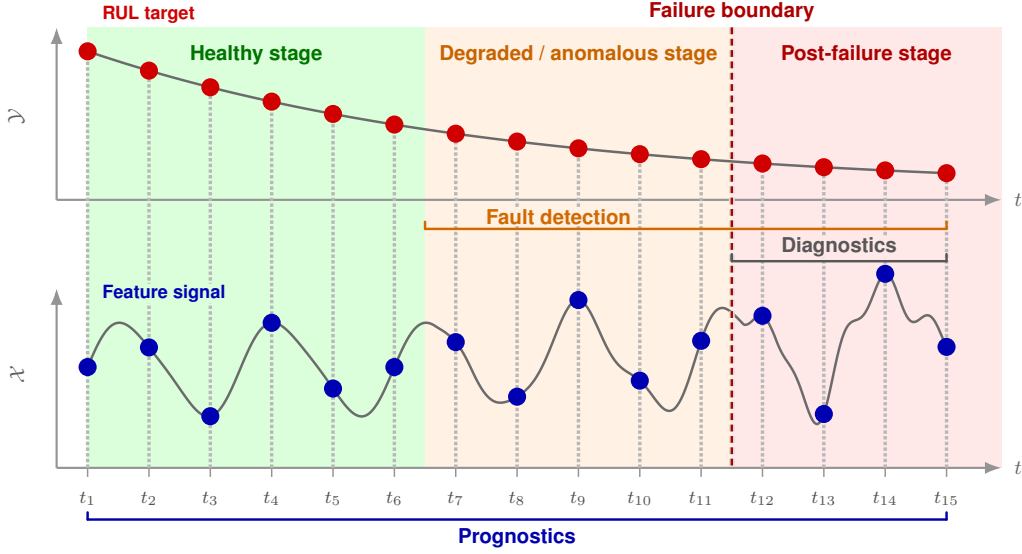

\centering
\input{figures/phm_signal/base_scene.tex}
\input{figures/phm_signal/task_overlays.tex}
\resizebox{0.98\linewidth}{!}{%
\begin{tikzpicture}[font=\sffamily, >=Latex]
\def\PHMOverviewStateLaneYOffset{-0.62cm}%
\def\PHMOverviewTaskAnomalyYOffset{3.7cm}%
\def\PHMOverviewTaskDiagnosticsYOffset{3.2cm}%
\def\PHMOverviewTaskPrognosticsYOffset{-0.8cm}%
\def\PHMOverviewFailureLabelYOffset{0.05cm}%
\PHMConfigureTaskOverlayGeometry{%
  \PHMOverviewStateLaneYOffset%
}{%
  \PHMOverviewTaskAnomalyYOffset%
}{%
  \PHMOverviewTaskDiagnosticsYOffset%
}{%
  \PHMOverviewTaskPrognosticsYOffset%
}{%
  \PHMOverviewFailureLabelYOffset%
}

\PHMActivateTaskOverlayLayers

\DrawPHMSignalBaseScene{phmoverview}
\DrawPHMTaskOverviewOverlay{phmoverview}
\end{tikzpicture}
}
\caption{Three PHM inference tasks on a single monitored unit. \emph{Top panel ($\gY$):} the prognostic target (e.g., remaining useful life) declines as degradation progresses from a healthy stage (green) through a degraded/anomalous stage (yellow) to post-failure (red). \emph{Bottom panel ($\gX$):} the raw feature signal (e.g., vibration or voltage) recorded over the same timeline. Fault detection operates over the degraded stage, flagging departures from nominal behavior; diagnostics classifies the fault type once sufficient post-failure evidence has accumulated; prognostics spans the full trajectory, estimating how much useful life remains at each time step. The three tasks thus share the same underlying sensor data and temporal support but differ in their supervisory targets, applicable time regimes, and output semantics---a heterogeneity that motivates the unified formalization in Section~\ref{sec:framework_formalization}.}
\label{fig:formalization_phm_tasks_overview}
\end{figure}

Broadly interpreted, PHM addresses four practical questions: Is something abnormal? What fault or operating mode is present? How is the asset degrading and how long can it continue operating? What action should be taken in response? These questions correspond to \emph{fault detection}, \emph{diagnostics}, \emph{prognostics}, and \emph{health management}, respectively \citep{lee2014prognostics,Su2024}. Figure~\ref{fig:formalization_phm_tasks_overview} illustrates how the first three tasks attach to different temporal regimes of the same monitored trajectory.

\paragraph{Fault Detection.}
Fault detection addresses the question of whether the observed behavior departs from an expected nominal pattern. Its output is commonly a fault score or a binary normal-versus-abnormal label, and it is often studied in one-class, unsupervised, or weakly supervised settings because abnormal examples are rare or only partially labeled. Intuitively, if vibration or process measurements begin to deviate from the patterns seen during healthy operation, fault detection is the task that flags that departure before a specific fault identity is necessarily known.

\paragraph{Diagnostics.}
Diagnostics addresses the question of what fault, degradation mode, or operational state is present once the asset's operational data---such as vibration signatures, temperature profiles, or electrical signals---are available for analysis. The typical output is a discrete class label, so diagnostics is often treated as a classification problem in which models distinguish among conditions such as normal operation, inner-race bearing damage, valve malfunction, or other fault categories. In practical terms, diagnostics does not merely ask whether something is wrong; it asks what kind of problem is most likely responsible for the observed behavior.

\paragraph{Prognostics.}
Prognostics addresses the question of how degradation is evolving and how long the asset can continue operating before a specified failure criterion is reached. Its output is usually a continuous quantity, such as \emph{remaining useful life} (RUL) or a scalar health indicator, and for that reason prognostics is typically formulated as a regression problem rather than a classification problem. A battery example is especially intuitive: from voltage, current, and temperature trajectories observed over repeated charge-discharge cycles, the goal is to estimate a continuous measure of future life or degradation severity rather than to assign a single fault class \citep{severson2019data,Su2024}.

\paragraph{Health Management.}
Health management addresses the decision question that follows from the inference tasks above: given current health evidence, what maintenance or operational action should be taken? Its outputs may include maintenance recommendations, inspection priorities, scheduling decisions, or risk-aware operating constraints. In this sense, health management consumes the outputs of fault detection, diagnostics, and prognostics together with cost, safety, and mission considerations in order to support action rather than merely description \citep{lee2014prognostics,zonta2020predictive}.

\picid{} operationalizes fault detection, diagnostics, and prognostics as executable benchmark tasks, while treating health management as the downstream decision layer that consumes those outputs in application-specific ways.

\subsection{Data, Labels, and Domain Heterogeneity}
\label{app:intro_phm:data_labels}

PHM data are usually organized as trajectories collected from monitored units over time. A unit may correspond to a bearing, battery cell, engine, hydraulic rig, or another engineered asset, and each unit can generate multivariate measurements under different operating conditions, duty cycles, and degradation paths \citep{lee2014prognostics,hagmeyer2021creation,mauthe2025overview}. As a result, PHM datasets rarely resemble a homogeneous table of independent samples: sequence lengths often differ across units, labels may be available only for selected intervals or terminal events, and some trajectories are truncated before failure, censored, or only partially observed.

The same PHM task can look very different across domains because both the sensing modality and the labeling/target structure change with the application. For example, rotating-machinery diagnostics may use high-rate vibration segments with discrete fault classes, while battery prognostics may use long charge--discharge histories with a continuous degradation target inferred from operational data \citep{helwig2015condition,severson2019data,mauthe2025overview}. In addition, run-to-failure data are hard to collect in many industrial settings, so public datasets are limited and heavily reused \citep{hagmeyer2021creation,mauthe2025overview}. This reuse makes ``small'' protocol choices (windowing, splitting, target construction) disproportionately important for the performance numbers a paper reports \citep{li2024small}.

This variation in sensing modality, label structure, sequence length, and degradation observability is one reason unifying PHM tasks across domains is difficult and one reason \picid{}'s unification claim is non-trivial. Representative datasets used in this paper are summarized in Appendix~\ref{sec:dataset_descriptions}.

\subsection{Representations and task targets}
\label{app:intro_phm:representations_targets}

\paragraph{Representations (features).}
Representations in PHM are not one-size-fits-all: depending on sensing modality and task, models may consume raw waveforms, engineered descriptors, or learned embeddings.
Some approaches operate directly on raw waveforms or process traces, whereas others construct engineered descriptors from the time, frequency, or time-frequency domains in order to expose degradation-relevant structure such as impulsiveness, spectral redistribution, or operating-state variation \citep{sim2020tutorial}.
For example, bearing diagnostics often uses spectral summaries or time--frequency features (e.g., STFT bands, envelope spectra, kurtosis/RMS over short windows), while battery prognostics may rely on cycle-level descriptors such as capacity fade, internal resistance proxies, or summaries of voltage--current curves.
More recently, learned representations based on autoencoders, convolutional networks, and recurrent architectures have been used to extract latent features from high-dimensional monitoring data without requiring every informative pattern to be specified manually \citep{zhao2019deep}.
In the protocol notation of Section~\ref{sec:framework_formalization}, this representation step is the feature pipeline $\mathcal{G}$ acting on raw unit inputs: $\gZ=\mathcal{G}(\gX;\Psi)$. When $\mathcal{G}$ aggregates or re-indexes time (e.g., produces one feature vector per window), the benchmark must attach a single supervisory label to each produced feature index; \picid{} treats this as an explicit alignment choice (Section~\ref{sec:framework_formalization}).

\paragraph{Targets by task.}
PHM targets differ primarily in \emph{what is being predicted} and \emph{at what temporal granularity} labels are available. In \picid{}, these are unified as a raw-time target $y(t)$ together with an aligned target sequence on the transformed grid (Section~\ref{sec:framework_formalization}).
For \emph{fault detection}, supervision may be a binary nominal/abnormal label or a fault score, and labels are often weak (e.g., only at coarse segments or only for a subset of runs).
For \emph{diagnostics}, the target is a discrete condition label (fault type, operating mode, or component state). A common practical issue is that the label may be provided per segment rather than per timestamp, so the benchmark must define how segment labels are assigned to windows.
For \emph{prognostics}, the target is continuous and typically framed as regression. The most direct target is remaining useful life (RUL), but PHM practice also uses \emph{health index} (HI) targets when exact failure-time labels are unavailable, noisy, or too indirect to supervise reliably at every time step \citep{lee2014prognostics,sim2020tutorial}. A health index is a scalar designed to summarize degradation progression over time, often with a monotonic trend, e.g., decreasing from healthy toward failed.
In all three tasks, the key detail strongly affecting reported performance is how labels are aligned to the inputs: e.g., whether a window inherits the label at its end time, a majority label over its support, or an offset label that predicts ahead. \picid{} makes this alignment explicit in the protocol so that comparisons across models and datasets remain meaningful.

\subsection{Why PHM Is Difficult to Standardize and Benchmark}
\label{app:intro_phm:benchmarking}

PHM is hard to benchmark fairly because performance depends on more than the model family. It also depends on the evaluation protocol: how units are split into train/validation/test, how trajectories are windowed, how continuous targets (RUL/health index) are constructed, how sparse labels are aligned to windows, and whether any learned preprocessing state (e.g., normalizers, imputers, feature extractors) is estimated using only training data and then frozen for validation/test \citep{Su2024,li2024small}. These details matter because PHM spans multiple tasks and domains whose data contracts are not naturally uniform.

Recent benchmark critiques make this sensitivity concrete. For example, \citet{hendriks2022towards} show that apparently standard train-test protocols on the CWRU bearing dataset can still reuse the same physical bearings across splits, thereby overstating generalization. Similarly, a recent overview of 98 publicly available degradation datasets finds that many lack a standardized task assignment, leaving users to decide independently whether a given dataset targets diagnosis, prognosis, or fault detection \citep{mauthe2025overview}---a source of implicit variation that further undermines cross-study comparability. \picid{} is motivated by this broader methodological problem. Its formalization therefore makes split semantics, target alignment, deterministic dataset construction, and leakage-safe evaluation explicit protocol objects rather than hidden implementation details, so that fault detection, diagnostics, and prognostics can be compared under the same auditable benchmark contract across heterogeneous PHM domains.

\section{Audit of reproducibility artifact availability in PHM Society Conference and IJPHM}
\label{app:phm_audit}

This appendix reports the audit underlying the main-text claim that direct, paper-level code and data release is rare in PHM. To contextualize the reproducibility challenges discussed in the main text, we audited paper-level code and data release across two venues: the PHM Society Conference and the International Journal of Prognostics and Health Management (IJPHM). Concretely, we ask whether a reader can find a \emph{directly accessible} artifact (e.g., a resolvable repository link or downloadable dataset) that plausibly corresponds to the paper's experimental pipeline, as opposed to a non-verifiable statement such as ``available online'' or ``upon request.''

\subsection{Scope and corpus construction}
\label{app:phm_audit:scope}

The audit covers the PHM Society Conference and IJPHM from 2022 to 2025. To keep the corpus focused on research-facing publications, inclusion rules are as follows: (i)~PHM Society Conference technical research papers for 2022; (ii)~technical research and industry experience papers from 2023 onward; (iii)~IJPHM technical papers, excluding technical briefs. Final labels are assigned only from full-paper text, since availability evidence often appears in methods sections, implementation notes, footnotes, acknowledgments, captions, or appendices rather than in abstracts alone. Under these criteria, the pooled corpus contains 329 papers: 205 from the PHM Society Conference and 124 from IJPHM, with full-text coverage for all included papers.

\subsection{Audit protocol}
\label{app:phm_audit:protocol}

Code and data are labeled independently on the five-point \texttt{A1--A5} scale summarized in Table~\ref{tab:audit_rubric}. The audit is intentionally conservative: its main result tracks direct paper-level release, not general ecosystem openness. In particular, we do not attempt to execute released code, request private access, or recover artifacts from non-resolving links; labels reflect evidence available from the paper text and any directly referenced public resources.

\begin{table}[H]
\centering
\small
\caption{Artifact-availability rubric used in the audit. Labels are assigned independently for code and for data.}
\label{tab:audit_rubric}
\begin{tabular}{@{}p{1.2cm}p{4.2cm}p{7.8cm}@{}}
\toprule
\textbf{Label} & \textbf{Meaning} & \textbf{Typical evidence} \\
\midrule
\texttt{A1} & Directly accessible, author-controlled release with the paper & ``our code is available at~\ldots''; ``we release the dataset at~\ldots'' \\
\texttt{A2} & Public-access signal without a verifiable paper-owned release & ``the software is publicly available online''; claim that a public dataset exists without a paper-owned release link \\
\texttt{A3} & Restricted access & ``available upon request'' \\
\texttt{A4} & Explicit non-availability & ``the data cannot be shared because of confidentiality'' \\
\texttt{A5} & No credible sharing statement & --- \\
\bottomrule
\end{tabular}
\end{table}%

Operationally, the audit reads extracted full-paper text, looks for URLs and availability statements, applies deterministic labeling rules for code and data, and records brief notes supporting each decision. We also track ``public-data context'' separately from strict paper-level release: reusing a public benchmark is informative, but it does not recover the paper-specific pipeline (preprocessing, split construction, and evaluation protocol).

\subsection{Main results}
\label{app:phm_audit:main_results}

The pooled strict result is highly concentrated in non-release outcomes. Under the \texttt{A1} criterion, 321 of 329 papers (97.6\%) provide neither public code nor public data, 6 papers (1.8\%) provide only code, no paper provides only data, and only 2 papers (0.6\%) provide both; in aggregate, only 8 of 329 papers (2.4\%) release any directly accessible artifact. The dominant failure mode is therefore not partial openness but the absence of direct paper-level artifact release, summarized in Table~\ref{tab:audit_strict_results} and Figure~\ref{fig:audit_overview}(a). Equivalently, by release channel, 8 papers (2.4\%) release code and only 2 (0.6\%) release data; we emphasize the code-release figure in the main text because code encodes the per-paper execution protocol---windowing, normalization, split construction, and evaluation---so even reuse of a public benchmark does not substitute for paper-level code when assessing direct reproducibility.

\begin{wraptable}{r}{0.52\linewidth}
\vspace{-0.6\baselineskip}
\centering
\small
\caption{Strict public-availability outcomes under \texttt{A1}.}
\label{tab:audit_strict_results}
\begin{tabular}{lrrr}
\toprule
 & Pooled & PHM Society & IJPHM \\
\midrule
Papers & 329 & 205 & 124 \\
Both & 2 & 2 & 0 \\
Only code & 6 & 3 & 3 \\
Only data & 0 & 0 & 0 \\
Neither & 321 & 200 & 121 \\
\bottomrule
\end{tabular}

\vspace{-0.8\baselineskip}
\end{wraptable}

\subsection{Nuance beyond strict release}
\label{app:phm_audit:nuance}

Two forms of nuance complicate the headline. First, the full label distribution grades the sharing statements that do exist beyond strict \texttt{A1}; second, broader ecosystem signals point to shared empirical resources that papers rely on but do not themselves release.

\paragraph{Full label distribution.}
The strict result does not imply that PHM research never touches shared empirical resources. The full label distribution shows a different nuance: direct release remains rare, but broader openness signals exist alongside it. Across the pooled corpus, code labels are distributed as \texttt{A1=8, A2=14, A3=2, A4=0, A5=305}, and data labels as \texttt{A1=2, A2=3, A3=2, A4=5, A5=317}. In other words, the scarcity of direct release is accompanied by a strong concentration in \texttt{A5} (no credible sharing statement), especially for data.

\begin{wraptable}{r}{0.52\linewidth}
\vspace{-0.6\baselineskip}
\centering
\scriptsize
\caption{Full label distribution (compact).}
\label{tab:audit_label_distribution}
\begin{tabular}{@{}p{1.25cm}p{2.55cm}p{2.55cm}@{}}
\toprule
\textbf{Scope} & \textbf{Code (\texttt{A1--A5})} & \textbf{Data (\texttt{A1--A5})} \\
\midrule
Pooled & 8, 14, 2, 0, 305 & 2, 3, 2, 5, 317 \\
PHM Soc. & 5, 9, 1, 0, 190 & 2, 2, 2, 3, 196 \\
IJPHM & 3, 5, 1, 0, 115 & 0, 1, 0, 2, 121 \\
\bottomrule
\end{tabular}

\vspace{-0.8\baselineskip}
\end{wraptable}

\begin{figure}[H]
\centering
\begin{minipage}[t]{0.48\linewidth}
\centering
\includegraphics[width=\linewidth]{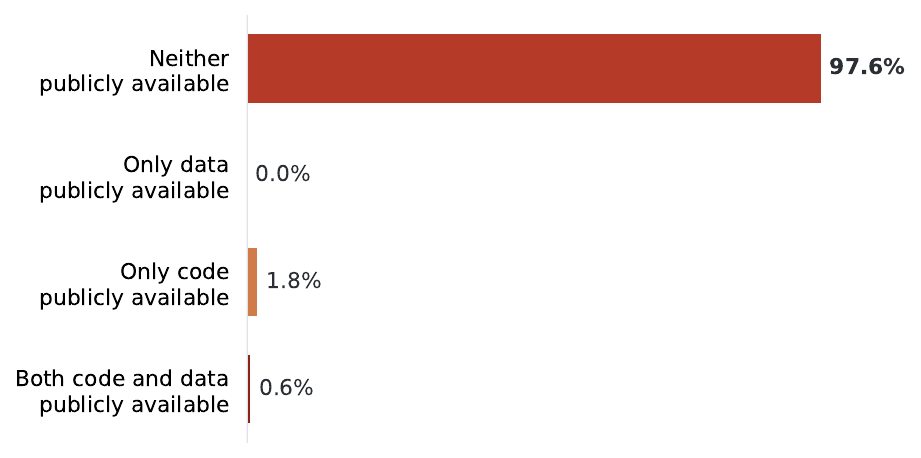}

{\small (a) Strict public-availability outcomes.}
\end{minipage}\hfill
\begin{minipage}[t]{0.48\linewidth}
\centering
\includegraphics[width=\linewidth]{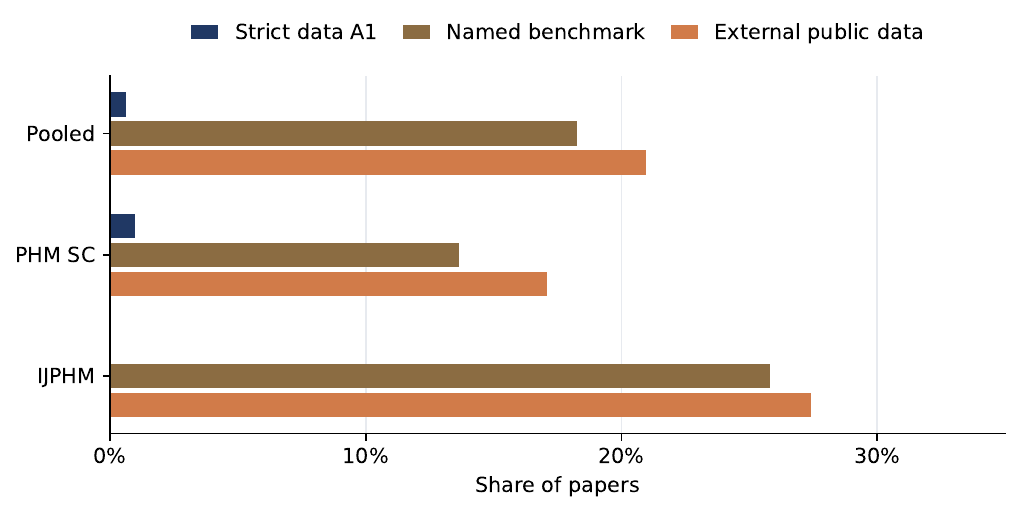}

{\small (b) Public-data context signals.}
\end{minipage}
\caption{Overview of the audit results. Panel~(a) summarizes the pooled strict public-availability outcomes across PHM Society Conference and IJPHM papers from 2022 to 2025, where ``publicly available'' denotes strict \texttt{A1}: a directly accessible, author-controlled code or data artifact released with the paper. Panel~(b) shows broader public-data context signals, tracked separately from strict \texttt{A1} release because benchmark reuse or external public-data access does not, by itself, recover the paper-specific experimental pipeline.}
\label{fig:audit_overview}
\end{figure}

\paragraph{Ecosystem-level public-data signals.}
Broader public-data context is also more common than direct author-controlled release. Across the pooled corpus, 69 papers (21.0\%) show at least one public-data context signal (benchmark datasets, external public repositories, or other explicit signals outside strict \texttt{A1}), of which 60 (18.2\%) specifically reuse named benchmarks. These signals are important because they show that the field does use shared empirical resources. They do not, however, substitute for the headline \texttt{A1} criterion, since benchmark access alone does not recover a paper's own preprocessing, split definitions, implementation details, or evaluation pipeline; Figure~\ref{fig:audit_overview}(b) visualizes these signals.

\subsection{Validation and limitations}
\label{app:phm_audit:validation}

The analyzer was iteratively refined through targeted spot checks and secondary LLM-assisted inspection of sampled papers. This auxiliary pass was used only as an error-discovery aid for revising the deterministic rules and rerunning the corpus; it was not a formal gold-standard human annotation study, and we do not present it as an accuracy benchmark. The final outputs reported here should therefore be read as results of a deterministic audit whose rules were stress-tested and corrected against difficult cases, not as simple keyword counts over titles and abstracts.

\section{Extended related work}
\label{app:extended_related_work}

This appendix expands the discussion in Section~\ref{sec:related_work} by providing additional context on the evaluation-infrastructure landscape surrounding \picid. The subsections mirror the clusters in the main-text related-work section and deepen the comparative discussion. The aim is threefold: to make the comparison with each group auditable, to record the specific design commitments of representative tools rather than only our summary of them, and to justify the positioning claims made in the main text.

\subsection{Evaluation infrastructures in machine learning}
\label{app:extended_related_work:ml}

Several mature machine-learning subfields show that shared evaluation infrastructure can turn individual studies into cumulative research programs. In natural language processing, the Hugging Face \textit{Datasets} library exposes hundreds of benchmarks behind a uniform API with versioning, schema metadata, and deterministic streaming \citep{lhoest-etal-2021-datasets}, reducing the marginal cost of cross-dataset evaluation while making dataset provenance explicit. In graph learning, PyTorch Geometric provides scalable message-passing abstractions, standardized dataset containers, and reproducible experimental pipelines that have become a de facto platform for representation-learning research \citep{fey2019pyg}. In reinforcement learning, the combination of environment APIs such as OpenAI Gym and trusted algorithm implementations such as Stable-Baselines3 has enabled controlled comparison across algorithm families \citep{stable-baselines3}.

Two lessons generalize across these ecosystems. First, evaluation contracts are most useful when they are \emph{model-agnostic}: any method that consumes the prescribed inputs and produces valid outputs can be compared under identical conditions. Second, the contract must be \emph{executable}: the greatest value comes when the same artifact that defines the evaluation also runs it. \picid{} adopts both principles for PHM. The formalization in Section~\ref{sec:framework_formalization} is model-agnostic, and the software realization in Section~\ref{sec:picid_library} executes that formalization directly rather than merely describing it.

\subsection{The time-series ML ecosystem}
\label{app:extended_related_work:tsml}

Time-series machine learning provides the closest analogue outside PHM to the type of infrastructure that \picid{} targets.

\paragraph{sktime and aeon.}
sktime \citep{sktimets} exposes unified estimator APIs for classification, regression, forecasting, and fault detection over time series, together with standardized split semantics, backtesting utilities, and reproducible evaluators. Its community fork aeon \citep{aeon24jmlr} continues this line with greater emphasis on performance, modernized dependencies, and expanded distance-based and deep-learning estimators. Both projects show that model-agnostic evaluation contracts can span broad time-series task families.

\paragraph{Forecasting-oriented toolkits.}
GluonTS \citep{gluonts} provides probabilistic forecasting models, calibrated evaluators, and large-scale dataset bindings for industrial-scale experimentation. Darts \citep{darts} unifies classical statistical and deep forecasting models behind a consistent interface with backtesting, probabilistic forecasts, and hierarchical reconciliation utilities. NeuralForecast \citep{olivares2022library_neuralforecast} focuses on deep forecasting architectures with reproducible hyperparameter recipes. AutoGluon--TimeSeries \citep{agtimeseries} represents the AutoML direction in time-series ML by automating model selection, ensembling, and retraining under standardized protocols.

\paragraph{Why generic time-series APIs are not sufficient for PHM.}
PHM imposes invariants that generic time-series APIs do not encode:
\begin{itemize}
    \item \textbf{Unit-level semantics.} PHM evaluations are defined over \emph{units} (e.g., turbofans, cells, bearings, tools) with run-to-failure or degradation semantics, rather than over isolated series. Splits must respect either unit boundaries (inter-unit) or a unit's own timeline (intra-unit), depending on the intended notion of generalization.
    \item \textbf{PHM-specific targets.} Remaining useful life, monotone health indicators, and fault or mode labels require domain-aware target transformations, such as clipping, piecewise-linear shaping, or monotonicity constraints, whose parameters must be estimated from training units only.
    \item \textbf{Transform-induced grid changes.} Feature pipelines in PHM frequently alter the temporal grid through STFTs, wavelets, feature windows, or downsampling. This creates a target-alignment problem that must be specified explicitly; generic APIs typically assume a fixed grid or leave alignment entirely to the user.
    \item \textbf{Unit-aware evaluation.} Prognostics metrics are often defined at the unit or trajectory level (e.g., last-window RUL per engine, per-unit PHM score), not as averages over i.i.d.\ samples. The evaluator must therefore support per-unit aggregation.
\end{itemize}

Relative to existing PHM toolchains, \picid{} is most comparable to the broader time-series ML ecosystem, but it is PHM-specific by construction: the protocol and software explicitly encode unit semantics, leakage-safe fitting, and target alignment (Sections~\ref{sec:framework_formalization} and~\ref{sec:picid_library}). For brevity, we group sktime, aeon, GluonTS, Darts, NeuralForecast, and AutoGluon--TimeSeries into a single ``TS-ML ecosystem'' cluster: these toolkits are strong on general-purpose experimentation, but they do not enforce PHM-specific invariants by default.

General-purpose time-series toolkits show how far executable, model-agnostic evaluation contracts can scale, but they do not encode PHM's unit semantics, target construction, and alignment constraints that \picid{} treats as protocol-level invariants.

\subsection{PHM data access and prognostic software utilities}
\label{app:extended_related_work:utilities}

A first PHM-specific cluster focuses on making datasets easier to access and on providing reusable prognostic software.

\paragraph{Data access and dataset utilities.}
PyPHM \citep{von2022computational} exposes common PHM datasets behind a class hierarchy with basic windowing, sampling, and preprocessing helpers. PHMD \citep{SOLISMARTIN2025102039} provides a larger curated catalog with standardized formats and metadata, emphasizing discoverability and loading. The collaborative prognostics data library of \citet{sikorska2016collaborative} focuses on secure sharing and metadata stewardship for reusable industrial datasets, which is particularly relevant in deployment settings where raw field data cannot leave the operator.

\paragraph{Prognostic software stacks.}
ProgPy \citep{teubert2023progpy} provides a substantially richer software layer for prognostics, including physics-based and data-driven models, state estimation, prognostic reasoning, and deployment-oriented evaluation hooks. Among the systems surveyed here, it is the most capable reusable stack for prognostic modeling and simulation.

These utilities complement \picid{} rather than replacing it: they make PHM data easier to access and help build prognostic applications, but they do not define (or enforce) a benchmark contract. In particular, they typically leave key evaluation degrees of freedom to the user (windowing, train-only preprocessing state, target alignment when transforms change the grid, and unit-aware aggregation). \picid{} sits one layer later: it can reuse their data abstractions while adding an executable evaluation contract so that the formal protocol and the code that runs it remain the same artifact.

\subsection{PHM evaluation platforms}
\label{app:extended_related_work:platforms}

A second cluster contributes evaluation platforms or platform-like comparative studies for narrower PHM settings.

\paragraph{PHM Society challenges and challenge analyses.}
The retrospective analysis of PHM Society challenges by \citet{jia2018review} helped establish shared public PHM tasks early in the field and documented the methodological choices associated with each challenge. These challenges remain among the most reused task definitions in PHM. At the same time, each challenge is still tied to a specific task and dataset, with its own evaluation script and surrounding infrastructure, so cumulative comparison across challenges continues to depend on user-maintained plumbing.

\paragraph{Cross-domain fault-diagnosis studies.}
\citet{zhao2020deep} released an open-source benchmark of deep-learning diagnosis algorithms for rotating machinery, with unified code, preprocessing, and evaluation across multiple datasets. \citet{zhao2024domain} extended this line to cross-domain fault diagnosis under distribution shift, combining a survey with a comparative study and released code. Both are important within their scope, but both remain vibration- and rotating-machinery-centric and therefore do not generalize naturally to battery degradation, hydraulic health, or other non-vibration PHM settings.

\paragraph{PHM-Vibench.}
The closest direct comparator to \picid{} is PHM-Vibench \citep{Li2026}, a factory-style framework for vibration-based PHM. PHM-Vibench introduces the Unified PHM Problem (UPHMP) formalism,
\[
\omega = (P, D, T, M, L, \Pi, E) \in \mathbb{P}\times\mathbb{D}\times\mathbb{T}\times\mathbb{M}\times\mathbb{L}\times\Pi\times\mathbb{E},
\]
whose components correspond to the domain-knowledge, data, task, model, loss, protocol, and evaluation spaces. Around this meta-setting, the framework instantiates Data, Model, Task, and Trainer ``factories'' from a single configuration file and integrates both leakage constraints of the form $\sigma_{\mathrm{train}}\cap\sigma_{\mathrm{test}}=\emptyset$ and temporal constraints for sliding windows of the form $\tau\geq\tau_0$ within each split. PHM-Vibench supports classification, fault detection, and RUL prediction, and emphasizes foundation-model adaptation through pre-trained backbones, few-shot learning, and lightweight fine-tuning, with reported demonstrations on bearing datasets such as CWRU, Ottawa, THU, Ottawa-2019, and HUST.

\paragraph{How \picid{} differs from PHM-Vibench.}
PHM-Vibench and \picid{} share several important design commitments: declarative meta-setting specification, protocol isolation as a structural invariant, and modular, replayable configurations. They differ, however, in three respects that make them complementary rather than overlapping.
\begin{enumerate}
    \item \emph{Scope.} PHM-Vibench is explicitly vibration-centered, and its reported demonstrations are limited to bearing datasets. \picid{}, by contrast, is constructed to span non-vibration domains such as turbofan degradation, battery capacity fade, semiconductor chemical--mechanical polishing, and hydraulic systems in addition to bearings. This difference is central to our positioning.
    \item \emph{Formal granularity.} UPHMP captures experiments at the meta-setting level, but does not explicitly treat the target-alignment problem that arises when feature extraction changes the temporal grid---for example, when an STFT or wavelet transform replaces raw time steps with time--frequency frames. \picid{} makes this alignment a first-class object through the auxiliary support map $a(\cdot)$ and the alignment operator $\mathcal{A}$ in the $(\mathcal{G},\widetilde{\mathcal{H}},\mathcal{S})$ triple (Section~\ref{subsec:construction}), so that the protocol records how labels are sampled or aggregated whenever the feature grid changes; see Appendix~\ref{app:yaml_example}, Section~\ref{app:transform_inventory} (STFTTransform) for a concrete transform example.
    \item \emph{Model orientation.} PHM-Vibench is optimized for factory-style experimentation centered on foundation models, with reported benchmarks built around TimesNet, FNO, PatchTST, and DLinear under zero-shot, few-shot, and fine-tuning regimes. \picid{} is model-class agnostic: its evaluation contract is designed so that classical, statistical, deep-learning, and foundation-model approaches can all be compared under identical conditions (Section~\ref{subsec:extensibility}).
\end{enumerate}

The resulting difference is structural rather than cosmetic. PHM-Vibench is a specialized platform optimized for vibration-based foundation-model research, whereas \picid{} is a general, cross-domain evaluation infrastructure in which leakage-safe evaluation and target alignment are formalized as part of the protocol itself.

Taken together, these platforms provide valuable baselines, but they are typically specialized to narrower settings (e.g., a single modality or challenge protocol). \picid{} instead targets cross-task, cross-domain reuse by making two fragile degrees of freedom explicit and executable: (i) leakage-safe evaluation (train-only fitting, split invariants) and (ii) target alignment when preprocessing changes the effective temporal grid.

\subsection{Benchmark-construction critiques}
\label{app:extended_related_work:critiques}

A third cluster shows that benchmark \emph{construction} can dominate apparent progress. \citet{hendriks2022towards} demonstrate that common train/test protocols on CWRU reuse the same physical bearings across splits, inflating reported performance and obscuring whether models are learning transferable diagnostic representations or bearing-specific signatures. \citet{matania2024test} show that segment- or record-level splitting of bearing vibration data yields severely over-optimistic results relative to group-aware protocols, and argue that leakage in bearing diagnosis is structural rather than accidental. In general time-series fault detection, \citet{Wu_2021} and \citet{Liu_2024} document widespread triviality and evaluation artifacts in widely used benchmarks, showing that superficial changes to splitting or scoring conventions can alter rankings without corresponding modeling gains.

These studies motivate \picid{}'s design choice to enforce leakage safety in software, not only in documentation. In \picid{}, training-only estimation of fitted parameters (feature state $\Psi$, target state $\Phi$, and any statistics used by the alignment operator $\mathcal{A}$) is built into the execution path, together with explicit inter-unit vs.\ intra-unit splitting and unit-aware aggregation. Concretely, users cannot accidentally fit a normalizer on test data or construct windows that cross unit boundaries: those failure modes are ruled out by the protocol contract.

\subsection{Methodology, reproducibility, and workflow frameworks}
\label{app:extended_related_work:methodology}

A fourth cluster argues that PHM needs stronger methodological standardization, but does not provide a reusable implementation. For example, \citet{uckun2008standardizing} call for shared conventions for problem definition and cross-study comparison, and \citet{astfalck2016modelling} emphasize that reproducibility depends on documenting data and validation choices (not only the final algorithm). This maps directly onto \picid{}'s protocol view: feature parameters $\Psi$, target parameters $\Phi$, and any alignment statistics are treated as explicit objects that must be fit on training data and then frozen. Reviews such as \citet{Su2024} similarly note the absence of unified guidelines, while deployment-oriented pipeline work (e.g., SBA Tools \citep{bishay2023design}) focuses on operational workflow rather than public evaluation standardization.

In parallel, the broader ML community has moved toward AutoML-style tooling that packages model selection, tuning, and evaluation into reproducible pipelines; AutoGluon--TimeSeries \citep{agtimeseries} is a time-series example. This direction complements \picid{}: AutoML systems work best when an evaluation contract is fixed, but they do not by themselves enforce PHM-specific invariants such as unit semantics, target alignment, and protocol isolation.

Overall, this literature clarifies what rigorous PHM evaluation should require, but it stops short of providing a reusable implementation that enforces those requirements. \picid{} targets that gap: it makes leakage safety, protocol isolation, transparent preprocessing, and explicit target alignment executable through the formalization (Section~\ref{sec:framework_formalization}) and the software realization (Section~\ref{sec:picid_library}).

\section{Complete framework formalization}
\label{app:complete_formalization}

This appendix provides a self-contained, comprehensive reference for the formal specification introduced in Section~\ref{sec:framework_formalization}. It reproduces the core definitions from the main text and extends them with full derivations, left-padding mechanics, multi-step supervision, multi-unit generalization, an illustrative example, training objectives, and the tabularization pathway. Sections~\ref{app:tasks_full}--\ref{app:partitioning_full} form the core specification; Sections~\ref{app:evaluation_full}--\ref{app:tabularization_full} extend it to evaluation, model families, and tabularization. Practitioners and future work can use this appendix as a single reference for the complete protocol.

\noindent\textbf{Pipeline overview.} For each unit, raw time-series inputs $(\gX,\gY)$ are transformed into aligned feature--target sequences via $\mathcal{G}$ and $\widetilde{\mathcal{H}}$, yielding $(\gZ,\gY')$ on a common transformed timeline. The windowing operator $\mathcal{S}$ then converts $(\gZ,\gY')$ into supervised instances $(\vW_k,y_k)$, which are partitioned into train/validation/test under an explicit leakage policy. Finally, models produce predictions $\hat{y}$ from windows, and evaluators aggregate window-level predictions either per window or per unit.

Table~\ref{tab:notation_full} provides the complete notation used throughout the framework specification. The main text now presents only the benchmark invariants needed to follow the contribution; this appendix collects the complete notation, including symbols for temporal-resolution changes, tabularization, and training.

\begin{table}[H]
\centering
\scriptsize
\caption{Complete notation for the framework specification.}
\label{tab:notation_full}

\renewcommand{\arraystretch}{1.10}
\begin{tabular}{@{}ll@{}}
\toprule
\textbf{Symbol} & \textbf{Description} \\
\midrule
\multicolumn{2}{l}{\textit{Raw unit data}} \\
$T$ & Total raw time steps for a unit \\
$t \in \{1,\dots,T\}$ & Raw time index \\
$\vx(t) \in \mathbb{R}^M$ & Raw sensor vector ($M$ channels) \\
$y(t) \in \mathcal{Y}$ & Raw target signal (task-dependent) \\
$\gX = \{\vx(t)\}_{t=1}^T$ & Raw sensor series \\
$\gY = \{y(t)\}_{t=1}^T$ & Raw target series \\
\midrule
\multicolumn{2}{l}{\textit{Feature transformation and target alignment}} \\
$\mathcal{G}(\cdot;\Psi)$ & Feature transformation pipeline with parameters $\Psi$ \\
$\gZ = \{\vz(j)\}_{j=1}^{T'}$ & Transformed feature series, $\vz(j)\in\mathbb{R}^F$ \\
$F$ & Feature dimension after transformation \\
$T'$ & Length of transformed series \\
$w$ & Pipeline-internal history window size (within windowed stages of $\mathcal{G}$; distinct from the benchmark sequencer's $L_{\mathrm{seq}}$) \\
$s_{\mathcal{G}}$ & Signal processing stride (within windowed stages of $\mathcal{G}$) \\
$a(j)$ & Auxiliary raw-time support for transformed index $j$ \\
$\mathcal{H}(\cdot;\Phi)$ & Target transformation pipeline (pointwise on raw time) \\
$\widetilde{\mathcal{H}}(\cdot;\Phi)$ & Target pipeline (transform + temporal alignment) \\
$\mathcal{A}$ & Alignment/aggregation stage inside $\widetilde{\mathcal{H}}$ \\
$\gY' = \{z_y(j)\}_{j=1}^{T'}$ & Aligned target series \\
\midrule
\multicolumn{2}{l}{\textit{Windowing / slicing}} \\
$L_{\mathrm{seq}}$ & History length (transformed steps per input window) \\
$\Delta$ & Stride between consecutive window starts \\
$\rho$ & Warm-start depth (maximum left-padding); $\rho=0$ strict (no padding) \\
$\delta$ & Supervision offset after window end; $\delta=0$ end-of-window \\
$L_{\mathrm{pred}}$ & Supervision segment length (default $1$ for window-level PHM) \\
$L_{\mathrm{req}}$ & Required right-side coverage, $L_{\mathrm{req}} = L_{\mathrm{seq}}+\delta+L_{\mathrm{pred}}$ \\
$\mathcal{K}$ & Admissible set of window start indices \\
$N_{\mathrm{slices}}$ & Number of admissible windows \\
$\vW_k \in \mathbb{R}^{L_{\mathrm{seq}}\times F}$ & Feature window (input) \\
$j_{\mathrm{sup}}(k)$ & Supervision index, $j_{\mathrm{sup}}(k) = k+L_{\mathrm{seq}}-1+\delta$ \\
$y_k$ & Window label, $y_k = z_y(j_{\mathrm{sup}}(k))$ \\
$\mathcal{P}$ & Padding operator for left-padding when $\rho>0$ (defines values for $j\le 0$) \\
$\widetilde{\vz}(j)$ & Padded feature series: $\widetilde{\vz}(j)=\vz(j)$ for $j\ge 1$, and $\widetilde{\vz}(j)=\mathcal{P}(\gZ,j)$ for $j\le 0$ \\
$L_{\mathrm{lbl}}$ & Label-overlap length for multi-step supervision segments \\
\midrule
\multicolumn{2}{l}{\textit{Multi-unit and partitioning}} \\
$U$ & Number of units \\
$\mathcal{D}^{\mathrm{train}},\mathcal{D}^{\mathrm{val}},\mathcal{D}^{\mathrm{test}}$ & Train / validation / test partitions \\
$\mathcal{D}_{\mathrm{global}}$ & Global dataset (union of all per-unit window sets) \\
\midrule
\multicolumn{2}{l}{\textit{Modeling and evaluation}} \\
$f(\cdot)$ & Predictor mapping a window to a prediction \\
$f_{\theta}(\cdot)$ & Sequential model parameterized by $\theta$ \\
$\mathcal{F}(\cdot)$ & Fitting operator for fit-predict models \\
$f_{\xi}(\cdot)$ & Fit-predict model with fitted parameter state $\xi$ \\
$\ell(\cdot,\cdot)$ & Per-sample loss / metric function \\
$\mathcal{L}_{\mathrm{test}}$ & Test performance aggregate \\
\midrule
\multicolumn{2}{l}{\textit{Tabularization and in-context learning}} \\
$\mathcal{T}$ & Tabular adapter (flattening) operator \\
$D_{\mathrm{tab}} = L_{\mathrm{seq}} F$ & Tabular dimension \\
$\vX_k = \mathcal{T}(\vW_k) \in \mathbb{R}^{D_{\mathrm{tab}}}$ & Tabularized window \\
$\mathcal{D}_{\mathrm{win}},\;\mathcal{D}_{\mathrm{tab}}$ & Window / tabular datasets \\
$f_{\phi}(\cdot)$ & Tabular / foundation model parameterized by $\phi$ \\
$\mathcal{C}$ & Context set for in-context learning \\
$\pi$ & Context selection rule \\
\bottomrule
\end{tabular}
\end{table}

\subsection{PHM tasks and target definitions}
\label{app:tasks_full}

The framework formalizes three canonical PHM task families (see Appendix~\ref{app:intro_phm} for task motivation and application domain context), each characterized by the codomain of the target signal $y(t)\in\mathcal{Y}$:

\begin{itemize}
    \item \textbf{Prognostics (RUL / Health Estimation).} $y(t)\in\mathbb{R}_{\ge 0}$ represents Remaining Useful Life, or a monotone health indicator $h(t)\in[0,1]$ that decreases from $1$ (healthy) toward $0$ (failed).

    \item \textbf{Diagnostics (Fault / Mode Classification).} $y(t)\in\{0,1,\dots,K-1\}$ is a discrete label indicating the fault class or operational mode active at time $t$.

    \item \textbf{Fault Detection.} The supervisory signal is either a binary label $y(t)\in\{0,1\}$ for nominal versus faulty operation, or a real-valued fault score $y(t)\in\mathbb{R}$ whose magnitude reflects the degree of abnormality.
\end{itemize}

All tasks are evaluated at the window level (Section~\ref{app:windowing_full}).

\subsection{Feature transformation}
\label{app:feature_transform_full}

The feature transformation operator $\mathcal{G}(\cdot;\Psi)$ is a composition of $P$ deterministic processing stages:
\[
\mathcal{G}(\cdot; \Psi) = g_P(\cdot; \psi_P) \circ g_{P-1}(\cdot; \psi_{P-1}) \circ \dots \circ g_1(\cdot; \psi_1),
\]
where $\Psi=\{\psi_1,\dots,\psi_P\}$ are the fitted parameters. Each stage $g_i$ may be pointwise (e.g., scaling) or windowed/time--frequency (e.g., STFT, wavelets), and may therefore alter the temporal resolution.

The output is a (possibly re-indexed) feature time-series:
\begin{equation}
    \gZ = \mathcal{G}(\gX;\Psi) = \{\vz(j)\}_{j=1}^{T'}, \quad \vz(j)\in\mathbb{R}^{F}.
\end{equation}

Note that windowing can occur both \emph{inside} $\mathcal{G}$ (e.g., STFT frames with internal window size $w$ and stride $s_{\mathcal{G}}$) and later at the benchmark level (Section~\ref{app:windowing_full}) when forming supervised learning windows of length $L_{\mathrm{seq}}$ with stride $\Delta$.%

When $\mathcal{G}$ changes the temporal resolution, the transformed length $T'$ is no longer equal to $T$. In such cases, each transformed index $j\in\{1,\dots,T'\}$ corresponds to a specific \emph{support} in raw time: the raw timestamps (or interval) used to compute $\vz(j)$. We represent this correspondence by an auxiliary raw-time support map
\[
a:\{1,\dots,T'\}\to\mathcal{I}_T,
\]
where $\mathcal{I}_T$ denotes raw timestamps or raw-time intervals. Pointwise stages preserve the current indexing (typically $T'=T$ and $a(j)=\{j\}$), whereas resolution-changing stages induce $T'$ and $a(j)$ through their own windowing or frame-placement rule. For the common case of an interior sliding-window stage with history length $w$ and stride $s_{\mathcal{G}}$, one has
\begin{equation}
    T' = \left\lfloor \frac{T - w}{s_{\mathcal{G}}} \right\rfloor + 1,
    \qquad
    a(j) = \{(j-1)s_{\mathcal{G}} + 1,\,\dots,\,(j-1)s_{\mathcal{G}} + w\}.
\end{equation}
More general time--frequency or boundary-aware stages (e.g., STFT with centered frames or causal-only framing) may induce different transformed lengths and support maps; in those cases, $T'$ and $a(\cdot)$ are part of the stage definition. In this formalization, $a(\cdot)$ is used only to define target alignment (Section~\ref{app:target_transform_full}), i.e., to specify which raw-time targets contribute to the label attached to transformed index $j$.

\textbf{Leakage policy.} All fitted parameters $\Psi$ are estimated exclusively using the training partition, then frozen for validation and test.

\subsection{Target transformation and alignment}
\label{app:target_transform_full}

The target transformation operator $\mathcal{H}(\cdot;\Phi)$ is a composition of $Q$ deterministic stages:
\begin{equation}
    \mathcal{H}(\cdot; \Phi) = h_Q(\cdot; \phi_Q) \circ \dots \circ h_1(\cdot; \phi_1),
\end{equation}
where $\Phi=\{\phi_1,\dots,\phi_Q\}$ are fitted parameters (e.g., clipping thresholds, scaling constants, calibration parameters) derived from the training partition.

The general target pipeline produces one aligned target value per transformed index:
\begin{equation}
    z_y(j) = \widetilde{\mathcal{H}}\big(\gY, a(j); \Phi\big), \quad j=1,\dots,T'.
\end{equation}
A common construction makes the alignment/aggregation stage $\mathcal{A}$ explicit:
\begin{equation}
    z_y(j) = \mathcal{A}\Big(\mathcal{H}\big(\gY;\,\Phi\big),a(j)\Big),
\end{equation}
where $\mathcal{A}$ implements pointwise sampling (when $a(j)$ is a timestamp) or aggregation over an interval (e.g., last/mean/max/majority). More generally, $\widetilde{\mathcal{H}}$ may place alignment at any position within the target pipeline, but in all cases yields one label per transformed index.

The aligned target series is $\gY'=\{z_y(j)\}_{j=1}^{T'}$.

\begin{figure}[t]
\centering
\input{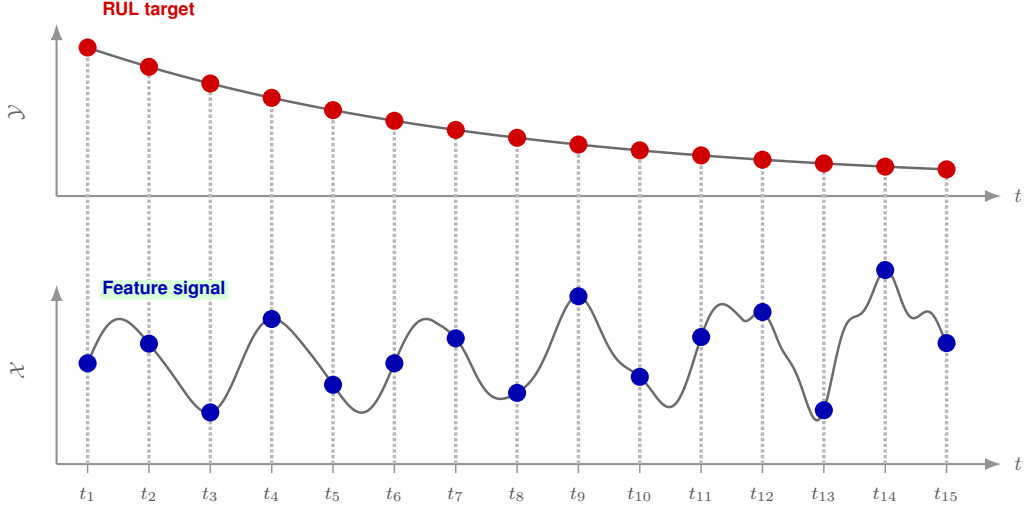}
\resizebox{0.98\linewidth}{!}{%
\begin{tikzpicture}[font=\sffamily, >=Latex]
\DrawPHMSignalBaseScene{phmraw}

\end{tikzpicture}
}
\caption{Toy PHM trajectory pair on a shared temporal support. The top panel shows a monotonically decreasing remaining-useful-life target, while the bottom panel shows a sensor signal with two clean healthy cycles through $t_6$, visible low-level degradation from $t_7$ to $t_{11}$, and a stronger high-degradation regime from $t_{12}$ onward, both sampled at the same $15$ timestamps. This paired support is the object on which the target-side transformation $\mathcal{H}$ and alignment operator $\mathcal{A}$ act before producing one aligned label per transformed index.}
\label{fig:formalization_transform_alignment}
\end{figure}

Figure~\ref{fig:formalization_transform_alignment} shows an illustrative PHM target-feature pair on a shared temporal support, which is the raw object acted on by the target-side transformation and alignment operators.

\textbf{Leakage policy.} All fitted parameters $\Phi$ and any statistics required by $\mathcal{A}$ are estimated exclusively using the training partition and frozen for validation and test.

\subsection{Windowing and label alignment}
\label{app:windowing_full}

The windowing operator $\mathcal{S}$ converts a transformed trajectory into supervised learning instances by slicing two aligned sequences---the feature sequence $\gZ$ and the aligned target sequence $\gY'$---into window-level samples.
\par\noindent
Intuitively, windowing fixes a history length $L_{\mathrm{seq}}$ and enumerates admissible window start indices along the transformed timeline. Each admissible start $k$ yields an input window $\vW_k$ by collecting $L_{\mathrm{seq}}$ consecutive transformed feature vectors (with optional warm-start padding), and it yields a label $y_k$ by selecting an aligned target value at a task-defined supervision index $j_{\mathrm{sup}}(k)$. The result is a windowed dataset $\mathcal{D}_{\mathrm{win}}=\{(\vW_k,y_k)\}$ that can be partitioned, batched, and evaluated consistently across model families.

\subsubsection{Windowing parameters}
\label{app:windowing_params}

Windowing is governed by five parameters. Intuitively, these specify (i) how much history the model sees, (ii) how densely windows are sampled, (iii) whether early windows may be left-padded, (iv) where supervision is read relative to the window, and (v) whether the target is a single value or a short segment. Formally: history length $L_{\mathrm{seq}}\in\mathbb{N}$; stride $\Delta\in\mathbb{N}$; warm-start depth $\rho\in\mathbb{Z}_{\ge 0}$ ($\rho=0$: strict, no padding; $\rho>0$: window may extend before index $1$); supervision offset $\delta\in\mathbb{Z}_{\ge 0}$ ($\delta=0$: end-of-window; $\delta>0$: $\delta$ steps later); and supervision segment length $L_{\mathrm{pred}}\in\mathbb{N}$ ($L_{\mathrm{pred}}=1$ for single window-level labels).

\paragraph{Admissibility precondition.}
\label{app:admissibility_precondition}
To guarantee that every admissible window carries a real-indexed supervision label, the warm-start depth must satisfy
\[
\rho \le L_{\mathrm{seq}} - 1 + \delta
\]
for single-step supervision (the multi-step analogue $\rho \le L_{\mathrm{seq}} - L_{\mathrm{lbl}} + \delta$ is stated in Section~\ref{app:multistep_supervision}). The closed-form $N_{\mathrm{slices}}$ and the label definition $y_k = z_y(j_{\mathrm{sup}}(k))$ below assume this precondition throughout; under $\rho = 0$ it holds trivially.

\subsubsection{Admissible window starts}
\label{app:admissible_window_starts}

Given the windowing parameters $L_{\mathrm{seq}}$, $\Delta$, $\rho$, $\delta$, and $L_{\mathrm{pred}}$, together with transformed length $T'$, we define the required right-side coverage length---i.e., the minimum number of transformed steps needed (from a window start) to extract one input window and its supervision:
\begin{equation}
    L_{\mathrm{req}} \triangleq L_{\mathrm{seq}} + \delta + L_{\mathrm{pred}}.
\end{equation}
Candidate window starts are enumerated as:
\begin{equation}
    k_m \triangleq 1-\rho + (m-1)\Delta,\qquad m=1,2,\dots
\end{equation}
subject to the constraint:
\begin{equation}
    k_m + L_{\mathrm{req}} - 1 \le T'.
\end{equation}
The number of admissible windows and the admissible start set are:
\begin{equation}
    N_{\mathrm{slices}} \triangleq \max\!\left\{\,0,\ \left\lfloor \frac{T' - L_{\mathrm{req}} + \rho}{\Delta}\right\rfloor + 1 \right\},\qquad
    \mathcal{K}\triangleq\{k_m\}_{m=1}^{N_{\mathrm{slices}}}.
\end{equation}
Here, $L_{\mathrm{req}}$ is the total right-side coverage required per window; $k_m$ enumerates starts with stride $\Delta$ from $1-\rho$ (the earliest possible start); the inequality enforces sufficient trajectory coverage; $N_{\mathrm{slices}}$ is the resulting number of windows; and $\mathcal{K}$ is the set of admissible start indices.

\subsubsection{Left-padding for the feature history}
\label{app:left_padding}

When $\rho>0$, some history-window indices may satisfy $k_m+i<1$. We introduce a padding operator $\mathcal{P}$ that defines feature values for indices $j\le 0$. The padded extension of the feature series is:
\begin{equation}
\widetilde{\vz}(j)=
\begin{cases}
\vz(j), & j\ge 1,\\
\mathcal{P}(\gZ,j), & j\le 0.
\end{cases}
\end{equation}
In strict mode ($\rho=0$), $\mathcal{P}$ is never invoked.

\subsubsection{History window}
\label{app:history_window}

For any admissible start $k\in\mathcal{K}$, the extracted input window is:
\begin{equation}
\vW_{k}
=
\big[\widetilde{\vz}(k),\ \widetilde{\vz}(k+1),\ \dots,\ \widetilde{\vz}(k+L_{\mathrm{seq}}-1)\big]^{\top}
\in \mathbb{R}^{L_{\mathrm{seq}}\times F}.
\end{equation}

\subsubsection{Window-level label}
\label{app:window_level_label}

In the default PHM benchmark instantiation (end-of-window supervision), the label is:
\begin{equation}
    y_k = z_y(k+L_{\mathrm{seq}}-1),
\end{equation}
which is always drawn from a real target index provided $\rho \le L_{\mathrm{seq}}-1$. More generally, for supervision offset $\delta\ge 0$, the supervision index is:
\begin{equation}
    j_{\mathrm{sup}}(k)=k+L_{\mathrm{seq}}-1+\delta,
\end{equation}
and the window-level label is $y_k=z_y(j_{\mathrm{sup}}(k))$, real-indexed for all $k\in\mathcal{K}$ under the admissibility precondition of Section~\ref{app:admissibility_precondition}.

\begin{figure}[t]
\centering
\input{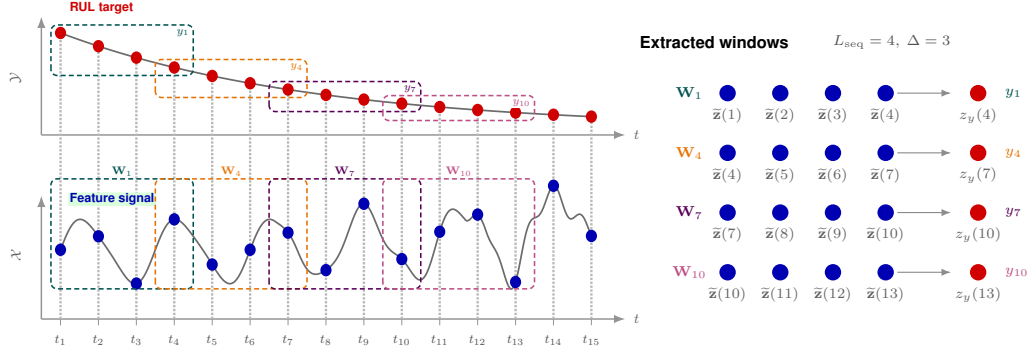}
\resizebox{0.98\linewidth}{!}{%
\begin{tikzpicture}[font=\sffamily, >=Latex]
\colorlet{wonewindow}{teal!65!black}
\colorlet{wfourwindow}{orange!90!black}
\colorlet{wsevenwindow}{violet!70!black}
\colorlet{wtenwindow}{magenta!75!black}
\tikzset{
  ruldot/.style={
    circle,
    draw=phmRUL,
    fill=red!12,
    line width=0.9pt,
    minimum size=0.34cm,
    inner sep=0pt
  },
  sourcewindow/.style={
    rounded corners=4pt,
    line width=1.0pt,
    dash pattern=on 3.2pt off 2.0pt
  },
  wonebox/.style={sourcewindow, draw=wonewindow},
  wfourbox/.style={sourcewindow, draw=wfourwindow},
  wsevenbox/.style={sourcewindow, draw=wsevenwindow},
  wtenbox/.style={sourcewindow, draw=wtenwindow}
}
\begin{scope}[xscale=0.90]
\DrawPHMSignalBaseScene{winslice}

\coordinate (wone-left) at ($(winslice-top-sw)!0.5!(winslice-time-0)$);
\coordinate (wone-right) at ($(winslice-time-3)!0.5!(winslice-time-4)$);
\coordinate (wfour-left) at ($(winslice-time-2)!0.5!(winslice-time-3)$);
\coordinate (wfour-right) at ($(winslice-time-6)!0.5!(winslice-time-7)$);
\coordinate (wseven-left) at ($(winslice-time-5)!0.5!(winslice-time-6)$);
\coordinate (wseven-right) at ($(winslice-time-9)!0.5!(winslice-time-10)$);
\coordinate (wten-left) at ($(winslice-time-8)!0.5!(winslice-time-9)$);
\coordinate (wten-right) at ($(winslice-time-12)!0.5!(winslice-time-13)$);

\coordinate (feature-window-y-sw) at ($(winslice-bottom-sw)+(0,0.68)$);
\coordinate (feature-window-y-ne) at ($(winslice-bottom-sw)+(0,3.16)$);

\draw[wonebox] ($ (wone-left |- feature-window-y-sw) $) rectangle ($ (wone-right |- feature-window-y-ne) $);
\draw[wfourbox] ($ (wfour-left |- feature-window-y-sw) $) rectangle ($ (wfour-right |- feature-window-y-ne) $);
\draw[wsevenbox] ($ (wseven-left |- feature-window-y-sw) $) rectangle ($ (wseven-right |- feature-window-y-ne) $);
\draw[wtenbox] ($ (wten-left |- feature-window-y-sw) $) rectangle ($ (wten-right |- feature-window-y-ne) $);

\node[
  ffmath,
  font=\sffamily\scriptsize\bfseries,
  text=wonewindow,
  anchor=south,
  fill=white,
  inner sep=1.0pt
] at ($($(wone-left |- feature-window-y-ne)!0.5!(wone-right |- feature-window-y-ne)$)+(0,0.03)$) {$\vW_1$};
\node[
  ffmath,
  font=\sffamily\scriptsize\bfseries,
  text=wfourwindow,
  anchor=south,
  fill=white,
  inner sep=1.0pt
] at ($($(wfour-left |- feature-window-y-ne)!0.5!(wfour-right |- feature-window-y-ne)$)+(0,0.03)$) {$\vW_4$};
\node[
  ffmath,
  font=\sffamily\scriptsize\bfseries,
  text=wsevenwindow,
  anchor=south,
  fill=white,
  inner sep=1.0pt
] at ($($(wseven-left |- feature-window-y-ne)!0.5!(wseven-right |- feature-window-y-ne)$)+(0,0.03)$) {$\vW_7$};
\node[
  ffmath,
  font=\sffamily\scriptsize\bfseries,
  text=wtenwindow,
  anchor=south,
  fill=white,
  inner sep=1.0pt
] at ($($(wten-left |- feature-window-y-ne)!0.5!(wten-right |- feature-window-y-ne)$)+(0,0.03)$) {$\vW_{10}$};

\coordinate (wone-target-sw) at ($(wone-left |- winslice-rul-3)+(0,-0.18)$);
\coordinate (wone-target-ne) at ($(wone-right |- winslice-rul-0)+(0,0.18)$);
\coordinate (wfour-target-sw) at ($(wfour-left |- winslice-rul-6)+(0,-0.18)$);
\coordinate (wfour-target-ne) at ($(wfour-right |- winslice-rul-3)+(0,0.18)$);
\coordinate (wseven-target-sw) at ($(wseven-left |- winslice-rul-9)+(0,-0.18)$);
\coordinate (wseven-target-ne) at ($(wseven-right |- winslice-rul-6)+(0,0.18)$);
\coordinate (wten-target-sw) at ($(wten-left |- winslice-rul-12)+(0,-0.18)$);
\coordinate (wten-target-ne) at ($(wten-right |- winslice-rul-9)+(0,0.18)$);

\draw[wonebox] (wone-target-sw) rectangle (wone-target-ne);
\draw[wfourbox] (wfour-target-sw) rectangle (wfour-target-ne);
\draw[wsevenbox] (wseven-target-sw) rectangle (wseven-target-ne);
\draw[wtenbox] (wten-target-sw) rectangle (wten-target-ne);

\node[
  ffmath,
  font=\sffamily\scriptsize\bfseries,
  text=wonewindow,
  anchor=north east,
  fill=white,
  inner sep=0.9pt
] at ($(wone-target-ne)+(-0.06,-0.05)$) {$y_1$};
\node[
  ffmath,
  font=\sffamily\scriptsize\bfseries,
  text=wfourwindow,
  anchor=north east,
  fill=white,
  inner sep=0.9pt
] at ($(wfour-target-ne)+(-0.06,-0.05)$) {$y_4$};
\node[
  ffmath,
  font=\sffamily\scriptsize\bfseries,
  text=wsevenwindow,
  anchor=north east,
  fill=white,
  inner sep=0.9pt
] at ($(wseven-target-ne)+(-0.06,-0.05)$) {$y_7$};
\node[
  ffmath,
  font=\sffamily\scriptsize\bfseries,
  text=wtenwindow,
  anchor=north east,
  fill=white,
  inner sep=0.9pt
] at ($(wten-target-ne)+(-0.06,-0.05)$) {$y_{10}$};
\end{scope}

\begin{scope}[cm={1.32,0,0,1.32,(-7.36cm,-1.42cm)}, transform shape]
\node[fftitle, anchor=west] at (16.18, 6.42) {Extracted windows};
\node[ffmath, anchor=west] at (19.40, 6.42) {$L_{\mathrm{seq}}=4,\ \Delta=3$};

\node[ffmath, text=wonewindow, anchor=west, text width=1.05cm, align=center]
  at (16.42, 5.54) {$\vW_1$};
\foreach \x in {17.72,18.62,19.52,20.42}
  \fill[phmSignal] (\x,5.54) circle[radius=0.14];
\draw[->, line width=0.72pt, draw=black!45] (20.62,5.54) -- (21.54,5.54);
\fill[phmRUL] (22.00, 5.54) circle[radius=0.14];
\node[ffmath, text=wonewindow, anchor=west] at (22.34, 5.54) {$y_1$};
\node[ffmath, text=black!62] at (17.72, 5.16) {$\widetilde{\vz}(1)$};
\node[ffmath, text=black!62] at (18.62, 5.16) {$\widetilde{\vz}(2)$};
\node[ffmath, text=black!62] at (19.52, 5.16) {$\widetilde{\vz}(3)$};
\node[ffmath, text=black!62] at (20.42, 5.16) {$\widetilde{\vz}(4)$};
\node[ffmath, text=black!62] at (22.00, 5.16) {$z_y(4)$};

\node[ffmath, text=wfourwindow, anchor=west, text width=1.05cm, align=center]
  at (16.42, 4.52) {$\vW_4$};
\foreach \x in {17.72,18.62,19.52,20.42}
  \fill[phmSignal] (\x,4.52) circle[radius=0.14];
\draw[->, line width=0.72pt, draw=black!45] (20.62,4.52) -- (21.54,4.52);
\fill[phmRUL] (22.00, 4.52) circle[radius=0.14];
\node[ffmath, text=wfourwindow, anchor=west] at (22.34, 4.52) {$y_4$};
\node[ffmath, text=black!62] at (17.72, 4.14) {$\widetilde{\vz}(4)$};
\node[ffmath, text=black!62] at (18.62, 4.14) {$\widetilde{\vz}(5)$};
\node[ffmath, text=black!62] at (19.52, 4.14) {$\widetilde{\vz}(6)$};
\node[ffmath, text=black!62] at (20.42, 4.14) {$\widetilde{\vz}(7)$};
\node[ffmath, text=black!62] at (22.00, 4.14) {$z_y(7)$};

\node[ffmath, text=wsevenwindow, anchor=west, text width=1.05cm, align=center]
  at (16.42, 3.50) {$\vW_7$};
\foreach \x in {17.72,18.62,19.52,20.42}
  \fill[phmSignal] (\x,3.50) circle[radius=0.14];
\draw[->, line width=0.72pt, draw=black!45] (20.62,3.50) -- (21.54,3.50);
\fill[phmRUL] (22.00, 3.50) circle[radius=0.14];
\node[ffmath, text=wsevenwindow, anchor=west] at (22.34, 3.50) {$y_7$};
\node[ffmath, text=black!62] at (17.72, 3.12) {$\widetilde{\vz}(7)$};
\node[ffmath, text=black!62] at (18.62, 3.12) {$\widetilde{\vz}(8)$};
\node[ffmath, text=black!62] at (19.52, 3.12) {$\widetilde{\vz}(9)$};
\node[ffmath, text=black!62] at (20.42, 3.12) {$\widetilde{\vz}(10)$};
\node[ffmath, text=black!62] at (22.00, 3.12) {$z_y(10)$};

\node[ffmath, text=wtenwindow, anchor=west, text width=1.05cm, align=center]
  at (16.42, 2.48) {$\vW_{10}$};
\foreach \x in {17.72,18.62,19.52,20.42}
  \fill[phmSignal] (\x,2.48) circle[radius=0.14];
\draw[->, line width=0.72pt, draw=black!45] (20.62,2.48) -- (21.54,2.48);
\fill[phmRUL] (22.00, 2.48) circle[radius=0.14];
\node[ffmath, text=wtenwindow, anchor=west] at (22.34, 2.48) {$y_{10}$};
\node[ffmath, text=black!62] at (17.72, 2.10) {$\widetilde{\vz}(10)$};
\node[ffmath, text=black!62] at (18.62, 2.10) {$\widetilde{\vz}(11)$};
\node[ffmath, text=black!62] at (19.52, 2.10) {$\widetilde{\vz}(12)$};
\node[ffmath, text=black!62] at (20.42, 2.10) {$\widetilde{\vz}(13)$};
\node[ffmath, text=black!62] at (22.00, 2.10) {$z_y(13)$};
\end{scope}

\end{tikzpicture}
}
\caption{Windowing and label alignment on an illustrative aligned PHM trajectory pair. On the left, the visible source windows are outlined directly on the shared feature and target trajectories for $\vW_1$, $\vW_4$, $\vW_7$, and $\vW_{10}$ under sequence length $L_{\mathrm{seq}}=4$ and stride $\Delta=3$. On the right, those same extracted windows are written explicitly row by row, with each ordered transformed feature sequence $\widetilde{\vz}(i)$ paired to its attached aligned target $z_y(j)$ under the default end-of-window rule $j_{\mathrm{sup}}(k)=k+L_{\mathrm{seq}}-1$.}
\label{fig:formalization_windowing_operator}
\end{figure}

Figure~\ref{fig:formalization_windowing_operator} illustrates the default end-of-window case: one highlighted history slice produces $\vW_k$, and the aligned target at the supervision index provides $y_k$.

\subsubsection{Windowed dataset and stride subsampling}
\label{app:windowed_dataset_stride}

The resulting windowed dataset for a single unit is:
\begin{equation}
    \mathcal{D}_{\mathrm{win}} \triangleq \{(\vW_k, y_k)\}_{k\in\mathcal{K}}.
\end{equation}
The total number of windows $N_{\mathrm{slices}}$ is directly controlled by the stride $\Delta$; adjusting $\Delta$ performs subsampling without altering window structure.

\subsubsection{Optional multi-step supervision segment}
\label{app:multistep_supervision}

If a dataset specifies a multi-step supervision segment (e.g., for auxiliary objectives or multi-horizon forecasting), one may define:
\begin{equation}
\mathbf{y}_{k}
=
\big[z_y(k+L_{\mathrm{seq}}-L_{\mathrm{lbl}}+\delta),\ \dots,\ z_y(k+L_{\mathrm{seq}}+\delta+L_{\mathrm{pred}}-1)\big]^{\top},
\end{equation}
where $L_{\mathrm{lbl}}\in\mathbb{N}$ denotes the segment overlap length. The vector $\mathbf{y}_k$ contains only real (non-padded) targets for all admissible $k\in\mathcal{K}$ provided the warm-start depth satisfies:
\begin{equation}
\rho \le L_{\mathrm{seq}}-L_{\mathrm{lbl}}+\delta,
\end{equation}
which strengthens the single-step precondition of Section~\ref{app:admissibility_precondition} when $L_{\mathrm{lbl}} > 1$, together with the admissibility condition $k+L_{\mathrm{req}}-1\le T'$ already enforced by $\mathcal{K}$.

\subsubsection{Generalization to multiple units}
\label{app:multiunit_windowing}

For datasets containing $U$ units with possibly different transformed lengths $\{T'_u\}_{u=1}^{U}$, the windowing construction is applied independently to each unit's pair $(\gZ^{(u)},\gY'^{(u)})$, yielding per-unit window sets $\mathcal{D}^{(u)}_{\mathrm{win}}=\{(\vW^{(u)}_{k},y^{(u)}_{k})\}$. The global dataset is then:
\begin{equation}
\mathcal{D}_{\mathrm{global}}=\bigcup_{u=1}^{U}\mathcal{D}^{(u)}_{\mathrm{win}}.
\end{equation}
Equivalently, one may index samples by the pair $(u,k)$, where $u$ identifies the unit and $k\in\mathcal{K}^{(u)}$ is that unit's window index, so each global sample has an unambiguous identity even when different units share the same transformed-time index values.
Partitioning (inter-unit or intra-unit temporal) is performed as specified in Section~\ref{app:partitioning_full}, with all fitted preprocessing parameters estimated on training data only.

\subsection{Partitioning and leakage control}
\label{app:partitioning_full}

This section formalizes the two partitioning regimes introduced in Section~\ref{subsec:partitioning} and states the leakage policy as a set of formal constraints. The global dataset
\[
\mathcal{D}_{\mathrm{global}} = \bigcup_{u=1}^{U}\mathcal{D}^{(u)}_{\mathrm{win}}
\]
is partitioned into three disjoint subsets
\[
\mathcal{D}^{\mathrm{train}} \cup \mathcal{D}^{\mathrm{val}} \cup \mathcal{D}^{\mathrm{test}} = \mathcal{D}_{\mathrm{global}}, \qquad
\mathcal{D}^{\mathrm{train}} \cap \mathcal{D}^{\mathrm{val}} = \mathcal{D}^{\mathrm{train}} \cap \mathcal{D}^{\mathrm{test}} = \mathcal{D}^{\mathrm{val}} \cap \mathcal{D}^{\mathrm{test}} = \emptyset.
\]
Two regimes induce this partition.

\subsubsection{Inter-unit partitioning}
\label{app:inter_unit_partitioning}

Let $\mathcal{U} = \{1,\dots,U\}$ denote the set of unit indices. Inter-unit partitioning assigns each unit to exactly one partition via three disjoint unit subsets satisfying
\[
\mathcal{U}^{\mathrm{train}} \cup \mathcal{U}^{\mathrm{val}} \cup \mathcal{U}^{\mathrm{test}} = \mathcal{U}, \qquad
\mathcal{U}^{\mathrm{train}} \cap \mathcal{U}^{\mathrm{val}} = \mathcal{U}^{\mathrm{train}} \cap \mathcal{U}^{\mathrm{test}} = \mathcal{U}^{\mathrm{val}} \cap \mathcal{U}^{\mathrm{test}} = \emptyset.
\]
Each unit's entire window set is routed to the partition of its unit:
\[
\mathcal{D}^{\mathrm{split}} = \bigcup_{u \in \mathcal{U}^{\mathrm{split}}} \mathcal{D}^{(u)}_{\mathrm{win}}, \qquad \mathrm{split} \in \{\mathrm{train},\mathrm{val},\mathrm{test}\}.
\]
This regime measures generalization to previously unseen units.

\subsubsection{Intra-unit temporal partitioning}
\label{app:intra_unit_partitioning}

Each unit $u$ is split chronologically along its transformed timeline by two boundaries $\tau^{(u)}_{\mathrm{train}} \le \tau^{(u)}_{\mathrm{val}} \le T'_u$. Let $\mathcal{K}^{(u)}$ denote the admissible window-start set for unit $u$ constructed by Section~\ref{app:windowing_full}. To partition windows we use their \emph{supervision index} $j_{\mathrm{sup}}(k) = k + L_{\mathrm{seq}} - 1 + \delta$ --- the transformed-time index at which the label of window $k$ is attached (Section~\ref{app:windowing_full}). Partitioning by $j_{\mathrm{sup}}(k)$ rather than by the window start $k$ ensures that every training window has its \emph{label} at or before the training boundary, so no future target information enters the training partition:
\begin{align*}
\mathcal{K}^{(u)}_{\mathrm{train}} &= \{\, k \in \mathcal{K}^{(u)} : j_{\mathrm{sup}}(k) \le \tau^{(u)}_{\mathrm{train}} \,\}, \\
\mathcal{K}^{(u)}_{\mathrm{val}} &= \{\, k \in \mathcal{K}^{(u)} : \tau^{(u)}_{\mathrm{train}} < j_{\mathrm{sup}}(k) \le \tau^{(u)}_{\mathrm{val}} \,\}, \\
\mathcal{K}^{(u)}_{\mathrm{test}} &= \{\, k \in \mathcal{K}^{(u)} : \tau^{(u)}_{\mathrm{val}} < j_{\mathrm{sup}}(k) \,\}.
\end{align*}
The three sets are pairwise disjoint by construction and their union equals $\mathcal{K}^{(u)}$. Per-unit partitioned window sets and their global aggregation are
\[
\mathcal{D}^{(u),\mathrm{split}}_{\mathrm{win}} = \{(\vW^{(u)}_k, y^{(u)}_k) : k \in \mathcal{K}^{(u)}_{\mathrm{split}}\}, \qquad
\mathcal{D}^{\mathrm{split}} = \bigcup_{u=1}^{U} \mathcal{D}^{(u),\mathrm{split}}_{\mathrm{win}}.
\]
This regime measures predictive ability within each unit's own trajectory.

\subsubsection{Leakage policy}
\label{app:partitioning_leakage}

In both regimes, the following constraints are part of the benchmark definition:
\begin{itemize}
    \item Feature parameters $\Psi$ are estimated using training data only (cf.\ Section~\ref{app:feature_transform_full}).
    \item Target-pipeline parameters $\Phi$ and any alignment statistics required by $\mathcal{A}$ are estimated using training data only (cf.\ Section~\ref{app:target_transform_full}).
    \item Any context selection rule $\pi$ for in-context learning draws only from the training partition (cf.\ Section~\ref{app:tabularization_full}).
    \item Hyper-parameters are selected using validation data only.
    \item Test data are accessed exclusively for final evaluation.
\end{itemize}
For intra-unit partitioning, ``training data'' means the portion of each unit available up to the training boundary $\tau^{(u)}_{\mathrm{train}}$ (in transformed time). In this regime, all preprocessing must be \emph{causal with respect to the split}: any fitted stage of $\mathcal{G}$ or $\mathcal{H}$ operating on unit $u$ may use only raw samples that influence transformed indices $j \le \tau^{(u)}_{\mathrm{train}}$ (as captured by the support map $a(\cdot)$ in Section~\ref{app:feature_transform_full}). This rules out statistics that implicitly peek at future timestamps within the same unit, such as scalers fit over the full unit trajectory or spectral normalization computed using post-boundary samples.

\subsection{Evaluation protocol}
\label{app:evaluation_full}

Any model evaluated under the framework must implement a function $f$ mapping an arbitrary input window $\vW$ to a prediction $\hat{y}$.

The framework supports two evaluator types; the choice between them is a property of the benchmark configuration for a given dataset, not of the task family.

\textbf{Window-level evaluator.} Let $\mathcal{K}^{\mathrm{test}}$ denote the set of window indices in the held-out test partition, so $\mathcal{D}^{\mathrm{test}} = \{(\vW_k, y_k)\}_{k \in \mathcal{K}^{\mathrm{test}}}$. Performance is computed as the mean per-sample metric
\[
\mathcal{L}_{\mathrm{test}} = \frac{1}{|\mathcal{D}^{\mathrm{test}}|}\sum_{k \in \mathcal{K}^{\mathrm{test}}} \ell(\hat{y}_k, y_k).
\]

\textbf{Per-unit evaluator.} Let $\mathcal{U}^{\mathrm{test}}$ denote the set of test units and, for each $u \in \mathcal{U}^{\mathrm{test}}$, let $\mathcal{K}^{(u)}_{\mathrm{test}} \subseteq \mathcal{K}^{(u)}$ denote the chronologically ordered window indices belonging to corresponding unit's test range (Section~\ref{app:partitioning_full}). Define the ordered prediction and ground-truth trajectories
\[
\hat{\mathbf{y}}^{(u)} = \{\hat{y}_k : k \in \mathcal{K}^{(u)}_{\mathrm{test}}\}, \qquad \mathbf{y}^{(u)} = \{y_k : k \in \mathcal{K}^{(u)}_{\mathrm{test}}\}.
\]
Unit-level metrics are then computed and averaged:
\[
\mathcal{L}_{\mathrm{test}}^{\mathrm{per\text{-}unit}} = \frac{1}{|\mathcal{U}^{\mathrm{test}}|}\sum_{u \in \mathcal{U}^{\mathrm{test}}} \ell^{(u)}\!\left(\hat{\mathbf{y}}^{(u)},\, \mathbf{y}^{(u)}\right),
\]
where $\ell^{(u)}$ is a unit-level metric (e.g., RMSE, PHM-score).

These two evaluators differ whenever units contribute different numbers of test windows or when the metric is defined at the trajectory level. For a per-window scoring function $\ell(\hat{y}_k,y_k)$, let $n_u \triangleq |\mathcal{K}^{(u)}_{\mathrm{test}}|$. The window-level evaluator can be rewritten as a weighted average over units:
\[
\mathcal{L}_{\mathrm{test}}
=
\frac{1}{\sum_{u \in \mathcal{U}^{\mathrm{test}}} n_u}
\sum_{u \in \mathcal{U}^{\mathrm{test}}}
\sum_{k \in \mathcal{K}^{(u)}_{\mathrm{test}}}
\ell(\hat{y}_k,y_k),
\]
so units with more test windows receive higher weight. By contrast, the per-unit evaluator assigns equal weight to each unit by aggregating within unit and then averaging across units, and it also supports trajectory-level scoring functions $\ell^{(u)}$ that cannot be expressed as sums of per-window losses. The per-unit evaluator is therefore preferred when the intended notion of generalization is ``per asset'' rather than ``per window,'' or when the task metric is naturally unit-level.

For intuition: if one unit contributes $10\times$ more test windows than another (e.g., it runs longer), the window-level evaluator implicitly weights that unit $10\times$ more; the per-unit evaluator weights both units equally.

All models are evaluated on identical windows, identical splits, and identical metrics; the evaluator choice determines whether aggregation is window-weighted or unit-weighted.

\subsection{Model families and fitting conventions}
\label{app:training_objectives}

While the benchmark specification is model-agnostic (it defines evaluation, not training), we describe the inference and fitting conventions for three representative model families currently supported by the PICID library.

\subsubsection{Sequential models}
\label{app:sequential_models}

Sequence models (e.g., LSTM, 1D-CNN, Transformers) operate directly on the window representation $\vW_k \in \mathcal{D}_{\mathrm{win}}^{\mathrm{train}}$ and are trained to minimize the empirical risk over the training partition:
\begin{equation}
    \theta^* = \operatorname*{arg\,min}_{\theta}\;\frac{1}{|\mathcal{D}_{\mathrm{win}}^{\mathrm{train}}|}\sum_{(\vW_k,y_k)\in\mathcal{D}_{\mathrm{win}}^{\mathrm{train}}} \ell\big(f_{\theta}(\vW_k),\;y_k\big),
\end{equation}
where $f_{\theta}$ is the model parameterized by $\theta$ and $\ell(\cdot,\cdot)$ is a task-dependent loss (e.g., MSE for prognostics, cross-entropy for diagnostics). At inference, $\hat{y}_k = f_{\theta^*}(\vW_k)$.

\subsubsection{Fit-predict models}
\label{app:fit_predict_models}

Fit-predict models (e.g., XGBoost, Random Forest, SVR) operate on the tabular representation $\vX_k \in \mathcal{D}_{\mathrm{tab}}^{\mathrm{train}}$ from Section~\ref{app:tabularization_full}. A fitting operator $\mathcal{F}$ maps the training tabular dataset to a fitted parameter state
\begin{equation}
    \xi^* = \mathcal{F}\!\left(\mathcal{D}_{\mathrm{tab}}^{\mathrm{train}}\right),
\end{equation}
produced in a single non-iterative pass rather than through gradient-based empirical risk minimization. Inference applies the resulting predictor $\hat{y}_k = f_{\xi^*}(\vX_k)$. The leakage policy of Section~\ref{app:partitioning_leakage} applies: $\mathcal{F}$ accesses only training-partition samples, and any internal model-selection procedure embedded in $\mathcal{F}$ is restricted to that partition.

\subsubsection{Tabular foundation models: zero-shot in-context learning}
\label{app:tabular_foundation_models}

Tabular foundation models (e.g., TabPFN, TabDPT) perform zero-shot in-context learning (ICL) using frozen pre-trained weights $\phi$. These models approximate the Bayesian posterior over test labels conditioned on a labeled context set $\mathcal{C}$ drawn from the training partition:
\begin{equation}
    \hat{y}_q = f_{\phi}(\mathcal{C},\;\vX_q), \qquad \mathcal{C} = \{(\vX_j,y_j)\}_{j=1}^{|\mathcal{C}|} \subset \mathcal{D}_{\mathrm{tab}}^{\mathrm{train}}.
\end{equation}
No weight updates are performed; the model adapts to the PHM task solely through the information provided in $\mathcal{C}$ at inference time. Context size is fixed in advance per benchmark configuration. If $|\mathcal{D}_{\mathrm{tab}}^{\mathrm{train}}|$ exceeds the model's context limit, $\mathcal{C}$ is constructed via a context-selection rule $\pi$ fixed by the benchmark configuration. Under intra-unit partitioning, every context element $j$ drawn from the query's own unit $u(k)$ must satisfy $j_{\mathrm{sup}}(j) \le \tau^{(u(k))}_{\mathrm{train}}$, i.e., its supervision index lies within that unit's training time range; when the query is itself a training sample it is excluded from its own context ($k \notin \mathcal{J}_k$, where $\mathcal{J}_k$ indexes the selected context elements). Any stochastic component of $\pi$ is seeded so that every compared model receives the same $\mathcal{C}_k$ for a given query, preserving the benchmark invariant that performance differences reflect model differences rather than context-sampling variation.

\subsubsection{Unified inference summary}
\label{app:unified_inference_summary}

Regardless of model family, all predictions are produced from the same benchmark inputs and evaluated on the same test instances:
\begin{equation}
    \hat{y}_k =
    \begin{cases}
        f_{\theta^*}(\vW_k) & \text{sequential,}\\
        f_{\xi^*}(\vX_k) & \text{fit-predict,}\\
        f_{\phi}(\mathcal{C},\vX_k) & \text{tabular ICL.}
    \end{cases}
\end{equation}

\subsection{Tabularization and dual representation}
\label{app:tabularization_full}

To support tabular methods alongside sequential models, we define a representation adapter:
\[
\mathcal{T} : \mathbb{R}^{L_{\mathrm{seq}} \times F} \rightarrow \mathbb{R}^{D_{\mathrm{tab}}}, \qquad D_{\mathrm{tab}} \triangleq L_{\mathrm{seq}}F,
\]
which flattens each window into a vector using time-major ordering:
\begin{align*}
\vX_k = \mathcal{T}(\vW_k)
&= \big[z_1(k),\dots,z_F(k),\;z_1(k+1),\dots,z_F(k+1),\\
&\qquad \dots,\;z_1(k+L_{\mathrm{seq}}-1),\dots,z_F(k+L_{\mathrm{seq}}-1)\big]^{\top}.
\end{align*}
The mapping admits an inverse $\mathcal{T}^{-1}$ that recovers $\vW_k$ from $\vX_k$ by restoring the $(L_{\mathrm{seq}},F)$ structure; no information is lost under $\mathcal{T}$.

The dual dataset representations are:
\begin{itemize}
    \item \textbf{Window (sequence) dataset:} $\mathcal{D}_{\mathrm{win}}=\{(\vW_k,y_k)\}_{k\in\mathcal{K}}$
    \item \textbf{Tabular dataset:} $\mathcal{D}_{\mathrm{tab}}=\{(\vX_k,y_k)\}_{k\in\mathcal{K}}$, where $\vX_k=\mathcal{T}(\vW_k)$
\end{itemize}
These contain identical information content and inherit the same split membership.

\section{Complete PICID library architecture}
\label{app:library_architecture}

This appendix expands Section~\ref{sec:picid_library} into a complete module-level view of the \picid{} software stack. The goal is to make the end-to-end benchmark executable and inspectable: raw datasources are loaded and split, transformed into aligned feature/target sequences, windowed into supervised samples, consumed by model wrappers, and scored by a shared evaluator/reporting layer. Although the library also supports forecasting configurations, we focus here on the three PHM task categories formalized in the main paper.

The protocol is realized through configuration-driven composition: experiment definitions, transform pipelines, datasource selection, and training settings are specified declaratively in YAML and composed into a deterministic run by the orchestrator. Each subsection states its \emph{protocol role} explicitly; Table~\ref{tab:module_protocol_map_app} summarizes the mapping, while Appendix~\ref{app:complete_formalization} remains the authoritative source for the full mathematical definitions.

\begin{table}[H]
\centering
\footnotesize
\caption{Module--protocol mapping for the complete library architecture appendix. Each row links a \picid{} module family to its formal role in Section~\ref{sec:framework_formalization}.}
\label{tab:module_protocol_map_app}

\renewcommand{\arraystretch}{1.15}
\begin{tabular}{@{}lll@{}}
\toprule
\textbf{Module} & \textbf{Protocol operator} & \textbf{Data flow} \\
\midrule
Datasources (\ref{app:lib_datasources}) & Ingestion, split policy $\mathcal{P}_{\mathrm{split}}$ & Raw files $\to$ $\gX,\gY,\mathcal{D}^{\mathrm{train/val/test}}$ \\
Typed containers (\ref{app:lib_typed_containers}) & Split-safe transport & Split payload $\to$ validated typed views \\
Transform pipeline (\ref{app:lib_transforms}) & $\mathcal{G}(\cdot;\Psi),\widetilde{\mathcal{H}}(\cdot;\Phi)$ & $\gX,\gY,\mathcal{D}^{\mathrm{train}} \to \gZ,\gY'$ \\
Datasets \& sequencers (\ref{app:lib_datasets}) & Windowing $\mathcal{S}$ & $\gZ,\gY' \to \{(\vW_k,y_k)\}_{k\in\mathcal{K}}$ \\
Datamodule (\ref{app:lib_datamodule}) & Batching & $\{(\vW_k,y_k)\} \to$ mini-batches \\
Model wrappers (\ref{app:lib_wrappers}) & Predictor $f(\vW)=\hat{y}$ & $\vW_k \to \hat{y}_k$ \\
Evaluator (\ref{app:lib_evaluation}) & $\ell(\cdot,\cdot),\mathcal{L}_{\mathrm{test}}$ & $\{(\hat{y}_k,y_k)\} \to$ task metrics \\
Extension interfaces (\ref{app:lib_interfaces}) & Protocol-preserving extensibility & Reader-facing extension surfaces \\
\bottomrule
\end{tabular}
\end{table}

\clearpage

\begin{algorithm}[H]
\caption{End-to-end experiment execution pipeline (high level).}
\label{alg:execution_pipeline}
\footnotesize
\begin{algorithmic}[1]
\Require Resolved configuration \texttt{cfg} (datasource, transforms, task definition, model, trainer, evaluator)
\Ensure Reported metrics and persisted run artifacts
\State $\texttt{datasource} \gets \textsc{Instantiate}(\texttt{cfg.datasource})$
\State $\texttt{transforms} \gets \textsc{BuildSequence}(\texttt{cfg.transforms})$
\State $\texttt{container} \gets \textsc{PreprocessWithCaching}(\texttt{datasource},\,\texttt{transforms},\,\texttt{cfg.cache})$ \Comment{See Algorithm~\ref{alg:preprocessor}}
\State $\texttt{dataset} \gets \textsc{BuildDataset}(\texttt{container},\,\texttt{cfg.task\_definition})$
\State $\texttt{datamodule} \gets \textsc{BuildDataModule}(\texttt{dataset},\,\texttt{cfg.datamodule})$
\State $\texttt{model} \gets \textsc{BuildWrapper}(\texttt{cfg.model},\,\texttt{cfg.loss},\,\texttt{cfg.optim})$
\State $\texttt{trainer} \gets \textsc{BuildTrainer}(\texttt{cfg.trainer},\,\texttt{cfg.callbacks},\,\texttt{cfg.loggers})$
\State $\texttt{evaluator} \gets \textsc{BuildEvaluator}(\texttt{cfg.evaluator})$
\State
\State \textbf{Training:} $\texttt{trainer.fit}(\texttt{model},\,\texttt{datamodule})$
\State \hspace{1.2em}\textbf{for} epoch $=1,\dots,E$ \textbf{do}
\State \hspace{2.4em}\textbf{for} batch $\in$ \texttt{datamodule.train} \textbf{do} \Comment{Windowed samples from $\mathcal{S}$}
\State \hspace{3.6em}$\hat{y} \gets \texttt{model}(\texttt{batch})$ \Comment{Forward pass}
\State \hspace{3.6em}$\mathcal{L} \gets \ell(\hat{y},\,y)$ \Comment{Task loss}
\State \hspace{3.6em}\textsc{BackpropAndStep}($\mathcal{L}$) \Comment{Optimizer/scheduler update}
\State \hspace{2.4em}\textbf{end for}
\State \hspace{2.4em}\textsc{ValidateAndCheckpoint}(\texttt{datamodule.val},\,\texttt{evaluator}) \Comment{Early stopping/model selection}
\State \hspace{1.2em}\textbf{end for}
\State
\State \textbf{Testing:} $\texttt{trainer.test}(\texttt{model},\,\texttt{datamodule.test})$
\State $\texttt{metrics} \gets \texttt{evaluator}(\hat{y},\,y)$ \Comment{Per-task aggregation \& (optional) inverse scaling}
\State \textsc{PersistRunArtifacts}(\texttt{cfg},\,\texttt{code\_version},\,\texttt{metrics},\,\texttt{optional predictions})
\end{algorithmic}
\end{algorithm}

\subsection{Datasources}
\label{app:lib_datasources}

\emph{Protocol role.}\; $\text{Raw files} \xrightarrow{\text{load},\,\mathcal{P}_{\mathrm{split}}} \gX=\{\vx(t)\}_{t=1}^{T},\gY=\{y(t)\}_{t=1}^{T},\mathcal{D}^{\mathrm{train}},\mathcal{D}^{\mathrm{val}},\mathcal{D}^{\mathrm{test}}$.

Datasources provide a single, uniform way to load raw PHM data and to define train/validation/test partitions. This layer is intentionally data-centric: it decides what a ``unit'' is, how measurements and targets are exposed as arrays, and how splits are produced; task-specific semantics (alignment, windowing, tabularization) are handled downstream. The library includes datasources spanning turbofan engines, bearings, batteries, buildings, railway traction, and aerospace domains, although this paper evaluates a smaller subset documented in Appendix~\ref{sec:dataset_descriptions}.

Concretely, the datasource subsystem supports four roles:
\begin{itemize}
    \item \textbf{Contract and lifecycle:} a common protocol for \emph{load}, \emph{split}, and \emph{get} that yields typed containers and tracks loader state.
    \item \textbf{Single-source loaders:} datasets comprising one logical operational unit (or one unit per file) that can be split into train/validation/test within that unit.
    \item \textbf{Multi-source composition:} datasets comprising multiple units collected under different operating conditions, with splits defined either within each unit or by assigning entire units to splits.
    \item \textbf{Predefined splits:} datasets that ship with fixed train/validation/test partitions, for which splitting is a no-op and the loader materializes the provided split structure.
\end{itemize}
This separation keeps benchmark semantics stable. Adding a new datasource usually means specifying the raw data contract (what keys exist, how units are organized, and what split policy applies), while downstream semantics (target alignment and windowing) remain shared across datasets. Because inter-unit vs.\ intra-unit splitting is fixed at the datasource level, partitioning becomes a protocol invariant rather than a model-level choice.

\subsection{Typed data containers}
\label{app:lib_typed_containers}

\emph{Protocol role.}\; Structural transport that preserves type safety, metadata, and split consistency across all pipeline stages.

A typed container hierarchy transports split-organized data through the pipeline. Higher-level dataset containers add split-aware storage, persistent identifiers/metadata (e.g., unit IDs, split membership, and any timestamps needed to interpret sequences), and basic consistency checks across splits. Multiple access patterns are supported (e.g., split-keyed dictionaries and unit-grouped iterators), so modules can interoperate without depending on a single in-memory layout.

This boundary is where the protocol becomes enforceable in software. Centralizing structural checks (e.g., verifying that feature and target containers refer to the same units, and that transformations do not scramble split membership or unit ordering) catches inconsistencies early, before they reach dataset construction or evaluation. The same abstraction also supports both dense arrays and ragged multi-unit representations\footnote{We use \emph{dense} to denote fixed-length arrays (e.g., a single segment/cycle with shape $(T,F)$), and \emph{ragged} to denote arrays where the sequenced axis has variable length across segments (e.g., multiple units or battery cycles, giving a structure like $(N_{\text{seg}},\text{var-}T,F)$). Ragged storage avoids padding/masking to a common length, which is especially useful in PHM where trajectories (or cycles within a unit) can differ substantially in duration.} without rewriting downstream modules.

\subsection{Transform pipeline and preprocessing}
\label{app:lib_transforms}

\emph{Protocol role.}\; $\gX,\gY,\mathcal{D}^{\mathrm{train}} \xrightarrow{\mathcal{G}(\cdot;\Psi),\,\widetilde{\mathcal{H}}(\cdot;\Phi)} \gZ=\{\vz(j)\}_{j=1}^{T'},\gY'=\{z_y(j)\}_{j=1}^{T'}$.

The transform system implements the feature pipeline $\mathcal{G}$ and the target transformation-and-alignment pipeline $\widetilde{\mathcal{H}}$ (Section~\ref{subsec:construction}, Appendix~\ref{app:target_transform_full}). Each transform is a composable unit with two separable components:
\begin{itemize}
    \item \textbf{Logic:} the algorithmic computation, including (i) an optional \emph{fit} step that estimates parameters from a reference split, and (ii) an \emph{apply} step that maps an input payload to an output payload using those parameters and user-specified hyperparameters (e.g., window length, frequency bands, clipping thresholds).
    \item \textbf{Metadata:} declarative wiring that tells the orchestrator how to run the logic: which container fields to read and write, whether fitting is enabled, which split is allowed for fitting, and whether statistics are pooled across units or computed per unit.
\end{itemize}
For example, a standardization transform may \emph{fit} mean/variance on the training split and then \emph{apply} $x \mapsto (x-\mu)/\sigma$ to every split; its metadata specifies the input field (e.g., raw features), the output field (e.g., standardized features), and the fitting split/scope.
A configuration manager instantiates the ordered transform sequence from YAML and executes it split-wise, freezing all fitted parameters after the training partition. Appendix~\ref{app:yaml_example} provides a detailed configuration walkthrough, transform inventory, and caching pseudocode.

This design resolves a common PHM tension: some operations should be unit-local (e.g., resampling or time--frequency feature extraction), while others need shared training statistics to keep scales comparable (e.g., global normalization constants). \picid{} separates what a transform computes from how it is fit/applied, and the explicit \texttt{fit\_on} constraint prevents accidental use of validation/test information; the orchestrator records the fitted state as part of the experiment definition.

\paragraph{Preprocessing orchestration and caching.} The preprocessing orchestrator sequences the full load\,$\to$\,split\,$\to$\,transform pipeline as one deterministic execution. For efficiency, it supports cache-enabled runs where intermediate states are stored under a deterministic hash of the datasource configuration, transform configuration, and code fingerprint. Cache checkpoints exist at three levels:
\begin{itemize}
    \item \textbf{Raw-loaded data:} the post-ingestion container before any transformation.
    \item \textbf{Boundary checkpoints:} intermediate states after selected transform prefixes.
    \item \textbf{Fully preprocessed outputs:} the final transformed containers consumed by dataset construction.
\end{itemize}
Caching belongs to the benchmark contract because it never changes what is computed; it only changes how efficiently the fixed computation is replayed.

\subsection{Datasets and sequencers}
\label{app:lib_datasets}

\emph{Protocol role.}\; $\gZ,\gY' \xrightarrow{\mathcal{S}} \{(\vW_k,y_k)\}_{k\in\mathcal{K}}$ --- windowed supervised instances.

Datasets are task-centric consumers of the transformed containers. They format preprocessed data for specific model families, while sequencers handle the dense or ragged index generation associated with window length $L_{\mathrm{seq}}$, stride $\Delta$, supervision offset $\delta$, and related slicing parameters. In other words, this layer is the software realization of the windowing operator $\mathcal{S}$ from Section~\ref{subsubsec:windowing} and Appendix~\ref{app:windowing_full}. The same preprocessed trajectory can therefore be turned into sequence-model samples, tabular fit-predict samples, or other model-facing views without redefining the upstream transformation contract.
Sequencers are implemented to make batching efficient for both dense and variable-length (ragged) multi-unit data: they precompute a compact window index table and materialize entire batches via vectorized indexing, avoiding Python-level per-window loops during training and evaluation.

Separating datasets from datasources enforces a clean boundary between data-centric and task-centric concerns. Datasources decide what raw units and partitions exist; datasets decide how those split-safe units are presented as samples. Sequencers isolate the indexing logic needed for variable-length units and global sample addressing (which time ranges are valid windows, and where their supervision indices fall), which keeps dataset implementations simple and avoids re-deriving the same ragged indexing logic in every dataset class.

\subsection{Datamodule and batching}
\label{app:lib_datamodule}

\emph{Protocol role.}\; $\{(\vW_k,y_k)\} \to$ mini-batches for the training loop.

The datamodule constructs partition-specific data loaders with configurable batch size, shuffling, collation, and optional subsetting. Batch structure is dataset-aware: sequential prognostic datasets may return features, targets, and unit-context identifiers, while other task families may expose different but still standardized batch payloads. The key design choice is that batching remains an operational layer rather than a task-definition layer: it preserves the sample semantics created by the dataset and does not redefine the benchmark contract.

This thinness is intentional. By keeping split logic and sample construction out of the datamodule, PICID ensures that model wrappers can adapt batch contents to backbone-specific needs without taking ownership of partitioning, window semantics, or leakage control.

\subsection{Model wrappers}
\label{app:lib_wrappers}

\emph{Protocol role.}\; $\vW_k \xrightarrow{f} \hat{y}_k$ --- predictor mapping.

Model wrappers provide the unified interface between heterogeneous model backbones and the shared training/evaluation loop. Feed-forward wrappers support batchwise deep learning models, while fit-predict wrappers support scikit-learn-style estimators and other non-gradient baselines. All wrappers emit a standardized output dictionary containing predictions together with the aligned targets forwarded from the input batch. This is the direct software realization of the model-agnostic evaluation contract from Section~\ref{subsec:evaluation}: any method that consumes the defined window representation and produces a valid prediction can be evaluated under identical benchmark conditions.

The wrapper abstraction is also what allows PICID to compare deep sequence models, tabular learners, and statistical baselines within one protocol. By decoupling model internals from the evaluation interface, the framework can accommodate fundamentally different fitting paradigms without weakening the shared benchmark boundaries. This is especially important for tabular or fit-predict pathways, which would otherwise require custom ad hoc evaluation pipelines.

\subsection{Evaluation and reporting}
\label{app:lib_evaluation}

\emph{Protocol role.}\; $\{(\hat{y}_k,y_k)\} \xrightarrow{\ell} \mathcal{L}_{\mathrm{test}}$, task-specific metrics.

The evaluator subsystem turns standardized $(\hat{y}_k,y_k)$ pairs into benchmark metrics. Concretely, it:
\begin{itemize}
    \item computes task-specific metrics for regression, classification, fault detection, and other supported settings;
    \item optionally applies inverse scaling so metrics are reported in physical units; and
    \item exposes a hook mechanism for side effects such as saving predictions, logging trend plots, or exporting visualizations without modifying the core metric logic.
\end{itemize}
In mathematical terms, this module is where the per-sample scoring function $\ell$ and the aggregate test metric $\mathcal{L}_{\mathrm{test}}$ are enforced on standardized $(\hat{y}_k,y_k)$ pairs.

This layer is also where PICID's stronger software-assurance features become visible. Run artifacts---including resolved configurations, checkpoint metadata, and reproducibility manifests---are persisted together with the metrics so that any reported result can be audited and replayed. In combination with deterministic seeding, cache-aware preprocessing, and the large automated test suite described in the main paper, the evaluator/reporting layer turns the benchmark protocol into traceable research software rather than a collection of scripts. Appendix~\ref{app:reproducibility} documents the full run-reproduction pathway.

\subsection{Public extension interfaces}
\label{app:lib_interfaces}

\emph{Protocol role.}\; Protocol-preserving extensibility for datasources, tasks, models, and evaluators.

Beyond the internal execution modules, \picid{} exposes a public interface layer that lets users extend the framework without editing the orchestration core. The high-level entry surface resolves configuration resources, instantiates datasources, and executes preprocessing. Schema objects define task definitions, model configurations, and evaluator configurations in a typed form, while wrapper-facing interfaces let custom backbones and trainers join the standard execution path. This extension layer matters for the paper's extensibility claim because it shows that PICID is not only a fixed benchmark bundle, but a library with stable public surfaces for adding new research components while preserving the benchmark invariants defined in Sections~\ref{sec:framework_formalization} and~\ref{sec:picid_library}.

In practice, this interface layer answers four reader-facing extension questions: how to add a datasource, how to define a task, how to plug in a model, and how to adapt evaluation/reporting. The design keeps those extension points public and typed, while leaving the invariant parts of the protocol---splits, alignment, windowing, and evaluation boundaries---owned by the framework.

\section{Transform system: configuration, inventory, and pipeline}
\label{app:yaml_example}

This appendix provides a detailed reference for the \picid{} transform system. Section~\ref{app:transform_walkthrough} walks through a representative configuration with commentary on the data-routing semantics. Section~\ref{app:transform_inventory} catalogues the available transforms. Section~\ref{app:preprocessing_pipeline} presents pseudocode for the preprocessing pipeline, including the three-tier caching strategy.

\subsection{Transform configuration walkthrough}
\label{app:transform_walkthrough}

Each transform in a \picid{} pipeline is specified as a YAML block with two sections: a \emph{transform} block that names the Python class (via Hydra's \texttt{\_target\_} field) and its hyperparameters, and a \emph{metadata} block that controls data routing. Listing~\ref{lst:transform_yaml} shows a minimal three-stage pipeline illustrating train-only fitting and key routing. The stages execute in declaration order; each stage's output becomes available to subsequent stages via key assignment.

\begin{lstlisting}[style=yamlstyle,caption={Three-stage transform configuration illustrating input routing (\texttt{apply\_to}), output routing (\texttt{assign\_to}), and fitting policy (\texttt{fit\_on}).},label={lst:transform_yaml}]
scaler_features:
  transform:
    _target_: picid.transforms.base_transforms.scaler.MinMaxScalerSklearn
  metadata:
    apply_to: features
    fit_on: train
    assign_to: features_scaled

scaler_target:
  transform:
    _target_: picid.transforms.base_transforms.scaler.MinMaxScalerSklearn
  metadata:
    apply_to: target
    fit_on: train
    assign_to: target_scaled

concatenate_features:
  transform:
    _target_: picid.transforms.base_transforms.concatenate.ConcatenateTransform
  metadata:
    apply_to: [features_scaled, target_scaled]
    assign_to: features
\end{lstlisting}

\paragraph{Stage-by-stage commentary.}
\begin{enumerate}[leftmargin=*]
    \item \texttt{scaler\_features} reads the \texttt{features} key from the container, fits a min-max scaler on the \emph{training split only} (\texttt{fit\_on:~train}), and writes the scaled output to a new key \texttt{features\_scaled} via \texttt{assign\_to}.
    \item \texttt{scaler\_target} applies the same pattern to the \texttt{target} key, writing to \texttt{target\_scaled}. Keeping feature and target transforms separate is important for evaluators that later apply inverse scaling to report metrics in physical units.
    \item \texttt{concatenate\_features} illustrates multi-input routing by merging \texttt{features\_scaled} and \texttt{target\_scaled} into a single \texttt{features} key. In a task pipeline, the second input is typically a derived covariate (e.g., descriptors) rather than the supervisory target; the routing semantics are the same.
\end{enumerate}

\paragraph{Key metadata fields.} The metadata block supports the following fields:
\begin{itemize}[leftmargin=*, nosep]
    \item \texttt{apply\_to} --- which container key(s) the transform reads. Can be a single string or a list (for multi-input transforms like concatenation).
    \item \texttt{assign\_to} --- the container key to write the output to. If omitted, the output overwrites the input key.
    \item \texttt{fit\_on} --- which split to use for parameter estimation (typically \texttt{train}). If omitted, the transform is stateless.
    \item \texttt{transform\_on\_keys} --- optionally restrict application to a subset of splits (e.g., \texttt{[train, val, test]}).
    \item \texttt{validate\_output} --- boolean flag to enable or disable post-transform integrity checks (default \texttt{true}).
\end{itemize}

\subsection{Transform inventory}
\label{app:transform_inventory}

Table~\ref{tab:transform_inventory} catalogues the transforms currently available in the \picid{} library. Each transform inherits from a base class (\texttt{DenseTransform}, \texttt{RaggedTransform}, or \texttt{RaggedOrDenseTransform}) and composes a fit-strategy mixin that determines how parameters are estimated across multi-unit data. The per-transform definitions that follow the table state each operator in the notation of Section~\ref{sec:framework_formalization} and Appendix~\ref{app:complete_formalization}, combining a conceptual explanation, benchmark-level operator definitions, and explicit support or dimensional consequences; selected window-aware transforms additionally include mini worked derivations anchored to Figure~\ref{fig:formalization_windowing_operator}, which supplies the shared support, extracted-window, and label-alignment context. These derivations are intentionally more explanatory than the rest of H.2, so that the reader can see exactly how each operator acts on an already displayed benchmark object.

\begin{table}[H]
\centering
\footnotesize
\caption{Transform inventory. \emph{Fit strategy}: CF = concatenate-fit (global statistics across units), NF = no-fit (stateless). Transforms marked with $\dagger$ support inverse transformation for evaluator-level metric de-scaling.}
\label{tab:transform_inventory}

\renewcommand{\arraystretch}{1.12}
\begin{adjustbox}{max width=\linewidth}
\begin{tabular}{@{}llll@{}}
\toprule
\textbf{Transform} & \textbf{Category} & \textbf{Description} & \textbf{Fit} \\
\midrule
\multicolumn{4}{l}{\emph{Scaling}} \\
MinMaxScalerSklearn$^\dagger$ & Scaling & Min-max normalization (scikit-learn adapter) & CF \\
StandardScalerSklearn$^\dagger$ & Scaling & Zero-mean, unit-variance standardization & CF \\
ConstantScaler$^\dagger$ & Scaling & Multiply by a constant factor & NF \\
\midrule
\multicolumn{4}{l}{\emph{Spectral \& statistical}} \\
STFTTransform & Spectral & Short-time Fourier transform with configurable windows & NF \\
SpectralStatsTransform & Spectral & Frequency-domain statistics via FFT & NF \\
TimeStatsTransform & Statistical & Time-domain statistics (mean, kurtosis, peak factor, etc.) & NF \\
WindowedAggregationTransform & Statistical & Window-based aggregation (mean, sum, min, max, last, etc.) & NF \\
\midrule
\multicolumn{4}{l}{\emph{Signal processing}} \\
CumSumTransform & Signal processing & Cumulative squared sum after ragged flattening, then unflattened back & NF \\
\midrule
\multicolumn{4}{l}{\emph{Structural}} \\
ConcatenateTransform & Structural & Concatenate multiple input arrays along a specified axis & NF \\
ReshapeTransform & Structural & Reshape arrays using einops patterns & NF \\
SubsampleTransform & Structural & Subsample data at fixed intervals & NF \\
PadToLength & Structural & Pad arrays to a target length along a specified axis & NF \\
RegularizeRaggedDataTransform & Structural & Convert a ragged axis to a regular Awkward array & NF \\
RaggedToDenseTransform & Structural & Convert ragged (Awkward) arrays to dense NumPy arrays & NF \\
ExpandScalarToReferenceFeatureSize & Structural & Expand scalar/array to match reference feature dimensions & NF \\
\midrule
\multicolumn{4}{l}{\emph{Imputation \& corruption}} \\
ImputationTransform & Imputation & Fill NaNs via zero/mean/LOCF/linear/spectral/stochastic/pattern-based repair & CF \\
MCARCorruptorTransform & Corruption & Inject missing-completely-at-random faults (point or block) & NF \\
\midrule
\multicolumn{4}{l}{\emph{Tabularization}} \\
TimeseriesTabularizer & Tabular & Tabularize time-series with windowing, padding, and subsetting & NF \\
\midrule
\multicolumn{4}{l}{\emph{Domain-specific}} \\
BatteryTransform & Battery & Append shifted target; polynomial/rolling helper routines are defined but inactive & NF \\
Sequence2Statistics & Battery & Aggregate selected sequence axes into mean/std summary vectors & NF \\
HealthIndexTransform$^\dagger$ & Bearings & Compute health indicator from bearing RUL lookups & NF \\
N\_CMAPSSFeaturesScaler & N-CMAPSS & Scale N-CMAPSS sensor features using fixed dataset statistics & NF \\
N\_CMAPSSDescriptorsScaler & N-CMAPSS & Scale N-CMAPSS operating-condition descriptors & NF \\
ConceptClassesBuilder & N-CMAPSS & Map concept indicators and dataset IDs to concept-class labels & CF \\
MinMaxScalerMZVAV$^\dagger$ & Building & Min-max scaler for MZVAV building data & CF \\
\midrule
\multicolumn{4}{l}{\emph{Analytics \& debugging}} \\
MissingValuesStatsLogger & Analytics & Log missing-value statistics (pass-through, no data modification) & NF \\
IdentityPassThrough & Utility & No-op pass-through transform & NF \\
\bottomrule
\end{tabular}
\end{adjustbox}
\end{table}

The grouped reference below is meant to be read in two layers. First, each entry states the operator using the notation of Section~\ref{sec:framework_formalization} and Appendix~\ref{app:complete_formalization}. Second, the worked examples use the window supports shown in Figure~\ref{fig:formalization_windowing_operator} to illustrate what the operator changes on the feature/target side and, crucially, how the window label is assigned (e.g., end-of-window versus offset supervision).

\subsubsection{Scaling}

\noindent Scaling transforms are pointwise numerical reparameterizations of already defined feature or target channels. When fitted, their parameters are estimated on the training partition only and stored in $\Psi$ for feature-side stages or $\Phi$ for target-side stages; when inverse maps exist, predictions can be returned to the original physical units. Unless noted otherwise, these operators preserve temporal support, so $T'=T$ and $a(j)=j$.

\paragraph{MinMaxScalerSklearn.}
This transform rescales each selected channel to a common numerical interval using channelwise minimum and maximum values estimated from the training partition. When instantiated as a feature-side stage in $\mathcal{G}(\cdot;\Psi)$, the fitted state is $\psi_i=\{m_q^{\min},m_q^{\max}\}_{q=1}^{d_{\mathrm{in}}}\subset\Psi$; when the same pointwise map is instantiated in $\mathcal{H}(\cdot;\Phi)$ for targets, the corresponding target-side state is $\phi_i=\{m_q^{\min},m_q^{\max}\}_{q=1}^{d_{\mathrm{in}}}\subset\Phi$. With the default sklearn \texttt{feature\_range}$=[0,1]$, the forward and inverse maps are
\[
z_q(j)=\frac{x_q(j)-m_q^{\min}}{m_q^{\max}-m_q^{\min}},\qquad
x_q(j)=z_q(j)\big(m_q^{\max}-m_q^{\min}\big)+m_q^{\min}.
\]
For a non-default range $[a,b]$, the forward map becomes $z_q(j)=a+(b-a)\,(x_q(j)-m_q^{\min})/(m_q^{\max}-m_q^{\min})$. It is pointwise, so the transformed index set is unchanged: $T'=T$ and $a(j)=j$.

\paragraph{StandardScalerSklearn.}
This transform centers and rescales each selected channel using training-partition means and standard deviations so that transformed coordinates have comparable scale. When instantiated in $\mathcal{G}(\cdot;\Psi)$, the fitted parameters are $\psi_i=\{(\mu_q,\sigma_q)\}_{q=1}^{d_{\mathrm{in}}}\subset\Psi$; when instantiated in $\mathcal{H}(\cdot;\Phi)$ for targets, the corresponding target-side state is $\phi_i=\{(\mu_q,\sigma_q)\}_{q=1}^{d_{\mathrm{in}}}\subset\Phi$. The operator applies
\[
z_q(j)=\frac{x_q(j)-\mu_q}{\sigma_q},\qquad
x_q(j)=\sigma_q z_q(j)+\mu_q.
\]
As a pointwise stage, it preserves temporal resolution and support: $T'=T$ and $a(j)=j$.

\paragraph{ConstantScaler.}
This transform multiplies the selected signal by a user-specified constant and is useful when targets or auxiliary channels must be expressed in a preferred engineering scale without introducing learned state. For a fixed constant $c\neq 0$, the stage is $z_f(j)=c\,x_f(j)$, with inverse $x_f(j)=z_f(j)/c$. Since $c$ is configuration-defined rather than estimated from data, it contributes no fitted parameter to $\Psi$ or $\Phi$. Again the operator is pointwise, so $T'=T$ and $a(j)=j$.

\subsubsection{Spectral \& statistical}

\paragraph{STFTTransform.}
This transform computes a short-time Fourier transform (STFT): it replaces raw time samples with features defined on time--frequency frames. The important consequence for the benchmark is that it changes the effective temporal grid: each output index corresponds to an analysis window, not a single raw timestamp, so target alignment must follow the returned frame supports. Let $w$ be the analysis-window length, $s_{\mathcal{G}}$ the hop, and $n_{\mathrm{fft}}$ the FFT length. The implementation delegates frame placement to SciPy's \texttt{ShortTimeFFT}, so the transformed index set $\{1,\dots,T'\}$ and support map $a(j)=I_j$ are the library-returned STFT frame grid induced by $(w,s_{\mathcal{G}})$, including boundary frames when present. For feature channel $f$ and discrete frequency bin $k\in\{0,\dots,n_{\mathrm{fft}}-1\}$, the complex coefficient array has the form
\[
S_f(j,k)=\sum_{t\in I_j} x_f(t)\,u_j(t)\,
\exp\!\left(-\frac{2\pi i\,k\, t}{n_{\mathrm{fft}}}\right),
\]
where $u_j$ is the translated analysis window on frame $j$. We adopt the absolute-time phase convention (the exponent is indexed by the raw sample $t$, matching SciPy's internal bookkeeping up to a per-frame linear-phase offset $\exp(-2\pi i\,k\,t_0(j)/n_{\mathrm{fft}})$, with $t_0(j)=\min I_j$). This offset is immaterial for magnitude, power, and log-power post-maps; when $\Gamma$ retains phase, the offset is absorbed into the stated frame-origin reference. The emitted feature vector is $\vz(j)=\Gamma(\{S_f(j,k)\}_{f,k})$, where $\Gamma$ is the configured post-map (magnitude, phase, paired real--imaginary coefficients, power, log-power, or subband aggregation). Per frame, the emitted feature dimension may be written as $F'=d_{\mathrm{in}}d_{\Gamma}$, where $d_{\Gamma}$ depends on the retained frequency or subband representation (e.g., one-sided $k\in\{0,\dots,\lfloor n_{\mathrm{fft}}/2\rfloor\}$ for real-valued inputs, or a subband aggregation thereof). Because $T'$ is the number of frames returned by \texttt{ShortTimeFFT.stft} rather than a manually truncated interior-window count, aligned supervision must be defined through $\widetilde{\mathcal{H}}(\gY,a(j);\Phi)$ on that exact frame support.

\paragraph{SpectralStatsTransform.}
This transform compresses a full dense segment into frequency-domain summary features when spectral shape matters more than samplewise timing. In the current implementation, one transform call consumes one dense array $\vx\in\mathbb{R}^{T\times d_{\mathrm{in}}}$ and returns one descriptor vector, so $T'=1$ and $a(1)=\{1,\dots,T\}$. Let $A_q(\omega)$ denote the one-sided amplitude spectrum computed from the full column signal $\{x_q(t)\}_{t=1}^{T}$ for channel $q$. The output is a concatenation of configured functionals
\[
\vz(1)=\bigoplus_{q=1}^{d_{\mathrm{in}}}\bigoplus_{r\in\mathcal{R}}
\rho_r\!\left(A_q\right),
\]
where $\mathcal{R}$ may include mean, maximum, minimum, RMS, variance, skewness, kurtosis, energy, peak factor, clearance factor, change coefficient, and entropy-based descriptors. Since each configured spectral statistic contributes one scalar per channel, the emitted dimension is
\[
D_{\mathrm{spec}} = d_{\mathrm{in}}\lvert\mathcal{R}\rvert,
\]
so the class realizes a whole-segment spectral summary operator rather than a framewise feature map.

\emph{Worked example.} Figure~\ref{fig:formalization_windowing_operator} provides a concrete reference for how supports and labels are inherited. We reuse the same displayed window support (e.g., the row labeled $\vW_7$) and show what changes when a transform replaces that support-restricted feature block with a descriptor vector. The label assignment stays the same: it is inherited from the aligned target at the supervision index fixed by the benchmark rule.%
For the window-aware examples below, we keep $\mathcal{H}=\mathrm{Id}$ and $\Phi=\varnothing$, so the target pipeline specializes to $\widetilde{\mathcal{H}}(\gY,I;\Phi)=\mathcal{A}_{\mathrm{last}}(\gY,I)$ on any interval support $I$. The figure uses the default end-of-window rule with $L_{\mathrm{seq}}=4$, so the label attached to $\vW_k$ is $y_k=z_y(j_{\mathrm{sup}}(k))$ with $j_{\mathrm{sup}}(k)=k+3$.%
To form a spectral summary, take the support underlying the row labeled $\vW_7$ from the right panel; the left panel shows that this support covers the four consecutive timestamps $t_7,\dots,t_{10}$. The transform's role is to replace that ordered four-step feature block by one descriptor vector:
\begin{align*}
\mathbf{u}_7 &= [\widetilde{\vz}(7),\widetilde{\vz}(8),\widetilde{\vz}(9),\widetilde{\vz}(10)]^{\top},\\
I_7 &= \{t_7,t_8,t_9,t_{10}\},\\
A_{u_7}(\omega) &\triangleq \text{the one-sided amplitude spectrum of }\mathbf{u}_7,\\
\vz_{\mathrm{spec}}(7) &= [\rho_{\mathrm{mean}}(A_{u_7}),\rho_{\mathrm{max}}(A_{u_7})]^{\top},\\
z_y(10) &= \widetilde{\mathcal{H}}(\gY,I_7;\Phi)
= \mathcal{A}_{\mathrm{last}}(\mathcal{H}(\gY;\Phi),I_7)
= \mathcal{A}_{\mathrm{last}}(\gY,I_7),\\
y_7 &= z_y(10).
\end{align*}
No supervision logic changes here. The figure already fixes the support and the attached label, so the supervised sample is simply $(\vz_{\mathrm{spec}}(7),y_7)$. The transform only changes the feature-side representation: it replaces the support-restricted block underlying $\vW_7$ with a spectral summary vector.

\paragraph{TimeStatsTransform.}
This transform replaces a full dense segment with time-domain summary features such as mean, standard deviation, peak-to-peak range, RMS, and shape coefficients. In the current implementation, one transform call consumes one dense array $\vx\in\mathbb{R}^{T\times d_{\mathrm{in}}}$ and returns one descriptor vector, so $T'=1$ and $a(1)=\{1,\dots,T\}$. The transform applies a configured family of time-domain functionals $\eta_r$ to each full column signal. The emitted vector is
\[
\vz(1)=\bigoplus_{q=1}^{d_{\mathrm{in}}}\bigoplus_{r\in\mathcal{R}}
\eta_r\!\left(\{x_q(t)\}_{t=1}^{T}\right),
\]
where $\mathcal{R}$ may include mean, maximum, minimum, RMS, absolute average, variance, standard deviation, skewness, kurtosis, absolute energy, peak factor, change coefficient, clearance factor, and the optional singular values of a Hankel embedding. If each statistic $r$ contributes $d_r$ coordinates, with $d_r=1$ for scalar descriptors and $d_r>1$ for vector-valued ones such as \texttt{hankel\_svd}, then the emitted dimension is
\[
D_{\mathrm{time}} = d_{\mathrm{in}}\sum_{r\in\mathcal{R}} d_r.
\]
Thus the transform collapses an entire selected segment to one time-domain descriptor point.

\emph{Worked example.} Reuse the same displayed support underlying $\vW_7$ so that the reader can compare two different summary families on exactly the same local transformed-feature block. The input support and the attached label therefore stay fixed, and only the feature-side descriptor changes:
\begin{align*}
\mathbf{u}_7 &= [\widetilde{\vz}(7),\widetilde{\vz}(8),\widetilde{\vz}(9),\widetilde{\vz}(10)]^{\top},\\
I_7 &= \{t_7,t_8,t_9,t_{10}\},\\
\mu_{W_7} &= \frac{1}{4}\sum_{i=7}^{10}\widetilde{z}(i),\\
\sigma_{W_7}^2 &= \frac{1}{4}\sum_{i=7}^{10}\big(\widetilde{z}(i)-\mu_{W_7}\big)^2,
\qquad
\sigma_{W_7} = \sqrt{\sigma_{W_7}^2},\\
r_{W_7} &= \max_{i=7,\dots,10}\widetilde{z}(i)-\min_{i=7,\dots,10}\widetilde{z}(i),\\
\vz_{\mathrm{time}}(7) &= [\mu_{W_7},\sigma_{W_7},r_{W_7}]^{\top},\\
z_y(10) &= \widetilde{\mathcal{H}}(\gY,I_7;\Phi)
= \mathcal{A}_{\mathrm{last}}(\mathcal{H}(\gY;\Phi),I_7)
= \mathcal{A}_{\mathrm{last}}(\gY,I_7),\\
y_7 &= z_y(10).
\end{align*}
The contrast with the spectral case is now controlled and explicit: the same four transformed feature entries and the same aligned target yield the supervised sample $(\vz_{\mathrm{time}}(7),y_7)$, but the emitted feature object is now a time-domain summary vector rather than a spectral one.

\paragraph{WindowedAggregationTransform.}
This transform performs deterministic local pooling, such as a windowed mean, sum, minimum, maximum, median, standard deviation, first sample, or last sample. It is a resolution-changing operator whose main effect is temporal smoothing or downsampling with an explicit aggregation rule, and in practice it is often the transform that makes feature support compatible with the target-alignment rule. For aggregation rule $\alpha$, window size $w$, stride $s_{\mathcal{G}}$, and support $a(j)=I_j$, the stage computes
\[
z_f(j)=\alpha\!\left(\{x_f(t): t\in I_j\}\right),\qquad
I_j=\{(j-1)s_{\mathcal{G}}+1,\dots,(j-1)s_{\mathcal{G}}+w\}.
\]
When the aggregated axis is temporal and channel count is preserved, the output lies in $\mathbb{R}^{T'\times d_{\mathrm{in}}}$ with
\[
T'=\left\lfloor \frac{T-w}{s_{\mathcal{G}}}\right\rfloor+1.
\]
Since $\alpha$ is deterministic and stateless, no parameter enters $\Psi$, but the support map $a(j)$ must still be propagated to $\widetilde{\mathcal{H}}(\gY,a(j);\Phi)$ for label alignment.

\emph{Worked example.} Figure~\ref{fig:formalization_windowing_operator} already displays the entire extracted-window family for $L_{\mathrm{seq}}=4$ and $\Delta=3$, so here the only new ingredient is the aggregation rule. Choose $\alpha=\mathrm{mean}$ and index the aggregated outputs by the displayed window starts $k\in\{1,4,7,10\}$ so that every equation lines up directly with one row of the figure. Each displayed row identifies the local support block that feeds one aggregated feature value. Under the specialization above, that same support first determines the aligned sequence-level value $z_y(k+3)=\widetilde{\mathcal{H}}(\gY,\{t_k,\dots,t_{k+3}\};\Phi)$, and the corresponding window-level label is then $y_k=z_y(k+3)$:
\begin{alignat*}{3}
\mathbf{u}_1 &=
[\widetilde{\vz}(1),\widetilde{\vz}(2),\widetilde{\vz}(3),\widetilde{\vz}(4)]^{\top}, \qquad&
\bar{z}_{\mathrm{agg}}(1) &=
\frac{1}{4}\sum_{i=1}^{4}\widetilde{z}(i), \qquad&
z_y(4) &= y_1,\\
\mathbf{u}_4 &=
[\widetilde{\vz}(4),\widetilde{\vz}(5),\widetilde{\vz}(6),\widetilde{\vz}(7)]^{\top}, \qquad&
\bar{z}_{\mathrm{agg}}(4) &=
\frac{1}{4}\sum_{i=4}^{7}\widetilde{z}(i), \qquad&
z_y(7) &= y_4,\\
\mathbf{u}_7 &=
[\widetilde{\vz}(7),\widetilde{\vz}(8),\widetilde{\vz}(9),\widetilde{\vz}(10)]^{\top}, \qquad&
\bar{z}_{\mathrm{agg}}(7) &=
\frac{1}{4}\sum_{i=7}^{10}\widetilde{z}(i), \qquad&
z_y(10) &= y_7,\\
\mathbf{u}_{10} &=
[\widetilde{\vz}(10),\widetilde{\vz}(11),\widetilde{\vz}(12),\widetilde{\vz}(13)]^{\top}, \qquad&
\bar{z}_{\mathrm{agg}}(10) &=
\frac{1}{4}\sum_{i=10}^{13}\widetilde{z}(i), \qquad&
z_y(13) &= y_{10}.
\end{alignat*}
The resulting supervised samples in this start-indexed view may be written as
\begin{align*}
\mathcal{D}_{\mathrm{agg}}^{\mathrm{toy}}
=
\{(\bar{z}_{\mathrm{agg}}(1),y_1),(\bar{z}_{\mathrm{agg}}(4),y_4),\\
&\qquad (\bar{z}_{\mathrm{agg}}(7),y_7),(\bar{z}_{\mathrm{agg}}(10),y_{10})\}.
\end{align*}
In the generic notation above, these four displayed starts correspond to the emitted indices $j=1,\dots,4$ and their support windows $a(j)$. The support map is thus not an abstract bookkeeping device: it is exactly the mechanism that tells us which raw-time evidence and which window-level target belong together.

\subsubsection{Signal processing}

\paragraph{CumSumTransform.}
This transform squares the selected ragged signal, flattens away one ragged level, applies a cumulative sum along the remaining last axis, and then restores the original outer partition. In other words, it rewrites each flattened row into cumulative squared coordinates rather than integrating along the original time axis. Let $\bar{\vx}^{(\ell)}=(\bar{x}^{(\ell)}_1,\dots,\bar{x}^{(\ell)}_{F_\ell})$ denote the $\ell$-th row after the call to \texttt{ak.flatten}. The implementation computes
\[
\bar{z}^{(\ell)}_q=\sum_{p=1}^{q}\big(\bar{x}^{(\ell)}_p\big)^2,\qquad q=1,\dots,F_\ell,
\]
and then restores the original outer ragged partition with \texttt{ak.unflatten(..., ak.num(x))}. No fitted parameter is introduced. The crucial implementation detail is that the cumulative sum runs along the last axis of the flattened rows, not along the original unit-time axis directly. A likely literature antecedent in bearing prognostics should be inserted here once confirmed: \textcolor{green}{<reference>}.

\subsubsection{Structural}

\noindent Structural transforms primarily alter layout, densification, broadcast structure, or temporal indexing rather than learned signal content. They should therefore be read as representation operators on already defined benchmark objects: some preserve support exactly, whereas others induce a new support map through decimation or padding.

\paragraph{ConcatenateTransform.}
This transform fuses several aligned arrays into a single tensor by concatenating them along a configured axis after coercing lower-rank inputs to compatible shapes. Let $\{\mathbf{u}^{(m)}(j)\}_{m=1}^{N_{\mathrm{src}}}$ denote the aligned sources at transformed index $j$. The emitted array is
\[
\vz(j)=\operatorname{concat}_{d}\!\big(\mathbf{u}^{(1)}(j),\dots,\mathbf{u}^{(N_{\mathrm{src}})}(j)\big),
\]
where $d$ is the configured concatenation axis after shape coercion to a common rank. If $d$ indexes channels, the support map and transformed length are unchanged and only feature width grows; if $d$ indexes time, the support at each emitted position is the ordered union of the contributing supports, so downstream target alignment must reference that combined interval.

\paragraph{ReshapeTransform.}
This transform re-expresses an array through an \texttt{einops}-style rearrangement pattern. The operator is a deterministic bijection $R_{\pi}$ induced by the configured pattern $\pi$, so $\vz=R_{\pi}(\vx)$ with identical entries under a new indexing convention. When $\pi$ only permutes or refactors feature axes, support and supervision alignment are unchanged; the transform changes tensor factorization, not numerical content, and introduces no fitted state.

\paragraph{SubsampleTransform.}
This transform retains every $r$-th entry along the leading axis of each selected array and discards the intermediate entries. For stride $r\in\mathbb{N}$, the implementation applies leading-axis slicing,
\[
\vz(j)=\vx\big(1+(j-1)r\big),\qquad
j=1,\dots,\left\lfloor \frac{T-1}{r}\right\rfloor+1.
\]
When the leading axis is temporal, this is explicit rate reduction with $T'=\lfloor (T-1)/r\rfloor+1$ and $a(j)=1+(j-1)r$, where $a(j)$ is a scalar raw-time index rather than an interval --- admissible under the ``timestamp or interval'' convention of Section~\ref{app:feature_transform_full}; otherwise it is simply leading-axis decimation. The transform is stateless, but temporal use requires target alignment only at the retained indices.

\paragraph{PadToLength.}
This transform pads the configured axis of \texttt{data["features"]} to a prescribed length by appending a fixed pad value. For target length $L_{\mathrm{pad}}$ and pad value $p$, the output along the padded axis is
\[
z(j)=
\begin{cases}
x(j), & 1\le j\le T,\\
p, & T<j\le L_{\mathrm{pad}}.
\end{cases}
\]
Observed entries are untouched, while only synthetic trailing values are added when the input is too short. If the padded axis is temporal, the transformed length becomes $T'=L_{\mathrm{pad}}$; the original support is preserved on $j\le T$, whereas padded indices correspond to artificial support used only for shape compatibility.

\paragraph{RegularizeRaggedDataTransform.}
This transform converts an Awkward ragged axis into a regular tensor-like axis when all realized segment lengths already agree. Let $\mathfrak{X}$ denote the ragged representation of such a collection. The operator is the identity on observed entries and ordering,
\[
\operatorname{Reg}(\mathfrak{X}) \equiv \mathfrak{X},
\]
but changes the container type from ragged to regular. It is therefore a shape-regularization step for later dense operators, not a numerical transform, and it leaves support unchanged.

\paragraph{RaggedToDenseTransform.}
This transform materializes a ragged array as a dense NumPy array once the lengths are regular enough to admit such a representation. If $\mathfrak{X}$ is a regularizable ragged container, the stage applies a densification map $\operatorname{Dense}(\mathfrak{X})$ that preserves entrywise values and ordering while replacing the storage layout by a dense tensor. The operator is numerically identity, introduces no fitted state, and changes only the class of admissible downstream kernels.

\paragraph{ExpandScalarToReferenceFeatureSize.}
This transform broadcasts a scalar or low-dimensional side channel so that it matches the first dimension of a reference feature block. Let $s(j)\in\mathbb{R}^{d_s}$ be the selected scalar or descriptor and let the reference feature width be $d_{\mathrm{ref}}$. For $d_s=1$, the emitted repeated side information is the $d_{\mathrm{ref}}$-vector filled with the scalar value,
\[
\bar{s}(j)=s(j)\,\mathbf{1}_{d_{\mathrm{ref}}}\in\mathbb{R}^{d_{\mathrm{ref}}};
\]
for $d_s>1$, each coordinate is independently broadcast along the reference axis, yielding the tiled block
\[
\bar{s}(j)=\mathbf{1}_{d_{\mathrm{ref}}}\,s(j)^{\top}\in\mathbb{R}^{d_{\mathrm{ref}}\times d_s},
\]
so that the result matches the reference axis and can be concatenated with other features along the channel axis. The operator is deterministic and stateless: feature width grows by repetition, while support and target alignment are preserved.

\subsubsection{Imputation \& corruption}

\paragraph{ImputationTransform.}
This transform fills missing values while keeping the original tensor layout, temporal support, and feature dimension intact. At the benchmark level it is an in-place repair operator on the transformed feature sequence: if $\Omega=\{(j,f):x_f(j)\ \text{is observed}\}$ denotes the observed set and $\Omega^c$ its missing complement, then
\[
\hat{x}_f(j)=
\begin{cases}
x_f(j), & (j,f)\in\Omega,\\
m_f(j;\psi_i), & (j,f)\in\Omega^c.
\end{cases}
\]
The configured strategy specializes the repair map $m_f$. For \texttt{mean} and \texttt{stochastic} modes, $\psi_i\subset\Psi$ contains train-only per-channel statistics such as fitted means and standard deviations. For \texttt{zero}, $m_f(j;\psi_i)\equiv 0$. For \texttt{locf}, \texttt{linear}, \texttt{spectral}, and \texttt{copy\_past}, $m_f$ is a local deterministic or stochastic reconstruction rule built from neighboring or historical observations. Stochastic modes (\texttt{stochastic} and any sampling-based branch of \texttt{copy\_past}) are seeded from a configuration-supplied RNG seed stored in $\psi_i$, so that the full transform $(g_i,\psi_i)$ remains a deterministic function of its inputs and participates unambiguously in the pipeline cache key of Algorithm~\ref{alg:preprocessor}. These implementation refinements change only the repaired values on $\Omega^c$; the support map $a(j)$ and target alignment remain unchanged.

\emph{Worked example.} On the shared toy feature trace, suppose $x(t_9)$ is missing while the neighboring values on $J=\{t_7,\dots,t_{11}\}$ are observed. Under \texttt{mean} imputation, $\hat{x}(t_9)=\bar{x}_{\mathrm{train}}$ for the fitted training-partition mean, whereas under \texttt{locf}, $\hat{x}(t_9)=x(t_8)$. In both cases the repaired interval is still indexed by $J$, so the aligned target under $\mathcal{H}=\mathrm{Id}$ and $\mathcal{A}_{\mathrm{last}}$ remains $z_y(1)=y(t_{11})$: imputation changes values, not support or supervision semantics.

\paragraph{MCARCorruptorTransform.}
This transform injects artificial missingness by replacing selected values with \texttt{NaN}. For each channel $f$, let $r_f\in[0,1]$ be the configured corruption ratio. In \texttt{point} mode the mask is sampled independently,
\[
M_f(j)\sim\operatorname{Bernoulli}(1-r_f),\qquad
z_f(j)=
\begin{cases}
x_f(j), & M_f(j)=1,\\
\mathrm{NaN}, & M_f(j)=0.
\end{cases}
\]
In \texttt{block} mode the zero entries of $M_f$ are generated as contiguous outages with lengths $L_b\sim\mathrm{Uniform}\{L_{\min},\dots,L_{\max}\}$ placed at uniformly sampled starts, without overlap, until the cumulative number of masked samples first reaches or exceeds $r_f T$; the realized corruption ratio therefore satisfies $|\{j:M_f(j)=0\}|/T\in[r_f,\,r_f+(L_{\max}-1)/T]$, so the target rate $r_f$ is achieved up to an additive slack of at most one block length. The benchmark object is unchanged except for its observed mask: temporal support, feature dimension, and target semantics remain fixed, while only the pattern of missing entries changes.

\subsubsection{Tabularization}

\paragraph{TimeseriesTabularizer.}
This transform converts sequential windows into the tabular representations expected by static learners and tabular foundation models. In the notation of Appendix~\ref{app:tabularization_full}, the canonical benchmark tabular adapter is
\[
\vX_k=\mathcal{T}(\vW_k)\in\mathbb{R}^{D_{\mathrm{tab}}},\qquad
D_{\mathrm{tab}}=L_{\mathrm{seq}}F.
\]
Using the time-major ordering fixed in Appendix~\ref{app:tabularization_full}, this means
\[
\vX_k
=
\big[z_1(k),\dots,z_F(k),\;z_1(k+1),\dots,z_F(k+1),\;\dots,\;z_1(k+L_{\mathrm{seq}}-1),\dots,z_F(k+L_{\mathrm{seq}}-1)\big]^{\top}.
\]
One transformed window $\vW_k$ therefore yields one tabular sample $\vX_k$ without changing split membership or benchmark supervision semantics. In software, the same class can either emit this canonical history-window flattening directly or concatenate it with other already aligned window-local fields selected by configuration, including additional feature keys, target-history slices, present-time descriptors, or horizon-specific side information. Those refinements enlarge the emitted software sample, but the benchmark tabularization map itself remains the history-window flattening above.

\emph{Worked example.} Reuse the row labeled $\vW_7$ from Figure~\ref{fig:formalization_windowing_operator}. The right panel already shows it as an ordered transformed-feature window, and tabularization simply rewrites that same object as one fixed-order sample for a static learner:
\[
\vW_7=
\begin{bmatrix}
\widetilde{z}(7)\\
\widetilde{z}(8)\\
\widetilde{z}(9)\\
\widetilde{z}(10)
\end{bmatrix}\in\mathbb{R}^{4\times 1}.
\]
The benchmark adapter therefore yields
\[
\vX_7=\mathcal{T}(\vW_7)=
\big[\widetilde{z}(7),\widetilde{z}(8),\widetilde{z}(9),\widetilde{z}(10)\big]^{\top}.
\]
In this single-channel toy case the coordinates look numerically unchanged, but the object type has changed in an important way: $\vW_7$ is still a sequential window with temporal structure, whereas $\vX_7$ is now a tabular sample in $\mathbb{R}^{4}$ that a static learner can consume directly. The attached supervision is still inherited from the same alignment chain,
\[
z_y(10)=\widetilde{\mathcal{H}}(\gY,\{t_7,\dots,t_{10}\};\Phi)
= \mathcal{A}_{\mathrm{last}}(\mathcal{H}(\gY;\Phi),\{t_7,\dots,t_{10}\}),
\qquad y_7=z_y(10),
\]
so the benchmark sample is $(\vX_7,y_7)$. If a configuration additionally appends aligned side information, that refinement enlarges the software representation without changing the benchmark window or its supervision semantics.

\subsubsection{Domain-specific}

\paragraph{BatteryTransform.}
This transform augments the feature stream with a causally shifted copy of the target, thereby exposing recent target history as an input covariate. Let $s_{\mathrm{lag}}$ denote the configured shift and let $y(j)$ be the target side channel made available to the transform. The emitted augmented feature is
\[
\vz(j)=\big[\vx(j);\tilde{y}(j)\big],\qquad
\tilde{y}(j)=
\begin{cases}
0, & j\le s_{\mathrm{lag}},\\
y(j-s_{\mathrm{lag}}), & j>s_{\mathrm{lag}}.
\end{cases}
\]
The operator is therefore causal, pointwise in the transformed index, and preserves temporal support while increasing feature width by the shifted-target channel. In the current implementation, helper routines for polynomial and rolling summaries are defined but do not contribute to $\vz(j)$, so the active class behavior reduces exactly to this benchmark augmentation map.

\paragraph{Sequence2Statistics.}
This transform replaces the selected sequence block by concatenated mean and standard-deviation summaries computed over axes $1$ and $2$ of the working tensor, optionally ignoring a designated padding value. Let $\mathbf{X}$ denote the working tensor after padding-value masking and any required promotion of a 2D input to higher rank. If the original input has shape $(T,F)$, the class promotes it to $\mathbf{X}\in\mathbb{R}^{1\times 1\times T\times F}$; if the original input has shape $(N,T,F)$, it keeps $\mathbf{X}\in\mathbb{R}^{N\times T\times F}$. The transform then computes
\[
\mu=\operatorname{nanmean}(\mathbf{X},\mathrm{axes}=(1,2)),\qquad
\sigma=\operatorname{nanstd}(\mathbf{X},\mathrm{axes}=(1,2)),
\qquad \text{(axes are 0-indexed, NumPy convention)}
\]
replaces NaNs in $\sigma$ by zero, and returns the one-dimensional vector
\[
\vz=\big[\mu^\top;\sigma^\top\big].
\]
Thus the class realizes a sequence-to-summary operator whose coordinates depend on the input regime: a 2D input yields per-channel means and standard deviations, whereas a 3D input yields one mean and one standard deviation for each leading slice.

\emph{Worked example.} Let the selected 2D block be
\[
\mathbf{X}_0=
\begin{bmatrix}
x_1(t_7) & x_2(t_7)\\
\vdots & \vdots\\
x_1(t_{11}) & x_2(t_{11})
\end{bmatrix}\in\mathbb{R}^{5\times 2}.
\]
After promotion to the working tensor shape $1\times 1\times 5\times 2$, the transform returns
\[
\mu=
\begin{bmatrix}
\frac{1}{5}\sum_{i=7}^{11}x_1(t_i)\\[2pt]
\frac{1}{5}\sum_{i=7}^{11}x_2(t_i)
\end{bmatrix},
\qquad
\sigma=
\begin{bmatrix}
\operatorname{std}\!\big(x_1(t_7{:}t_{11})\big)\\[2pt]
\operatorname{std}\!\big(x_2(t_7{:}t_{11})\big)
\end{bmatrix},
\qquad
\vz=\big[\mu^\top;\sigma^\top\big].
\]
The original $5\times 2$ sequence is therefore collapsed to a four-coordinate summary vector; if the input were already 3D, the same formula would instead summarize each leading slice across axes $1$ and $2$.

\paragraph{HealthIndexTransform.}
This transform maps a unit-specific runtime-like channel to a normalized health index using dataset-level lifetime lookups. Let $u$ denote the current unit and let $L(u)>0$ be its configured total-life lookup value. Applied to the selected scalar channel $r(j)$, the transform computes
\[
z(j)=\frac{r(j)}{L(u)},\qquad
r(j)=L(u)\,z(j).
\]
The constants $L(u)$ are fixed dataset metadata rather than training-estimated parameters, so they do not belong to $\Psi$ or $\Phi$. Under the admissibility assumption $0\le r(j)\le L(u)$, the transform takes values in the unit interval, $z(j)\in[0,1]$, consistent with the monotone health-indicator codomain introduced in Section~\ref{subsec:tasks}. The operator is pointwise, inverse-capable, and preserves the transformed temporal grid; the implementation additionally enforces range checks and a dataset-specific nonincrease check on samples separated by the configured decrease period.

\paragraph{N\_CMAPSSFeaturesScaler.}
This transform applies a fixed, dataset-specific normalization to the N-CMAPSS sensor channels using constants bundled with the library. Let the current feature vector have dimension $d_{\mathrm{feat}}$, and let $\mu^{\star},\sigma^{\star}\in\mathbb{R}^{d_{\mathrm{feat}}}$ or $m^{\star},M^{\star}\in\mathbb{R}^{d_{\mathrm{feat}}}$ denote the hard-coded feature statistics. The transform applies either
\[
\vz(j)=\frac{\vx(j)-\mu^{\star}}{\sigma^{\star}}
\]
for \texttt{standard} scaling or
\[
\vz(j)=2\frac{\vx(j)-m^{\star}}{M^{\star}-m^{\star}}-1
\]
for \texttt{min-max} scaling. These constants are benchmark-defined and therefore fixed a priori rather than fitted from the current training partition; only channel scale changes, while support and supervision semantics are preserved.

\paragraph{N\_CMAPSSDescriptorsScaler.}
This transform performs the same style of fixed normalization as the previous transform, but on the operating-condition descriptor channels rather than the main sensor channels. If $\mathbf{w}(j)\in\mathbb{R}^{d_{\mathrm{desc}}}$ denotes the descriptor vector at transformed index $j$, the stage applies either
\[
\tilde{\mathbf{w}}(j)=\frac{\mathbf{w}(j)-\mu_{\mathrm{desc}}^{\star}}{\sigma_{\mathrm{desc}}^{\star}}
\]
or
\[
\tilde{\mathbf{w}}(j)=2\frac{\mathbf{w}(j)-m_{\mathrm{desc}}^{\star}}{M_{\mathrm{desc}}^{\star}-m_{\mathrm{desc}}^{\star}}-1,
\]
using the hard-coded descriptor statistics bundled with the library. No fitted state is added to $\Psi$ because these constants are benchmark-defined rather than learned from the current split, and the descriptor time grid is preserved exactly.

\paragraph{ConceptClassesBuilder.}
This transform converts concept indicators into a single discrete class label and then modulates that label by the dataset identifier. During fitting, it estimates a lookup $\lambda:\mathcal{I}_{\mathrm{DS}}\rightarrow\{1,\dots,|\mathcal{I}_{\mathrm{DS}}|\}$ from the unique training-set dataset identifiers, so $\phi_i=\lambda\subset\Phi$ is train-only state. For a concept vector $\mathbf{c}(j)\in\{0,1\}^{d_{\mathrm{c}}}$ with at most one active coordinate and dataset identifier $d\in\mathcal{I}_{\mathrm{DS}}$, define
\[
\tilde{c}(j)=
\sum_{m=1}^{d_{\mathrm{c}}} m\,c_m(j)\in\{1,\dots,d_{\mathrm{c}}\},
\qquad
z_y(j)=\big(\lambda(d)-1\big)\,d_{\mathrm{c}}+\tilde{c}(j)\in\{1,\dots,|\mathcal{I}_{\mathrm{DS}}|\,d_{\mathrm{c}}\}.
\]
The encoding is injective on $(\lambda(d),\tilde{c}(j))$, so each (dataset, concept) pair maps to a unique class code; distinct pairs never collide. The temporal index is preserved, but target semantics are remapped from multi-hot indicators to a scalar class code. The implementation rounds numerically non-binary concept values before this mapping and rejects rows with multiple simultaneously active concepts.

\emph{Worked example.} Suppose $\mathbf{c}(j)=[0,1,0]^{\top}$ (so $d_{\mathrm{c}}=3$) and the dataset identifier $d$ has fitted lookup value $\lambda(d)=2$. Then
\[
\tilde{c}(j)=1\cdot 0+2\cdot 1+3\cdot 0=2,
\qquad
z_y(j)=(\lambda(d)-1)\,d_{\mathrm{c}}+\tilde{c}(j)=(2-1)\cdot 3+2=5.
\]
The transform therefore leaves the temporal support unchanged while replacing a one-hot concept indicator with the scalar class code $5$.

\paragraph{MinMaxScalerMZVAV.}
This transform is the MZVAV-specific min--max scaler used for building data. With per-channel training extrema $\psi_i=\{m_f^{\min},m_f^{\max}\}_{f=1}^{d_{\mathrm{in}}}\subset\Psi$ estimated from the training partition by the internal sklearn scaler, the operator applies the same pointwise map as the generic min--max adapter,
\[
z_f(j)=\frac{x_f(j)-m_f^{\min}}{m_f^{\max}-m_f^{\min}},\qquad
x_f(j)=z_f(j)\big(m_f^{\max}-m_f^{\min}\big)+m_f^{\min}.
\]
The distinction from \texttt{MinMaxScalerSklearn} is representational rather than mathematical: the transformed output is rewrapped into the dataset's Awkward container after scaling. Temporal support is preserved exactly, and inverse transformation is available.

\subsubsection{Analytics \& debugging}

\noindent These transforms are identity operators on the benchmark tensors and differ only by auxiliary side effects such as logging, visualization, or explicit no-op staging. Their formulas are therefore best read as identity maps paired with reporting operators.

\paragraph{MissingValuesStatsLogger.}
This transform inspects missing-value patterns and writes a report to disk, but it deliberately leaves the data unchanged. As an operator on the data stream, the numerical map is identity, $g_i(\gX)=\gX$, augmented by a side effect $r_i(\gX)$ that records per-unit and global \texttt{NaN} statistics to the configured logging path. Since the benchmark tensor is unchanged, feature dimension, support, and target semantics are all preserved.

\paragraph{IdentityPassThrough.}
This transform is the explicit no-op element of the transform library. The operator is exactly the identity map,
$g_i(\gX)=\gX$ and $h_i(\gY)=\gY$, with no fitted state, no side effects, and no change to support, feature dimension, container structure, or target semantics.

\subsection{Preprocessing pipeline}
\label{app:preprocessing_pipeline}

The preprocessing orchestrator sequences the full datasource-loading, splitting, and transform pipeline as a single deterministic execution. Algorithm~\ref{alg:preprocessor} presents the pipeline logic, highlighting the three-tier caching strategy and the train-only fitting guarantee.

\begin{algorithm}[H]
\caption{Preprocessing pipeline with three-tier caching.}
\label{alg:preprocessor}
\begin{algorithmic}[1]
\Require Datasource config $C_d$, transform config $C_t$, cache paths
\Ensure Preprocessed \texttt{SplitDatasetContainer}
\State $k_{\mathrm{pre}} \gets \textsc{Hash}(C_d,\; C_t,\; \textsc{CodeFingerprint})$ \Comment{Final cache key}
\If{$\textsc{PreprocessedCache}(k_{\mathrm{pre}})$ exists}
    \State \Return $\textsc{Load}(\textsc{PreprocessedCache}(k_{\mathrm{pre}}))$ \Comment{\textbf{Tier 3}: full cache hit}
\EndIf
\State
\State $\texttt{datasource} \gets \textsc{Instantiate}(C_d)$
\State $k_{\mathrm{ds}} \gets \textsc{Hash}(\texttt{datasource}.\textsc{get\_cache\_fingerprint}(),\; \textsc{CodeFingerprint})$ \Comment{Loaded/split cache key}
\If{$\textsc{LoadedSplitCache}(k_{\mathrm{ds}})$ exists}
    \State $\texttt{container} \gets \textsc{Load}(\textsc{LoadedSplitCache}(k_{\mathrm{ds}}))$ \Comment{\textbf{Tier 1}: loaded \& split (cached)}
\Else
    \State $\texttt{datasource}.\textsc{LoadData}()$
    \State $\texttt{datasource}.\textsc{SplitData}(\textsc{SplitPolicy})$ \Comment{Apply split policy}
    \State $\texttt{container} \gets \texttt{datasource}.\textsc{GetData}()$ \Comment{\textbf{Tier 1}: loaded \& split}
    \State $\textsc{SaveLoadedSplit}(\texttt{container},\; k_{\mathrm{ds}})$
\EndIf
\State
\State $\texttt{transforms} \gets \textsc{BuildSequence}(C_t)$ \Comment{Ordered transform list}
\State $\texttt{start} \gets 0$
\For{$b$ $= |\texttt{transforms}| - 1$ \textbf{downto} $0$} \Comment{Reverse boundary search}
    \If{$\texttt{transforms}[b].\texttt{is\_cache\_point}$}
        \State $k_b \gets \textsc{Hash}(C_d,\; C_t^{\le b},\; \textsc{CodeFingerprint})$ \Comment{Prefix-dependent boundary key}
        \If{$\textsc{BoundaryCache}(k_b)$ is valid}
            \State $\texttt{container} \gets \textsc{Load}(\textsc{BoundaryCache}(k_b))$ \Comment{\textbf{Tier 2}: boundary hit}
            \State $\texttt{start} \gets b + 1$
            \State \textbf{break}
        \EndIf
    \EndIf
\EndFor
\State
\For{$i = \texttt{start}$ \textbf{to} $|\texttt{transforms}| - 1$}
    \State $g_i \gets \texttt{transforms}[i]$
    \If{$g_i.\texttt{requires\_fit}$}
        \State $g_i.\textsc{Fit}(\texttt{container}[\texttt{train}])$ \Comment{Train-only parameter estimation ($\Psi$/$\Phi$)}
    \EndIf
    \For{$s \in \{\texttt{train}, \texttt{val}, \texttt{test}\}$}
        \State $\texttt{container}[s] \gets g_i.\textsc{Apply}(\texttt{container}[s])$ \Comment{Frozen parameters}
    \EndFor
    \If{$g_i.\texttt{is\_cache\_point}$}
        \State $k_i \gets \textsc{Hash}(C_d,\; C_t^{\le i},\; \textsc{CodeFingerprint})$
        \State $\textsc{SaveBoundary}(\texttt{container},\; k_i)$ \Comment{Persist \textbf{Tier 2} checkpoint}
    \EndIf
\EndFor
\State
\State $\textsc{SavePreprocessed}(\texttt{container},\; k_{\mathrm{pre}})$ \Comment{Persist \textbf{Tier 3}}
\State \Return $\texttt{container}$
\end{algorithmic}
\end{algorithm}

\paragraph{Three-tier caching.} The caching strategy exploits the sequential structure of the transform pipeline to avoid redundant computation:
\begin{itemize}[leftmargin=*]
    \item \textbf{Tier~1} (\emph{loaded and split}): the raw datasource output after loading and splitting, before any transforms are applied. This tier is useful when the datasource involves expensive I/O (e.g., parsing large HDF5 files) but transform configurations change frequently during experimentation.
    \item \textbf{Tier~2} (\emph{boundary checkpoints}): intermediate snapshots saved after transforms marked with \texttt{cache\_point:~true} in their metadata. The orchestrator searches boundary checkpoints in \emph{reverse} order to find the most recent valid checkpoint, then re-runs only the remaining downstream transforms. This is particularly valuable when early-stage transforms are expensive (e.g., spectral analysis on high-frequency vibration data) and later stages (e.g., scaling) are being tuned.
    \item \textbf{Tier~3} (\emph{fully preprocessed}): the final output after all transforms. A cache hit at this tier skips the entire pipeline.
\end{itemize}
Cache invalidation is automatic: the cache key is a deterministic hash of the datasource configuration, the transform configuration, and a fingerprint of the relevant source code files. Any change to these inputs produces a different key, forcing re-computation from the appropriate tier. File locking ensures safe concurrent access when multiple experiment runs share a cache directory.

\section{Dataset descriptions}
\label{sec:dataset_descriptions}

This appendix documents the datasets evaluated in this study (the benchmark subset), not the full datasource inventory supported by \picid{}. We cover battery prognostics (NB14, UNIBO21), bearing prognostics (PRONOSTIA, XJTU-SY), turbofan prognostics and concept diagnostics (N-CMAPSS families), hydraulic diagnostics (HSF15), building diagnostics (MZVAV), and the PHME20 challenge prognostics task. Each subsection summarizes the source data, split regime, target construction, normalization, and evaluation protocol.

\begin{table}[!t]
\centering
\footnotesize
\caption{Dataset summary for the benchmark subset evaluated in this study.}
\label{tab:datasources_experiments}
\begin{adjustbox}{max width=\linewidth}
\begin{tabular}{@{}p{2.0cm}p{1.9cm}p{4.2cm}p{1.7cm}p{2.9cm}@{}}
\toprule
\textbf{Dataset} & \textbf{Domain} & \textbf{Info} & \textbf{Task} & \textbf{Target} \\
\midrule
NB14 & Battery & NASA randomized battery usage; multiple usage profiles/temperatures; evaluated via ah-RUL target. & Prognostics & ah-RUL (normalized remaining discharge throughput) \\
UNIBO21 & Battery & University of Bologna battery aging; diverse cycling conditions; evaluated via ah-RUL target. & Prognostics & ah-RUL (normalized remaining discharge throughput) \\
PHME20 & Turbofan & PHM Europe 2020 challenge; engine degradation trajectories under varying operating conditions. & Prognostics & RUL (time-to-failure) \\
N-CMAPSS Multi & Turbofan & Pooled multi-source subset (DS01/DS04/DS05/DS07); 60-step aggregation and shared preprocessing. & Prognostics & RUL (scaled) \\
N-CMAPSS DS02 & Turbofan & Protocol-aligned DS02 subset with fixed unit splits (flight class 2); 60-step aggregation. & Prognostics & RUL (scaled) \\
N-CMAPSS Multi & Turbofan & Same multi-source data as above, but framed as operating concept recognition after aggregation. & Diagnostics & Concept class (multi-class) \\
PRONOSTIA & Bearing & Run-to-failure accelerated bearing tests; vibration signals mapped to a health-index-derived degradation target. & Prognostics & Health-index-derived RUL \\
XJTU-SY & Bearing & Run-to-failure bearing tests with multiple conditions; vibration-based features with health-index-derived target. & Prognostics & Health-index-derived RUL \\
HSF15 & Hydraulic & Condition monitoring of a hydraulic test rig; four component tasks (accumulator/cooler/pump/valve). & Diagnostics & Component fault class (3--4 way) \\
MZVAV & Building & Simulated commercial building HVAC faults; four grouped fault categories at day level. & Diagnostics & Fault class (4-way) \\
\bottomrule
\end{tabular}
\end{adjustbox}
\end{table}

\subsection{Battery datasets}
\label{sec:battery_datasets}

We evaluate two battery prognostics families, NB14~\citep{bole2014adaptation} and UNIBO21~\citep{univbo_dataset}, using the ah-RUL target of \citet{bosello2023charge}. Both evaluations use unit-disjoint splits (cells do not cross train/val/test), min--max scaling of sensor channels and the target, and per-unit prognostics reporting on both normalized and inverse-scaled predictions. Table~\ref{tab:battery_splits} summarizes the splits.

\begin{table}[H]
\centering
\footnotesize
\caption{Train/Validation/Test splits for the battery datasets.}
\label{tab:battery_splits}

\vspace{0.5em}
\textbf{(a) NB14 --- NASA Randomized Battery dataset}
\label{tab:nasa_split}

\vspace{0.3em}
\begin{adjustbox}{width=\linewidth}
\begin{tabular}{@{}lllll@{}}
\toprule
\textbf{Group} & \textbf{Original Cells} & \textbf{Training Set} & \textbf{Validation Set} & \textbf{Test Set} \\
\midrule
Group 1 & RW1, RW2, RW7, RW8       & RW1, RW2                 & RW7  & RW8 \\
Group 2 & RW3, RW4, RW5, RW6       & RW4 \textit{(RW3 excl.)} & RW5  & RW6 \\
Group 3 & RW9, RW10, RW11, RW12    & \multicolumn{3}{c}{\textit{Excluded (unrealistic profile)}} \\
Group 4 & RW13, RW14, RW15, RW16   & RW13, RW14               & RW15 & RW16 \\
Group 5 & RW17, RW18, RW19, RW20   & RW17 \textit{(RW20 excl.)} & RW18 & RW19 \\
Group 6 & RW21, RW22, RW23, RW24   & RW21, RW22               & RW23 & RW24 \\
Group 7 & RW25, RW26, RW27, RW28   & RW25, RW26               & RW27 & RW28 \\
\midrule
\textbf{Total} & \textbf{28 Batteries} & \textbf{10 Batteries} & \textbf{6 Batteries} & \textbf{6 Batteries} \\
\bottomrule
\end{tabular}
\end{adjustbox}

\vspace{1em}
\textbf{(b) UNIBO21 --- UNIBO Powertools dataset}
\label{tab:unibo_split}

\vspace{0.3em}
\begin{adjustbox}{width=\linewidth}
\begin{tabular}{@{}lllll@{}}
\toprule
\textbf{Group} & \textbf{Original Cells} & \textbf{Training Set} & \textbf{Validation Set} & \textbf{Test Set} \\
\midrule
DM-3.0-S  & 000, 001, 002, 003            & 000, 001                & 002                  & 003 \\
DM-3.0-H  & 009, 010, 011                 & 009                     & 010                  & 011 \\
DM-3.0-P  & 013-017, (047, 049)           & 014, 015, 017           & 016                  & 013 \\
EE-2.85-S & 006, 007, 008, 042            & 007, 008                & 042                  & 006 \\
EE-2.85-H & 043, 044                      & 043                     & ---                  & 044 \\
DP-2.00-S & 018, 036-039, 050, 051, (019) & 018, 036, 037, 038, 050 & 051                  & 039 \\
DM-4.00-S & 040, 041                      & 040                     & ---                  & 041 \\
\midrule
\textbf{Total} & \textbf{27 (+3 excl.)} & \textbf{15 Batteries} & \textbf{5 Batteries} & \textbf{7 Batteries} \\
\bottomrule
\end{tabular}
\end{adjustbox}
\end{table}

\subsubsection{Target variable formulation (ah-RUL)}
\label{sec:rul_methodology}

Cycle count alone is often a poor proxy for battery degradation because degradation depends on how the cell is used (charge/discharge profiles). Following \citet{bosello2023charge}, we use ah-RUL: the normalized remaining cumulative discharge throughput (in Ampere-hours). Intuitively, ah-RUL answers ``how much discharge the cell can still deliver'' rather than ``how many cycles remain.''

The ah-RUL at cycle $n$ equals the total normalized cumulative discharge throughput at End-of-Life (EoL) cycle $n_{EoL}$ minus the cumulative discharge throughput already processed up to cycle $n$, $Q_{acc}(n)$:
\begin{equation}
Q_{RUL}(n) = Q_{acc}(n_{EoL}) - Q_{acc}(n)
\label{eq:rul_general}
\end{equation}
where $Q_{RUL}(n)$ is set to 0 for all cycles $n \geq n_{EoL}$. The cumulative discharge throughput $Q_{acc}(n)$ is computed by integrating the discharge current, $I_d(t)$, and normalizing by the cell's nominal capacity, $Q_{nom}$:
\begin{equation}
Q_{acc}(n) = \frac{1}{Q_{nom}} \sum_{i=1}^{n} \left( \int_{t_i}^{t_{i+1}} I_d(t) dt \right)
\label{eq:rul_general_acc}
\end{equation}

\subsubsection{NB14}
\label{sec:nb14_dataset}

\paragraph{Description.}
The NASA Randomized Battery Usage dataset \citep{bole2014adaptation} comprises 28 lithium cobalt oxide 18,650 cells (2.2\,Ah nominal capacity), arranged in seven groups of four cells according to operational profile and temperature. Aging was conducted with randomized current loads between 0.5\,A and 4\,A to emulate real-world usage, and characterization cycles were inserted every fifty randomized cycles.

\paragraph{Data filtering and splitting.}
Following \citet{bosello2023charge}, we filter NB14 by excluding Group~3 (RW9--RW12) due to unrealistic random-walk profiles, removing RW3 due to corrupted temperature measurements, and removing RW20 because it reports near-zero sensor values for most of its lifetime. The remaining 22 cells are split into 10 training, 6 validation, and 6 test units, with one held-out test cell per operating group (Table~\ref{tab:battery_splits}(a)).

\paragraph{Target creation.}
For NB14, end-of-life is defined as the last recorded cycle for each cell, assuming that each selected unit is run to completion. The discharge current is stored as a positive quantity, so the implementation directly integrates the positive-current segments when constructing Eq.~\eqref{eq:rul_general_acc}.

\paragraph{Normalization.}
The preprocessing pipeline min-max scales the raw channels and the ah-RUL target, computes time-domain and spectral statistics, concatenates them, and min-max rescales the concatenated descriptor block. Fit-predict models reuse the same processed features through a tabularized variant.

\paragraph{Evaluation metrics.}
NB14 is evaluated with the \texttt{per\_unit} evaluator, which logs \texttt{phm\_score}, MAE, MSE, and RMSE on inverse-scaled ah-RUL predictions.

\subsubsection{UNIBO21}
\label{sec:unibo21_dataset}

\paragraph{Description.}
The UNIBO Powertools dataset \citep{univbo_dataset} records 30 lithium-ion batteries aged under laboratory conditions representative of power-tool usage. Cells span multiple manufacturers, nominal capacities, and current protocols, leading to heterogeneous degradation trajectories that are well suited for unit-level prognostics evaluation.

\paragraph{Data filtering and splitting.}
Following \citet{bosello2023charge}, cells 019, 047, and 049 are excluded because of corrupted data or incomplete lifetimes. The remaining 27 cells are split into 15 training, 5 validation, and 7 test units, with at least one held-out unit per group whenever possible. We use the split listed in Table~\ref{tab:battery_splits}(b).

\paragraph{Target creation.}
For UNIBO21, end-of-life is the first cycle at which the measured capacity drops below the cell-specific failure threshold. Because raw discharge current is negative in this dataset, the implementation flips the sign before integrating the discharge segments in Eq.~\eqref{eq:rul_general_acc}.

\paragraph{Normalization.}
The preprocessing pipeline min-max scales the raw channels and ah-RUL target, extracts time-domain and spectral statistics, concatenates them, and applies a final min-max scaling stage. Fit-predict models reuse the same processed representation through a tabularized variant.

\paragraph{Evaluation metrics.}
UNIBO21 is evaluated with the \texttt{per\_unit} evaluator, which logs \texttt{phm\_score}, MAE, MSE, and RMSE on inverse-scaled ah-RUL predictions.

\subsection{Bearing datasets}
\label{sec:bearing_datasets}

We evaluate two bearing prognostics families, PRONOSTIA and XJTU-SY, using the same high-level protocol. We use the in-domain fold-1 split regime, standardize vibration channels, convert runtime to a normalized degradation target via a health-index transform, and report per-unit prognostics metrics. Table~\ref{tab:bearing_conditions} summarizes operating conditions.

\begin{table}[H]
\centering
\footnotesize
\caption{Operating conditions for the PRONOSTIA and XJTU-SY bearing datasets.}
\label{tab:bearing_conditions}
\begin{adjustbox}{max width=\linewidth}
\begin{tabular}{@{}llcc@{}}
\toprule
\textbf{Dataset} & \textbf{Condition} & \textbf{Speed (rpm)} & \textbf{Radial Load (N)} \\
\midrule
PRONOSTIA & Condition 1 & 1800 & 4\,000 \\
PRONOSTIA & Condition 2 & 1650 & 4\,200 \\
PRONOSTIA & Condition 3 & 1500 & 5\,000 \\
\midrule
XJTU-SY   & Condition 1 & 2100 & 12\,000 \\
XJTU-SY   & Condition 2 & 2250 & 11\,000 \\
XJTU-SY   & Condition 3 & 2400 & 10\,000 \\
\bottomrule
\end{tabular}
\end{adjustbox}
\end{table}

\subsubsection{PRONOSTIA (FEMTO-ST)}
\label{sec:pronostia_dataset}

\paragraph{Description.}
The PRONOSTIA dataset \citep{nectoux2012pronostia} is an accelerated bearing degradation benchmark with three operating conditions (Table~\ref{tab:bearing_conditions}). Each run contains horizontal and vertical acceleration measurements together with runtime metadata used for health-target construction.

\paragraph{Data filtering and splitting.}
Our evaluation does not use the original challenge learning/test split. Instead, we apply the in-domain fold-1 assignment: 9 bearings for training (\texttt{1\_4}, \texttt{1\_5}, \texttt{1\_6}, \texttt{1\_7}, \texttt{2\_4}, \texttt{2\_5}, \texttt{2\_6}, \texttt{2\_7}, \texttt{3\_1}), 3 for validation (\texttt{1\_1}, \texttt{2\_1}, \texttt{3\_2}), and 5 for test (\texttt{1\_2}, \texttt{1\_3}, \texttt{2\_2}, \texttt{2\_3}, \texttt{3\_3}).

\paragraph{Target creation.}
The transform family constructs a normalized health-indicator trajectory from runtime via \texttt{HealthIndexTransform}. In idealized form, the health indicator is
\begin{equation}
HI = 1 - \frac{\text{Runtime}}{\text{Total Lifetime}},
\label{eq:hi_target}
\end{equation}
and the transform writes this quantity as the degradation target used by the benchmark pipeline.

\paragraph{Normalization.}
The preprocessing pipeline standard-scales the raw vibration channels, computes a cumulative-sum feature, extracts time-domain and spectral statistics, and applies a final min-max scaling stage to the concatenated representation. Fit-predict models reuse the same processed features through a tabularized variant.

\paragraph{Evaluation metrics.}
PRONOSTIA is evaluated with the \texttt{per\_unit} evaluator. The logged metrics are \texttt{phm\_score}, MAE, MSE, and RMSE on inverse-scaled predictions, where \texttt{phm\_score} corresponds to the challenge-style asymmetric error measure.
\par\noindent\textbf{PHM-score.}
Following the IEEE PHM 2012 scoring convention~\citep{nectoux2012pronostia}, define the percentage error of a prediction as
\[
e \triangleq 100\,\frac{y-\hat{y}}{y+\epsilon},
\]
where $y$ is the ground-truth target, $\hat{y}$ is the prediction, and $\epsilon>0$ avoids division by zero. Early predictions ($e>0$, underestimation) are penalized less than late predictions ($e\le 0$, overestimation) via the asymmetric score
\[
A(e)=
\begin{cases}
\exp\!\left(-\ln(0.5)\,\frac{e}{5}\right), & e\le 0,\\
\exp\!\left(\ln(0.5)\,\frac{e}{20}\right), & e>0,
\end{cases}
\]
and \texttt{phm\_score} reports the mean of $A(e)$ over the evaluated predictions (per-unit aggregation averages this score across bearings).

\subsubsection{XJTU-SY}
\label{sec:xjtu_dataset}

\paragraph{Description.}
The XJTU-SY dataset \citep{yaguo2019xjtu} provides 15 run-to-failure bearings under three operating conditions (Table~\ref{tab:bearing_conditions}). Each bearing is recorded through two high-frequency vibration channels until failure, and the full lifetime per bearing is listed in Table~\ref{tab:xjtu_split}.
\begin{table}[H]
\centering
\footnotesize
\caption{XJTU-SY data splits and bearing lifetimes (min).}
\label{tab:xjtu_split}
\begin{adjustbox}{max width=\linewidth}
\begin{tabular}{@{}llc@{}}
\toprule
\textbf{Condition} & \textbf{Bearing Name} & \textbf{Bearing Lifetime (min)} \\
\midrule
Condition 1 & Bearing 1\_1 & 123 \\
Condition 1 & Bearing 1\_2 & 161 \\
Condition 1 & Bearing 1\_3 & 158 \\
Condition 1 & Bearing 1\_4 & 122 \\
Condition 1 & Bearing 1\_5 & 52 \\
\midrule
Condition 2 & Bearing 2\_1 & 491 \\
Condition 2 & Bearing 2\_2 & 161 \\
Condition 2 & Bearing 2\_3 & 533 \\
Condition 2 & Bearing 2\_4 & 42 \\
Condition 2 & Bearing 2\_5 & 339 \\
\midrule
Condition 3 & Bearing 3\_1 & 2538 \\
Condition 3 & Bearing 3\_2 & 2496 \\
Condition 3 & Bearing 3\_3 & 371 \\
Condition 3 & Bearing 3\_4 & 1515 \\
Condition 3 & Bearing 3\_5 & 114 \\
\bottomrule
\end{tabular}
\end{adjustbox}
\end{table}

\paragraph{Data filtering and splitting.}
Our evaluation uses the in-domain fold-1 split: 9 bearings for training (\texttt{1\_3}, \texttt{1\_4}, \texttt{1\_5}, \texttt{2\_3}, \texttt{2\_4}, \texttt{2\_5}, \texttt{3\_3}, \texttt{3\_4}, \texttt{3\_5}), 3 for validation (\texttt{1\_1}, \texttt{2\_1}, \texttt{3\_1}), and 3 for test (\texttt{1\_2}, \texttt{2\_2}, \texttt{3\_2}).

\paragraph{Target creation.}
As for PRONOSTIA, the transform family uses \texttt{HealthIndexTransform} to convert runtime into a normalized degradation target. Table~\ref{tab:xjtu_split} provides the bearing lifetimes that underlie this transformation.

\paragraph{Normalization.}
The preprocessing pipeline standard-scales the raw vibration channels, computes a cumulative-sum feature, extracts time-domain and spectral statistics, and min-max rescales the concatenated descriptor block. Fit-predict models reuse the same features through a tabularized variant.

\paragraph{Evaluation metrics.}
XJTU-SY is evaluated with the \texttt{per\_unit} evaluator, which logs \texttt{phm\_score}, MAE, MSE, and RMSE on inverse-scaled predictions.

\subsection{N-CMAPSS families}
\label{sec:ncmapss_families}

Our evaluation uses three N-CMAPSS-derived families~\citep{arias2021aircraft, frederick2007user}: multi-source prognostics, DS02 prognostics, and multi-source concept diagnostics. All three share the same core preprocessing logic: standard scaling of sensor features and operating descriptors, constant scaling of the RUL target by 0.01, non-overlapping 60-step temporal aggregation, and concatenation of the aggregated features and descriptors.

\subsubsection{Multi-source prognostics}
\label{sec:ncmapss_multisource_prognostics}

\paragraph{Description and splitting.}
The multi-source prognostics family pools four N-CMAPSS sources (DS01, DS04, DS05, and DS07). For each source, units 1--5 are used for training, unit 6 for validation, and units 7--10 for test. The prognostics task predicts RUL after the 60-step temporal aggregation described above.

\paragraph{Evaluation metrics.}
This family uses the \texttt{rul} evaluator, which logs MAE, MSE, RMSE, and \texttt{nasa\_score}.

\subsubsection{DS02 prognostics}
\label{sec:ncmapss_ds02_prognostics}

\paragraph{Description and splitting.}
The DS02 family isolates the subset of N-CMAPSS used in the source paper's protocol. Units 2, 5, 10, 16, and 20 are assigned to training, unit 18 to validation, and units 11, 14, and 15 to test, as summarized in Table~\ref{tab:cmapss_split}. All units belong to flight class~2.

\begin{table}[H]
\centering
\footnotesize
\caption{N-CMAPSS DS02 data splits and flight classes.}
\label{tab:cmapss_split}
\begin{tabular}{@{}llcllcllc@{}}
\toprule
\textbf{Split} & \textbf{Unit} & \textbf{Class} &
\textbf{Split} & \textbf{Unit} & \textbf{Class} &
\textbf{Split} & \textbf{Unit} & \textbf{Class} \\
\midrule
Train & 2  & 2 & Train & 16 & 2 & Test & 11 & 2 \\
Train & 5  & 2 & Train & 20 & 2 & Test & 14 & 2 \\
Train & 10 & 2 & Val   & 18 & 2 & Test & 15 & 2 \\
\bottomrule
\end{tabular}
\end{table}

\paragraph{Evaluation metrics.}
As in the multi-source prognostics family, evaluation uses the \texttt{rul} evaluator and logs MAE, MSE, RMSE, and \texttt{nasa\_score}.

\subsubsection{Multi-source diagnostics}
\label{sec:ncmapss_multisource_diagnostics}

\paragraph{Description and splitting.}
The multi-source diagnostics family reuses the same multi-source datasource but changes the task definition to concept classification. After 60-step aggregation, the transform builds unified concept classes from the source-specific concept annotations. We therefore treat this family as concept diagnostics rather than generic fault-class prediction.

\paragraph{Evaluation metrics.}
This family uses the \texttt{classification} evaluator and logs \texttt{f1}, \texttt{accuracy}, \texttt{precision}, \texttt{recall}, and \texttt{auroc}.

\subsection{Hydraulic diagnostics (HSF15)}
\label{sec:hsf15_dataset}

We evaluate HSF15~\citep{hsf15_helwig} as four separate diagnostics tasks, one per component: accumulator (4 classes), cooler (3 classes), pump (3 classes), and valve (4 classes). All four use the same data loader and default fold, but differ in the target component and number of fault classes.

The preprocessing pipeline min-max scales the raw cycle measurements, extracts time-domain and spectral statistics, and min-max rescales the concatenated descriptor block. The target is reduced to the component-specific last label and evaluated with the classification evaluator, which logs F1, accuracy, precision, recall, and AUROC. Fit-predict models reuse the same statistics-based representation through a tabularized variant.

\subsection{MZVAV}
\label{sec:mzvav_dataset}

\paragraph{Description.}
MZVAV~\citep{Granderson2020} is a simulated building fault-diagnostics dataset for a small commercial building equipped with three air-handling units. The evaluation groups the original faults into four classes: outdoor-air-damper stuck, heating-coil-valve leaking, cooling-coil-valve leaking, and unfaulted.

\paragraph{Data filtering and splitting.}
Our evaluation uses the stratified day-level split summarized in Table~\ref{tab:mzvav_fault_groups}, with fault days preserved across train, validation, and test.

\begin{table}[H]
\centering
\footnotesize
\caption{Number of faulty days per fault group in MZVAV.}
\label{tab:mzvav_fault_groups}
\begin{tabular}{@{}lccc@{}}
\toprule
\textbf{Fault Category} & \textbf{Fault Days} & \textbf{Fault Category} & \textbf{Fault Days} \\ \midrule
OA Damper Stuck & 5 & Cooling Valve Leak & 5 \\
Heating Valve Leak & 3 & Unfaulted & 13 \\
\bottomrule
\end{tabular}
\end{table}

\paragraph{Target creation and normalization.}
The preprocessing pipeline applies a precomputed min-max scaler to the sensor channels and routes the scalar target directly to the fault-classification label used by the diagnostics models. Fit-predict models reuse the same min-max-scaled representation through a tabularized variant.

\paragraph{Evaluation metrics.}
MZVAV uses the \texttt{classification} evaluator and logs \texttt{f1}, \texttt{accuracy}, \texttt{precision}, \texttt{recall}, and \texttt{auroc}.

\subsection{PHME20}
\label{sec:phme20_dataset}

\paragraph{Description.}
PHME20~\citep{PHME20-GTU} is the PHM Society 2020 European Conference Data Challenge dataset. It records an experimental industrial filtration system in which a particulate-laden gas stream is driven through a filter element that progressively clogs as dust accumulates, raising the differential pressure across the filter until an operationally defined threshold is reached. Each run captures one complete filter lifetime from a clean state to end-of-life, and the dataset comprises multiple such run-to-failure trials collected under different operating conditions---varying dust types and feed regimes---so that the prognostics task is exercised across heterogeneous degradation profiles rather than a single nominal scenario. Sensor instrumentation provides differential-pressure, flow, and particulate-loading channels recorded over time, and a per-timestep RUL label is supplied alongside the sensor stream. It is used exclusively as a prognostics task with a direct RUL target.

\paragraph{Data filtering and splitting.}
We follow the default challenge-provided data split, which assigns disjoint filter runs to training, validation, and test partitions. No additional dataset-specific filtering is applied.

\paragraph{Target creation.}
Unlike the bearing families, no health-index transform is required: the dataset already provides a per-timestep RUL signal, which we adopt directly as the prognostics target. The RUL is expressed on a positive scale that decreases monotonically toward end-of-life within each run, and is rescaled but not otherwise reparameterized.

\paragraph{Normalization.}
The preprocessing pipeline min-max scales both the sensor channels and the RUL target, without adding handcrafted time-domain or spectral features. Fit-predict models use a tabularized variant of the same normalized representation.

\paragraph{Evaluation metrics.}
PHME20 uses the \texttt{rul} evaluator and logs MAE, MSE, RMSE, and \texttt{nasa\_score}. The asymmetric \texttt{nasa\_score} is logged alongside the symmetric error metrics because late RUL predictions carry higher operational cost in maintenance scheduling for filtration systems.

\section{Model inventory}
\label{app:models}

This appendix lists the models supported in \picid{}. Each model is registered under a Hydra config key, so experiments can select it via configuration rather than code changes.

\begin{itemize}
\item \textbf{MLP} --- Multi-layer perceptron; \texttt{mlp}.
\item \textbf{LSTM} --- Long short-term memory recurrent networks with gated memory cells that propagate a hidden state step-by-step, enabling nonlinear sequence modeling with adaptive retention of long-term information; the bi-directional variant integrates future covariates effectively \citep{siami-naminiPerformanceLSTMBiLSTM2019}. \texttt{lstm}.
\item \textbf{1D-CNN} --- Convolution-based models that slide learnable kernels over temporal inputs to extract local trends, scale efficiently over long histories, and capture multi-resolution features through residual blocks \citep{lecunDeepLearning2015}. \texttt{cnn\_1d}.
\item \textbf{Timeseries Transformer} --- Transformer for time-series; \texttt{timeseries\_transformer}.
\item \textbf{PatchTST} --- A channel-independent patching Transformer that tokenizes univariate time series into subseries-level patches, enabling longer receptive fields with lower attention cost \citep{Yuqietal-2023-PatchTST}. \texttt{patchtst}.
\item \textbf{Crossformer} --- A Transformer for multivariate time series that tokenizes inputs with cross-dimensional embeddings and applies a two-stage attention layer to model both cross-time and cross-feature dependencies \citep{zhang2023crossformer}. \texttt{crossformer}.
\item \textbf{Spacetimeformer (STF)} --- A long-range Transformer that jointly learns temporal and spatial interactions by treating spatiotemporal values as tokens, combining sequence and graph-like reasoning \citep{grigsby2021longrange}. \texttt{stf}.
\item \textbf{TiDE} --- A dense residual model built on MLP-based encoder--decoders and quasi-linear networks for long-term forecasting \citep{daslong}. \texttt{tide}.
\item \textbf{Linear forecaster} --- Linear model; \texttt{linear\_forecaster}.
\item \textbf{Linear / polynomial / exponential regression} --- Statistical baselines; \texttt{linear\_regression}, \texttt{polynomial\_regression}, \texttt{exponential\_regression}.
\item \textbf{SES, naive, mean, drift, persistence} --- Simple baselines; \texttt{ses}, \texttt{naive}, \texttt{mean}, \texttt{drift}, \texttt{persistence}.
\item \textbf{Similar period} --- Similar-period forecasting; \texttt{similar\_period}.
\item \textbf{XGBoost} --- Classical tree-based baseline that leverages gradient boosting for consistently strong performance on regression and classification tasks \citep{chen2016xgboost}. \texttt{xgboost\_fit\_predict}, \texttt{xgboost\_batch\_context}.
\item \textbf{CatBoost} --- Gradient boosting; \texttt{catboost}.
\item \textbf{AutoGluon} --- AutoML tabular; \texttt{autogluon\_fit\_predict}.
\item \textbf{TabPFN} --- Tabular Prior-Fitted Networks \citep{hollmanntabpfn,hollmann2025accurate} are trained on massive causally-generated synthetic tabular datasets and perform Bayesian inference via in-context learning; they have been shown to generalize to real-world tabular tasks, often outperforming tree-based methods in both speed and accuracy. \texttt{tabpfn\_fit\_predict}, \texttt{tabpfn\_batch\_context}.
\item \textbf{TabDPT} --- A transformer-based tabular foundation model trained on real datasets that uses retrieval-based self-supervised pre-training and in-context learning to generalize to unseen tabular data without task-specific training or hyperparameter tuning \citep{ma2024tabdpt}. \texttt{tabdpt\_fit\_predict}.
\item \textbf{CART-E} --- CART-E wrapper; \texttt{carte\_fit\_predict}.
\item \textbf{Isolation Forest} --- Fault detection; \texttt{isolation\_forest\_fit\_predict}.
\end{itemize}

\section{Full experimental results}
\label{app:full_experiments}

This appendix presents the complete experimental setup and results for the benchmark evaluation. It documents the evaluated experiment families, preprocessing schemas, and hyperparameter search spaces. For the condensed main-text summary, see Section~\ref{sec:empirical_validation}; for dataset-level protocol details, see Appendix~\ref{sec:dataset_descriptions}.

\subsection{Experimental setup}
\label{app:exp_setup_full}

\paragraph{Experiment definitions.}
Each experiment family is defined by a Hydra configuration \cite{Yadan2019Hydra} that specifies the datasource, transform pipeline, model, evaluator, seed set, and hyperparameter search space. The benchmark covers gradient-trained and fit-predict models for both diagnostics and prognostics tasks. XGBoost uses a separate fit-predict configuration that reuses the same experiment definitions and transform families as the other tabular models.

\paragraph{Learning tasks and datasets.}
The evaluation covers two PHM task categories. Diagnostics comprises multiclass fault classification on MZVAV~\citep{Granderson2020}, four component-specific hydraulic diagnostics tasks on HSF15~\citep{hsf15_helwig} (accumulator, cooler, pump, and valve), and concept classification on N-CMAPSS Multi~\citep{arias2021aircraft, frederick2007user}. Prognostics comprises ah-RUL regression on NB14~\citep{bole2014adaptation} and UNIBO21~\citep{univbo_dataset}, direct RUL regression on PHME20~\citep{PHME20-GTU}, RUL prediction on the N-CMAPSS Multi and DS02 families, and bearing prognostics on XJTU-SY~\citep{yaguo2019xjtu}.

Target semantics differ by family. NB14 and UNIBO21 predict ah-RUL, i.e., remaining cumulative discharge throughput in Ampere-hours; PHME20 and the N-CMAPSS families predict remaining useful life; and the bearing family transforms runtime trajectories into a normalized degradation target through the configured health-index transform, while evaluation is reported with the per-unit metric suite. All families use disjoint train/validation/test entities according to their datasource definitions, with MZVAV as the only day-stratified diagnostics exception.

\paragraph{Models.}
The benchmark compares five model families (Section~\ref{subsec:exp_setup}): (i) simple baselines (Linear, Exp, MLP), (ii) deep sequence models (LSTM, CNN-1D, TiDE), (iii) transformers (TST, STF, CF, PTST), (iv) tabular models (XGBoost), and (v) tabular foundation models (TabPFN, TabDPT). Linear is regression-only and is therefore omitted from diagnostics; for diagnostics, a linear classifier serves the analogous baseline role.

\paragraph{Training and evaluation.}
Every model--dataset configuration is repeated over five random seeds. Gradient-trained models search over \mbox{$\texttt{seq\_len} \in \{1,10,50\}$} and \mbox{$\texttt{lr} \in \{0.001, 0.0005, 0.0001\}$}, with batch size 512, a maximum of 200 epochs, and early stopping. Fit-predict models search over context/stride pairs $(1,1)$, $(5,1)$, $(10,5)$, $(20,5)$, and $(50,50)$. Diagnostics selects the best configuration on \texttt{val/f1}; prognostics (generic RUL objective) selects on \texttt{val/loss}. Evaluation uses three evaluator families: \texttt{classification} (diagnostics), \texttt{rul} (direct RUL regression), and \texttt{per\_unit} (battery and bearing families).

\subsection{Transformation schemas}
\label{app:transform_schemas}

All experiment families obey the same leakage-prevention invariant: any fitted normalization or feature-extraction statistic is estimated on the training partition only and then reused unchanged for validation and test. The subsections below describe the preprocessing pipeline applied to each dataset group; individual transform operators are drawn from the inventory in Appendix~\ref{app:transform_inventory}. Across all families, the transformed tensors are subsequently windowed according to the sequence-length and stride settings described in Section~\ref{app:hyperparameter_search}.

\subsubsection{Battery datasets (NB14 and UNIBO21)}
\label{app:schema_battery}

For both NB14 and UNIBO21, raw sensor channels and the ah-RUL target are min-max scaled (MinMaxScaler). Two descriptor branches are then extracted from the cycle traces: time-domain statistics---mean, variance, kurtosis, peak factor, and related summaries (TimeStatsTransform)---and frequency-domain statistics via FFT (SpectralStatsTransform). The resulting descriptor vectors are concatenated (ConcatenateTransform) and re-scaled with a final min-max transform. For fit-predict models, the pipeline additionally tabularizes the processed history into a single feature vector for one-shot inference (TimeseriesTabularizer). This pipeline corresponds to the \texttt{combined} / \texttt{combined\_fit\_predict} configuration group.

\subsubsection{Bearing dataset (XJTU-SY)}
\label{app:schema_bearings}

For XJTU-SY, raw vibration channels are standardized with training-partition statistics (StandardScaler). Runtime targets are converted into a normalized degradation trajectory via HealthIndexTransform (see Eq.~\eqref{eq:hi_target}). An additional cumulative-sum feature is computed and min-max scaled, and time-domain plus spectral descriptors (TimeStatsTransform, SpectralStatsTransform) are extracted from the standardized vibration signals. The resulting features are concatenated and min-max rescaled. For fit-predict models, the pipeline tabularizes the combined representation while preserving unit identifiers for per-unit evaluation. This pipeline corresponds to the \texttt{combined} / \texttt{combined\_fit\_predict} configuration group.

\subsubsection{N-CMAPSS families}
\label{app:schema_ncmapss}

Sensor features and operating descriptors are standardized with fixed N-CMAPSS scalers (N\_CMAPSSDescriptorsScaler, StandardScaler), and the RUL label is multiplied by the constant factor 0.01 (ConstantScaler). Each flight is then temporally aggregated with non-overlapping windows of 60~timesteps (WindowedAggregationTransform), and the aggregated sensor features and descriptors are concatenated into a single input representation. For the multi-source diagnostics family, the transform additionally builds unified concept classes from the source-specific concept annotations. Fit-predict models tabularize the aggregated histories after the same scaling and concatenation stages. This pipeline corresponds to the \texttt{depater2023\_default} / \texttt{depater2023\_fit\_predict\_history} configuration group.

\subsubsection{Building diagnostics (MZVAV)}
\label{app:schema_mzvav}

No additional feature engineering is applied for MZVAV. The pipeline rescales the sensor channels with a precomputed dataset-specific min-max scaler (MinMaxScalerMZVAV) and routes the scalar target directly to the fault-classification key consumed by the diagnostics models. For fit-predict models, the pipeline tabularizes the resulting history windows without introducing a separate descriptor stage. This pipeline corresponds to the \texttt{default} / \texttt{fit\_predict\_history} configuration group.

\subsubsection{Hydraulic diagnostics (HSF15)}
\label{app:schema_hsf15}

All four HSF15 component tasks share the same preprocessing pipeline. Features are min-max scaled (MinMaxScaler), the component-specific target is reduced to the last label in each window and assigned to the fault-classification key, and the sensor burst is summarized through time-domain statistics (TimeStatsTransform) and spectral statistics (SpectralStatsTransform), followed by concatenation and a final min-max rescaling. For fit-predict models, the pipeline tabularizes the statistics-based representation for XGBoost, TabPFN, and TabDPT. This pipeline corresponds to the \texttt{default} / \texttt{statistics\_fit\_predict} configuration group.

\subsubsection{PHM challenge prognostics (PHME20)}
\label{app:schema_phme20}

Both features and the direct RUL target are min-max scaled (MinMaxScaler), with no additional handcrafted feature extraction. For fit-predict models, the pipeline adds history tabularization for one-shot inference. This pipeline corresponds to the \texttt{normalize\_feature\_target} / \texttt{normalize\_feature\_target\_fit\_predict} configuration group.

\subsection{Hyperparameter search}
\label{app:hyperparameter_search}

The benchmark exposes two hyperparameter-search families: a gradient-trained family for sequence models and simple neural baselines, and a fit-predict family for tabular/foundation models including XGBoost. Table~\ref{tab:hyperparameter_search} summarizes the search spaces.

\begin{table}[H]
\centering
\footnotesize
\caption{Hyperparameter search families used in the benchmark evaluation.}
\label{tab:hyperparameter_search}
\renewcommand{\arraystretch}{1.08}
\begin{adjustbox}{max width=\linewidth}
\begin{tabular}{@{}p{2.1cm}p{4.1cm}p{4.8cm}p{2.4cm}p{2.1cm}@{}}
\toprule
\textbf{Search family} & \textbf{Models} & \textbf{Search space} & \textbf{Fixed settings} & \textbf{Selection rule} \\
\midrule
Gradient-trained & LSTM, 1D-CNN, STF, Crossformer, Timeseries Transformer, TiDE, PatchTST, MLP, linear classifier, linear regression, exponential regression & \texttt{seq\_len} $\in \{1,10,50\}$; \texttt{lr} $\in \{0.001, 0.0005, 0.0001\}$ & Batch size 512; max 200 epochs; early stopping & \texttt{val/f1} for diagnostics; \texttt{val/loss} for prognostics \\
Fit-predict & XGBoost, TabPFN, TabDPT & Context/stride pairs $(1,1)$, $(5,1)$, $(10,5)$, $(20,5)$, $(50,50)$ & Five seeds; deterministic transform reuse; one-shot fit/predict evaluation & \texttt{val/f1} for diagnostics; \texttt{val/loss} for prognostics \\
\bottomrule
\end{tabular}
\end{adjustbox}
\end{table}

All successful runs persist the resolved configuration, Hydra override trace, run metadata, and reproduction instructions alongside the predictions and evaluator outputs. The corresponding artifact protocol is described in Appendix~\ref{app:reproducibility}.

\subsection{Reading the result tables}
\label{app:reading_guide}

The result tables in Sections~\ref{app:full_diagnostics_results}--\ref{app:full_prognostics_results} share a common reading convention, summarized here once so individual captions can stay terse.

\paragraph{Dataset abbreviations.}
Column headers use short identifiers consistently across both the main-text and appendix tables. Their full meanings are:

\begin{table}[H]
\centering
\footnotesize
\renewcommand{\arraystretch}{1.05}
\begin{tabular}{@{}llp{6.0cm}@{}}
\toprule
\textbf{Abbreviation} & \textbf{Task} & \textbf{Source dataset (domain)} \\
\midrule
NC-DS02 & Prognostics & N-CMAPSS DS02 (turbofan engine RUL) \\
NC-P    & Prognostics & N-CMAPSS Multi-source (turbofan engine RUL) \\
NB14    & Prognostics & NASA Randomized Battery Usage (battery ah-RUL) \\
PHME20  & Prognostics & PHM 2020 Challenge (industrial filtration RUL) \\
Unibo   & Prognostics & UNIBO Powertools (battery ah-RUL) \\
XJTU-SY & Prognostics & XJTU-SY (bearing degradation) \\
NC-D    & Diagnostics & N-CMAPSS Multi-source (turbofan concept classification) \\
HSF15-A & Diagnostics & HSF15 (hydraulic accumulator, 4-way) \\
HSF15-C & Diagnostics & HSF15 (hydraulic cooler, 3-way) \\
HSF15-P & Diagnostics & HSF15 (hydraulic pump, 3-way) \\
HSF15-V & Diagnostics & HSF15 (hydraulic valve, 4-way) \\
MZVAV   & Diagnostics & MZVAV (multi-zone HVAC fault, 4-way) \\
\bottomrule
\end{tabular}
\end{table}%

\paragraph{Cell convention.}
Each cell reports mean $\pm$ std over five independent seeds (Section~\ref{subsec:exp_setup}). Bold cells are the best in their column; underlined cells are the second best. Rows are grouped by model family in the order: simple baselines, deep sequence models, transformers, tabular models, tabular foundation models.

\paragraph{Metric scaling and direction.}
Diagnostics metrics (F1, AUROC, Accuracy) are reported on a 0--100 scale; prognostics MAE/MSE are reported both in the framework's normalized target space (reported $\times 100$) and in original engineering units (denormalized) for practitioner interpretation. Arrows in column headers indicate metric direction ($\downarrow$ lower is better, $\uparrow$ higher is better). The \emph{Avg rank} column is the mean of per-task ranks across the columns of that table.

\subsection{Diagnostics results}
\label{app:full_diagnostics_results}

Diagnostics is evaluated by F1 (the headline metric in Section~\ref{sec:empirical_validation}) and complemented here by AUROC and Accuracy. The three metrics agree on the top group: TabDPT, TabPFN, CNN-1D, and XGBoost cluster within $1.0$ average-rank of each other on F1, with LSTM joining the leading tier on F1 and Accuracy. MZVAV (multi-zone HVAC fault classification under a day-stratified split) is the hardest task in every metric and the only family where the gap to chance is small. The metric-robustness check therefore supports the F1 choice in the main text.

\begin{table}[H]
\caption{F1 score on diagnostics ($\uparrow$). Same numbers as the diagnostics floor of Table~\ref{tab:results_main_combined}, reproduced here with full per-task breakdown.}
\label{tab:results_test_best_rerun_f1}
\begin{adjustbox}{max width=\textwidth}
\begin{tabular}{lrrrrrrr}
\toprule
Model & \multicolumn{1}{c}{NC-D} & \multicolumn{1}{c}{HSF15-A} & \multicolumn{1}{c}{HSF15-C} & \multicolumn{1}{c}{HSF15-P} & \multicolumn{1}{c}{HSF15-V} & \multicolumn{1}{c}{MZVAV} & \multicolumn{1}{c}{Average rank} \\
\midrule
Linear & 72.34 ± 3.04 & 58.75 ± 1.81 & 98.35 ± 0.82 & 54.40 ± 11.97 & 32.55 ± 2.36 & 39.89 ± 8.77 & 7.33 \\
MLP & 79.49 ± 1.80 & 91.02 ± 2.25 & \underline{99.91 ± 0.14} & 97.32 ± 0.60 & 80.99 ± 29.38 & 60.10 ± 6.39 & 5.00 \\
LSTM & \cellcolor[gray]{0.85}\textbf{88.84 ± 0.73} & 94.59 ± 0.97 & \cellcolor[gray]{0.85}\textbf{100.00 ± 0.00} & 95.94 ± 2.83 & 97.35 ± 3.32 & 51.31 ± 6.01 & 3.83 \\
CNN-1D & \underline{87.53 ± 2.83} & 94.03 ± 1.98 & \cellcolor[gray]{0.85}\textbf{100.00 ± 0.00} & 98.73 ± 0.52 & 97.92 ± 0.83 & \underline{66.11 ± 5.74} & 3.00 \\
TiDE & 32.57 ± 4.65 & 42.90 ± 5.07 & 61.57 ± 10.41 & 59.37 ± 7.68 & 42.11 ± 14.36 & 25.19 ± 5.29 & 8.17 \\
TST & 26.34 ± 3.96 & 37.37 ± 4.38 & 59.18 ± 10.29 & 46.07 ± 4.13 & 35.40 ± 5.77 & 24.93 ± 4.38 & 10.00 \\
STF & 24.55 ± 3.60 & 40.01 ± 5.75 & 65.94 ± 20.61 & 50.39 ± 11.54 & 37.34 ± 5.68 & 38.01 ± 5.56 & 8.67 \\
CF & 23.74 ± 0.93 & 25.04 ± 5.28 & 59.19 ± 10.05 & 29.46 ± 2.61 & 23.98 ± 2.95 & 17.34 ± 4.24 & 11.50 \\
PTST & 19.57 ± 0.26 & 31.56 ± 4.04 & 41.57 ± 5.18 & 41.22 ± 6.63 & 26.20 ± 2.44 & 25.81 ± 4.15 & 11.00 \\
XGBoost & 48.13 ± 0.00 & \underline{98.07 ± 0.00} & \cellcolor[gray]{0.85}\textbf{100.00 ± 0.00} & \underline{99.66 ± 0.00} & \underline{99.65 ± 0.00} & 57.08 ± 0.00 & 3.17 \\
TabPFN & 67.15 ± 1.46 & \cellcolor[gray]{0.85}\textbf{99.47 ± 0.00} & \cellcolor[gray]{0.85}\textbf{100.00 ± 0.00} & \cellcolor[gray]{0.85}\textbf{100.00 ± 0.00} & \cellcolor[gray]{0.85}\textbf{100.00 ± 0.00} & 58.32 ± 2.44 & 2.33 \\
TabDPT & 85.21 ± 0.16 & 96.66 ± 1.03 & \cellcolor[gray]{0.85}\textbf{100.00 ± 0.00} & 99.06 ± 0.25 & 98.92 ± 0.35 & \cellcolor[gray]{0.85}\textbf{71.29 ± 0.48} & 2.33 \\
\bottomrule
\end{tabular}
\end{adjustbox}
\end{table}

\begin{table}[H]
\caption{AUROC on diagnostics ($\uparrow$).}
\label{tab:results_test_best_rerun_auroc}
\begin{adjustbox}{max width=\textwidth}
\begin{tabular}{lrrrrrrr}
\toprule
Model & \multicolumn{1}{c}{NC-D} & \multicolumn{1}{c}{HSF15-A} & \multicolumn{1}{c}{HSF15-C} & \multicolumn{1}{c}{HSF15-P} & \multicolumn{1}{c}{HSF15-V} & \multicolumn{1}{c}{MZVAV} & \multicolumn{1}{c}{Average rank} \\
\midrule
Linear & 93.76 ± 1.29 & 85.10 ± 1.35 & \underline{99.43 ± 0.63} & 77.41 ± 12.99 & 64.74 ± 4.49 & 71.47 ± 9.58 & 7.17 \\
MLP & 96.31 ± 0.67 & 98.95 ± 0.39 & \cellcolor[gray]{0.85}\textbf{100.00 ± 0.00} & 99.84 ± 0.07 & 90.02 ± 21.11 & 85.24 ± 2.03 & 5.17 \\
LSTM & \cellcolor[gray]{0.85}\textbf{98.41 ± 0.16} & 99.54 ± 0.11 & \cellcolor[gray]{0.85}\textbf{100.00 ± 0.00} & 99.68 ± 0.27 & 99.88 ± 0.17 & \underline{86.32 ± 4.06} & 3.33 \\
CNN-1D & \underline{98.11 ± 0.62} & 99.44 ± 0.19 & \cellcolor[gray]{0.85}\textbf{100.00 ± 0.00} & 99.94 ± 0.06 & 99.94 ± 0.05 & 85.45 ± 2.29 & 3.17 \\
TiDE & 62.71 ± 3.84 & 69.00 ± 5.66 & 81.46 ± 5.14 & 78.78 ± 11.24 & 64.91 ± 11.98 & 60.77 ± 9.35 & 8.00 \\
TST & 57.03 ± 2.91 & 62.99 ± 6.37 & 79.87 ± 9.04 & 65.30 ± 6.07 & 61.60 ± 5.19 & 57.44 ± 9.01 & 9.33 \\
STF & 55.23 ± 3.04 & 58.44 ± 5.10 & 81.96 ± 14.13 & 64.97 ± 9.06 & 59.98 ± 15.81 & 62.61 ± 10.35 & 9.50 \\
CF & 52.00 ± 0.87 & 53.48 ± 5.54 & 73.59 ± 8.17 & 37.35 ± 0.53 & 51.76 ± 0.94 & 47.64 ± 6.33 & 11.67 \\
PTST & 50.01 ± 0.24 & 58.74 ± 1.64 & 59.83 ± 4.32 & 62.24 ± 5.03 & 52.84 ± 2.86 & 52.97 ± 5.01 & 11.17 \\
XGBoost & 82.56 ± 0.00 & \underline{99.94 ± 0.00} & \cellcolor[gray]{0.85}\textbf{100.00 ± 0.00} & \cellcolor[gray]{0.85}\textbf{100.00 ± 0.00} & \cellcolor[gray]{0.85}\textbf{100.00 ± 0.00} & 84.51 ± 0.00 & 3.17 \\
TabPFN & 93.97 ± 0.31 & \cellcolor[gray]{0.85}\textbf{100.00 ± 0.00} & \cellcolor[gray]{0.85}\textbf{100.00 ± 0.00} & \cellcolor[gray]{0.85}\textbf{100.00 ± 0.00} & \cellcolor[gray]{0.85}\textbf{100.00 ± 0.00} & 83.67 ± 0.77 & 2.67 \\
TabDPT & 97.38 ± 0.06 & 99.91 ± 0.05 & \cellcolor[gray]{0.85}\textbf{100.00 ± 0.00} & \underline{99.99 ± 0.00} & \underline{99.99 ± 0.00} & \cellcolor[gray]{0.85}\textbf{88.57 ± 0.73} & 2.50 \\
\bottomrule
\end{tabular}
\end{adjustbox}
\end{table}

\begin{table}[H]
\caption{Accuracy on diagnostics ($\uparrow$).}
\label{tab:results_test_best_rerun_accuracy}
\begin{adjustbox}{max width=\textwidth}
\begin{tabular}{lrrrrrrr}
\toprule
Model & \multicolumn{1}{c}{NC-D} & \multicolumn{1}{c}{HSF15-A} & \multicolumn{1}{c}{HSF15-C} & \multicolumn{1}{c}{HSF15-P} & \multicolumn{1}{c}{HSF15-V} & \multicolumn{1}{c}{MZVAV} & \multicolumn{1}{c}{Average rank} \\
\midrule
Linear & 73.79 ± 2.35 & 61.47 ± 1.91 & 98.36 ± 0.82 & 63.08 ± 14.59 & 45.93 ± 2.01 & 61.56 ± 4.94 & 7.33 \\
MLP & 79.72 ± 1.82 & 92.39 ± 2.06 & \underline{99.91 ± 0.14} & 98.05 ± 0.41 & 85.28 ± 22.19 & 71.54 ± 4.90 & 5.00 \\
LSTM & \cellcolor[gray]{0.85}\textbf{88.42 ± 0.72} & 95.42 ± 0.82 & \cellcolor[gray]{0.85}\textbf{100.00 ± 0.00} & 96.96 ± 1.94 & 97.64 ± 3.10 & 70.71 ± 4.77 & 3.50 \\
CNN-1D & \underline{86.91 ± 2.91} & 95.00 ± 1.57 & \cellcolor[gray]{0.85}\textbf{100.00 ± 0.00} & 99.09 ± 0.36 & 98.38 ± 0.67 & \underline{77.81 ± 3.10} & 3.00 \\
TiDE & 35.30 ± 3.73 & 49.24 ± 10.75 & 65.39 ± 8.98 & 70.93 ± 3.78 & 57.08 ± 10.30 & 30.15 ± 7.47 & 8.50 \\
TST & 26.80 ± 3.37 & 43.36 ± 8.23 & 62.53 ± 10.12 & 52.24 ± 6.62 & 45.19 ± 6.74 & 45.19 ± 9.25 & 9.83 \\
STF & 24.82 ± 3.12 & 47.02 ± 9.33 & 72.30 ± 15.62 & 64.90 ± 9.85 & 57.59 ± 4.35 & 50.34 ± 9.76 & 8.50 \\
CF & 28.90 ± 3.18 & 30.71 ± 5.80 & 61.83 ± 10.46 & 35.15 ± 4.49 & 33.06 ± 3.25 & 27.02 ± 10.45 & 11.17 \\
PTST & 21.56 ± 0.93 & 34.41 ± 4.03 & 42.44 ± 4.66 & 43.36 ± 7.01 & 31.85 ± 5.87 & 39.70 ± 5.78 & 11.33 \\
XGBoost & 54.75 ± 0.00 & \underline{98.53 ± 0.00} & \cellcolor[gray]{0.85}\textbf{100.00 ± 0.00} & \underline{99.77 ± 0.00} & \underline{99.77 ± 0.00} & 62.62 ± 0.00 & 3.33 \\
TabPFN & 70.61 ± 1.07 & \cellcolor[gray]{0.85}\textbf{99.58 ± 0.00} & \cellcolor[gray]{0.85}\textbf{100.00 ± 0.00} & \cellcolor[gray]{0.85}\textbf{100.00 ± 0.00} & \cellcolor[gray]{0.85}\textbf{100.00 ± 0.00} & 65.55 ± 1.96 & 2.50 \\
TabDPT & 84.74 ± 0.16 & 97.27 ± 0.83 & \cellcolor[gray]{0.85}\textbf{100.00 ± 0.00} & 99.27 ± 0.19 & 99.17 ± 0.26 & \cellcolor[gray]{0.85}\textbf{80.10 ± 0.36} & 2.33 \\
\bottomrule
\end{tabular}
\end{adjustbox}
\end{table}

\paragraph{Cross-family observations.}
Across the three diagnostics metrics, the model families separate clearly. Tabular foundation models (TabDPT, TabPFN) and XGBoost rank in the top tier even though they consume tabularized windows rather than raw sequences; this is most visible on HSF15 (where many models reach 100 F1), but it also holds on harder tasks such as NC-D and MZVAV. Deep sequence models split into a strong pair (LSTM, CNN-1D) and a weaker TiDE variant under this training budget. By contrast, the transformer family (TST, STF, CF, PTST) performs near chance on most diagnostics tasks, despite consuming identical inputs under the same protocol. Simple baselines (MLP, Linear) remain competitive on several HSF15 components, reinforcing that hydraulic component diagnostics is comparatively easy relative to MZVAV and NC-D. Finally, the robustness check is consistent: F1, AUROC, and Accuracy induce nearly identical rankings, so the headline conclusions do not depend on the specific metric choice.

\subsection{Prognostics results}
\label{app:full_prognostics_results}

The prognostics metric surface is reported in three families. Aggregate MAE and MSE (normalized and denormalized) are computed by pooling all predictions and all units before averaging. Per-unit aggregation, defined only on battery and bearing tasks where the framework's \texttt{per\_unit} evaluator is active, computes one error per monitored unit before averaging across units; this tightens rankings among leading models and reduces the influence of long-trajectory units that dominate window-level pools. Domain-specific scores (NASA, PHM) are reported with the per-task scoping enforced by the framework's metric registry.

\subsubsection{Aggregate errors}
\label{app:prog_aggregate}

Tables~\ref{tab:results_app_mae} and~\ref{tab:results_app_mse} report MAE and MSE in a two-block format. The top block reports errors in the normalized target space (reported $\times 100$), which is used to compute cross-task \emph{Avg rank}; the bottom block reports the same predictions in original engineering units for practitioner interpretation. MSE emphasizes occasional large errors, but it preserves the headline ordering among the leading models. Because MSE penalizes large residuals quadratically, it is sensitive to rare catastrophic predictions; we report it alongside MAE to make these failure modes visible.

\begin{table}[H]
\centering
\footnotesize
\caption{MAE on prognostics. Top block: normalized target space ($\times 100$) ($\downarrow$); bottom block: original engineering units ($\downarrow$). Same numbers as the prognostics floor of Table~\ref{tab:results_main_combined} (top), reproduced with full per-task breakdown.}
\label{tab:results_app_mae}
\begin{adjustbox}{max width=\textwidth}
\begin{tabular}{lrrrrrrr}
\toprule
Model & \multicolumn{1}{c}{NC-DS02} & \multicolumn{1}{c}{NC-P} & \multicolumn{1}{c}{NB14} & \multicolumn{1}{c}{PHME20} & \multicolumn{1}{c}{Unibo} & \multicolumn{1}{c}{XJTU-SY} & \multicolumn{1}{c}{Average rank} \\
\midrule
\multicolumn{8}{l}{\textit{Normalized target space ($\downarrow$)}} \\
\midrule
Linear & 10.13 ± 0.14 & 16.11 ± 0.60 & 41.69 ± 12.02 & 12.19 ± 0.36 & 27.59 ± 14.36 & 76.80 ± 60.41 & 12.50 \\
Exp & 5.35 ± 0.06 & 10.96 ± 0.09 & 30.47 ± 47.76 & 8.82 ± 0.52 & 12.19 ± 0.31 & 27.22 ± 4.06 & 9.67 \\
MLP & 6.37 ± 0.23 & 13.17 ± 0.78 & 14.38 ± 9.77 & 4.62 ± 1.15 & 12.50 ± 0.76 & 30.64 ± 2.67 & 10.33 \\
LSTM & \underline{4.93 ± 0.13} & 7.56 ± 0.31 & 3.80 ± 0.22 & 3.73 ± 0.98 & 6.50 ± 0.16 & \cellcolor[gray]{0.85}\textbf{21.89 ± 0.40} & 3.67 \\
CNN-1D & 5.33 ± 0.37 & 7.53 ± 0.22 & 8.89 ± 1.70 & 5.35 ± 3.71 & 12.41 ± 1.15 & 31.02 ± 8.25 & 8.67 \\
TiDE & 5.29 ± 0.22 & 7.62 ± 0.20 & \cellcolor[gray]{0.85}\textbf{3.44 ± 0.17} & 4.20 ± 0.66 & 6.46 ± 0.78 & 25.11 ± 2.38 & 5.17 \\
TST & 5.31 ± 0.13 & \underline{7.02 ± 0.17} & 6.28 ± 0.25 & 4.11 ± 0.84 & 7.23 ± 0.39 & 33.30 ± 7.72 & 7.00 \\
STF & \cellcolor[gray]{0.85}\textbf{4.89 ± 0.10} & 7.35 ± 1.16 & 10.67 ± 3.16 & 3.91 ± 1.00 & 8.89 ± 0.81 & 28.49 ± 4.01 & 6.17 \\
CF & 5.76 ± 0.51 & 9.98 ± 0.57 & \underline{3.57 ± 0.07} & 3.87 ± 0.85 & 5.58 ± 1.08 & \underline{22.09 ± 1.06} & 5.00 \\
PTST & 16.62 ± 0.04 & 21.55 ± 0.03 & 5.22 ± 0.10 & 15.09 ± 1.13 & 11.18 ± 1.11 & 25.42 ± 1.48 & 10.33 \\
XGBoost & 8.52 ± 0.00 & 15.24 ± 0.00 & 4.48 ± 0.00 & 2.68 ± 0.00 & 4.06 ± 0.00 & 24.59 ± 0.00 & 6.50 \\
TabPFN & 4.96 ± 0.04 & 7.79 ± 0.04 & 3.91 ± 0.03 & \cellcolor[gray]{0.85}\textbf{1.95 ± 0.03} & \cellcolor[gray]{0.85}\textbf{3.72 ± 0.06} & 22.27 ± 0.35 & 3.33 \\
TabDPT & 5.07 ± 0.06 & \cellcolor[gray]{0.85}\textbf{6.85 ± 0.02} & 3.63 ± 0.04 & \underline{2.19 ± 0.01} & \underline{3.94 ± 0.05} & 23.24 ± 0.45 & 2.67 \\
\midrule
\multicolumn{8}{l}{\textit{Original engineering units ($\downarrow$)}} \\
\midrule
Linear & 10.13 ± 0.14 & 16.11 ± 0.60 & 451.64 ± 130.15 & 43.13 ± 1.27 & 135.80 ± 70.68 & 907.40 ± 774.36 & 12.50 \\
Exp & 5.35 ± 0.06 & 10.96 ± 0.09 & 330.02 ± 517.36 & 31.21 ± 1.85 & 60.02 ± 1.51 & 349.71 ± 60.98 & 9.67 \\
MLP & 6.37 ± 0.23 & 13.17 ± 0.78 & 155.77 ± 105.84 & 16.35 ± 4.06 & 61.51 ± 3.72 & 384.95 ± 40.56 & 10.50 \\
LSTM & \underline{4.93 ± 0.13} & 7.56 ± 0.31 & 41.13 ± 2.34 & 13.19 ± 3.47 & 32.00 ± 0.77 & \cellcolor[gray]{0.85}\textbf{271.33 ± 3.45} & 3.67 \\
CNN-1D & 5.33 ± 0.37 & 7.53 ± 0.22 & 96.25 ± 18.38 & 18.95 ± 13.12 & 61.10 ± 5.66 & 380.05 ± 96.92 & 8.50 \\
TiDE & 5.29 ± 0.22 & 7.62 ± 0.20 & \cellcolor[gray]{0.85}\textbf{37.22 ± 1.80} & 14.86 ± 2.35 & 31.78 ± 3.86 & 320.41 ± 24.80 & 5.33 \\
TST & 5.31 ± 0.13 & \underline{7.02 ± 0.17} & 68.00 ± 2.71 & 14.56 ± 2.96 & 35.60 ± 1.94 & 424.88 ± 100.25 & 7.00 \\
STF & \cellcolor[gray]{0.85}\textbf{4.89 ± 0.10} & 7.35 ± 1.16 & 115.61 ± 34.23 & 13.83 ± 3.53 & 43.76 ± 4.01 & 361.41 ± 54.88 & 6.17 \\
CF & 5.76 ± 0.51 & 9.98 ± 0.57 & \underline{38.70 ± 0.80} & 13.71 ± 3.02 & 27.48 ± 5.31 & \underline{282.20 ± 19.10} & 5.00 \\
PTST & 16.62 ± 0.04 & 21.55 ± 0.03 & 56.53 ± 1.09 & 53.40 ± 3.99 & 55.02 ± 5.47 & 312.68 ± 23.46 & 10.17 \\
XGBoost & 8.52 ± 0.00 & 15.24 ± 0.00 & 48.54 ± 0.00 & 9.48 ± 0.00 & 19.98 ± 0.00 & 311.19 ± 0.00 & 6.50 \\
TabPFN & 4.96 ± 0.04 & 7.79 ± 0.04 & 42.32 ± 0.28 & \cellcolor[gray]{0.85}\textbf{6.89 ± 0.09} & \cellcolor[gray]{0.85}\textbf{18.33 ± 0.30} & 292.43 ± 5.26 & 3.50 \\
TabDPT & 5.07 ± 0.06 & \cellcolor[gray]{0.85}\textbf{6.85 ± 0.02} & 39.35 ± 0.47 & \underline{7.75 ± 0.04} & \underline{19.42 ± 0.24} & 282.83 ± 5.84 & 2.50 \\
\bottomrule
\end{tabular}
\end{adjustbox}
\end{table}

\begin{table}[H]
\centering
\footnotesize
\caption{MSE on prognostics. Top block: normalized target space ($\times 100$) ($\downarrow$); bottom block: original engineering units ($\downarrow$). MSE redistributes weight onto large per-window errors but preserves the leading-model ranking from MAE.}
\label{tab:results_app_mse}
\begin{adjustbox}{max width=\textwidth}
\begin{tabular}{lrrrrrrr}
\toprule
Model & \multicolumn{1}{c}{NC-DS02} & \multicolumn{1}{c}{NC-P} & \multicolumn{1}{c}{NB14} & \multicolumn{1}{c}{PHME20} & \multicolumn{1}{c}{Unibo} & \multicolumn{1}{c}{XJTU-SY} & \multicolumn{1}{c}{Average rank} \\
\midrule
\multicolumn{8}{l}{\textit{Normalized target space ($\downarrow$)}} \\
\midrule
Linear & 1.37 ± 0.04 & 3.95 ± 0.38 & 32.16 ± 16.14 & 2.13 ± 0.10 & 17.83 ± 17.83 & 107.37 ± 128.29 & 12.50 \\
Exp & 0.47 ± 0.01 & 1.92 ± 0.03 & 28.60 ± 61.10 & 1.18 ± 0.13 & 2.59 ± 0.08 & 11.21 ± 3.67 & 9.50 \\
MLP & 0.76 ± 0.05 & 2.77 ± 0.35 & 5.09 ± 6.91 & 0.35 ± 0.15 & 2.91 ± 0.51 & 14.80 ± 3.26 & 10.50 \\
LSTM & \underline{0.43 ± 0.02} & 1.04 ± 0.09 & 0.27 ± 0.03 & 0.23 ± 0.11 & 1.38 ± 0.07 & \cellcolor[gray]{0.85}\textbf{6.80 ± 0.22} & 3.83 \\
CNN-1D & 0.48 ± 0.05 & 1.01 ± 0.06 & 1.31 ± 0.60 & 0.60 ± 0.75 & 2.54 ± 0.48 & 14.34 ± 8.16 & 8.33 \\
TiDE & 0.47 ± 0.04 & 1.09 ± 0.04 & \cellcolor[gray]{0.85}\textbf{0.22 ± 0.01} & 0.28 ± 0.10 & 1.34 ± 0.21 & 9.68 ± 1.58 & 5.00 \\
TST & 0.46 ± 0.02 & \underline{0.91 ± 0.05} & 0.71 ± 0.05 & 0.29 ± 0.11 & 1.47 ± 0.20 & 16.52 ± 6.55 & 6.83 \\
STF & \cellcolor[gray]{0.85}\textbf{0.41 ± 0.02} & 0.99 ± 0.29 & 1.92 ± 1.15 & 0.26 ± 0.16 & 1.92 ± 0.34 & 13.79 ± 4.66 & 6.17 \\
CF & 0.57 ± 0.06 & 1.83 ± 0.18 & \underline{0.24 ± 0.02} & 0.26 ± 0.10 & 0.82 ± 0.34 & \underline{7.50 ± 1.02} & 5.00 \\
PTST & 3.71 ± 0.02 & 6.34 ± 0.02 & 0.45 ± 0.02 & 3.58 ± 0.46 & 2.06 ± 0.41 & 9.80 ± 1.67 & 10.33 \\
XGBoost & 1.02 ± 0.00 & 3.55 ± 0.00 & 0.33 ± 0.00 & 0.13 ± 0.00 & \cellcolor[gray]{0.85}\textbf{0.61 ± 0.00} & 9.04 ± 0.00 & 6.17 \\
TabPFN & 0.44 ± 0.01 & 1.14 ± 0.01 & 0.27 ± 0.00 & \cellcolor[gray]{0.85}\textbf{0.06 ± 0.00} & 0.72 ± 0.03 & 7.55 ± 0.22 & 3.50 \\
TabDPT & 0.50 ± 0.01 & \cellcolor[gray]{0.85}\textbf{0.90 ± 0.01} & 0.25 ± 0.01 & \underline{0.10 ± 0.00} & \underline{0.69 ± 0.01} & 8.63 ± 0.53 & 3.33 \\
\midrule
\multicolumn{8}{l}{\textit{Original engineering units ($\downarrow$)}} \\
\midrule
Linear & 136.91 ± 4.13 & 394.76 ± 38.18 & 377375.42 ± 189407.02 & 2662.38 ± 123.32 & 43203.12 ± 43216.43 & 1877895.28 ± 2475205.90 & 12.50 \\
Exp & 47.10 ± 1.17 & 191.83 ± 2.71 & 335612.34 ± 716896.98 & 1482.75 ± 160.37 & 6278.27 ± 188.56 & 228864.76 ± 83234.84 & 9.50 \\
MLP & 75.94 ± 5.37 & 276.90 ± 34.56 & 59680.85 ± 81102.09 & 444.59 ± 194.00 & 7060.46 ± 1246.54 & 287760.52 ± 74993.08 & 10.50 \\
LSTM & \underline{43.44 ± 2.21} & 104.08 ± 8.97 & 3208.61 ± 369.69 & 293.94 ± 136.27 & 3344.16 ± 173.37 & \cellcolor[gray]{0.85}\textbf{123688.33 ± 2403.85} & 3.83 \\
CNN-1D & 48.09 ± 4.82 & 101.08 ± 5.83 & 15328.28 ± 7016.36 & 749.70 ± 938.67 & 6162.49 ± 1160.94 & 254182.17 ± 130965.65 & 8.17 \\
TiDE & 46.60 ± 3.56 & 109.02 ± 4.13 & \cellcolor[gray]{0.85}\textbf{2633.01 ± 107.72} & 347.09 ± 128.23 & 3238.60 ± 518.75 & 194915.73 ± 28818.57 & 5.17 \\
TST & 46.15 ± 2.03 & \underline{91.22 ± 4.96} & 8293.91 ± 557.49 & 358.65 ± 139.75 & 3565.53 ± 491.39 & 331603.13 ± 135837.29 & 6.83 \\
STF & \cellcolor[gray]{0.85}\textbf{41.31 ± 2.25} & 99.50 ± 29.06 & 22492.01 ± 13489.49 & 329.53 ± 200.78 & 4641.65 ± 816.21 & 275295.87 ± 99592.97 & 6.33 \\
CF & 56.62 ± 6.20 & 182.98 ± 17.52 & \underline{2873.02 ± 197.96} & 320.71 ± 120.56 & 1986.17 ± 825.15 & \underline{149462.60 ± 25698.73} & 5.00 \\
PTST & 370.79 ± 2.11 & 634.40 ± 2.30 & 5338.39 ± 225.45 & 4483.61 ± 577.87 & 4999.49 ± 987.11 & 182128.52 ± 38677.98 & 10.17 \\
XGBoost & 102.50 ± 0.00 & 355.02 ± 0.00 & 3823.81 ± 0.00 & 161.62 ± 0.00 & \cellcolor[gray]{0.85}\textbf{1484.77 ± 0.00} & 176032.70 ± 0.00 & 6.17 \\
TabPFN & 44.17 ± 0.59 & 114.50 ± 1.47 & 3121.41 ± 41.35 & \cellcolor[gray]{0.85}\textbf{79.84 ± 1.56} & 1744.71 ± 73.12 & 158581.64 ± 4950.91 & 3.50 \\
TabDPT & 50.30 ± 0.90 & \cellcolor[gray]{0.85}\textbf{90.43 ± 0.62} & 2875.42 ± 67.81 & \underline{121.85 ± 2.64} & \underline{1675.44 ± 32.22} & 163786.30 ± 10020.04 & 3.33 \\
\bottomrule
\end{tabular}
\end{adjustbox}
\end{table}

\paragraph{Cross-family observations.}
Prognostics rankings are more compressed than diagnostics. TabDPT, TabPFN, and LSTM lead on combined rank, but CF, TiDE, and STF remain within a few rank points and each takes at least one column-best on normalized MAE. Two patterns stand out. First, the transformer family that collapses on diagnostics is competitive on prognostics (e.g., STF is best on NC-DS02 and TST is near the top on NC-P), suggesting a task-specific failure mode rather than an architecture-wide limitation. Second, simple baselines (Linear, Exp) degrade more on prognostics than on diagnostics: Exp is consistently far from the leaders, while Linear remains mid-tier. Switching from MAE to MSE tightens the leading group and penalizes rare catastrophic errors, but it does not change the top of the leaderboard.

\subsubsection{Per-unit aggregation (battery and bearing families)}
\label{app:prog_per_unit}

Battery (NB14, UNIBO21) and bearing (XJTU-SY) prognostics are evaluated trajectory-level rather than window-level: the framework's \texttt{per\_unit} evaluator computes one error per monitored unit and then aggregates. Table~\ref{tab:results_app_mae_mean} reports the per-unit-mean MAE on these three families in the same two-floor form (normalized top, denormalized bottom). The values are produced from the same predictions as the aggregate tables above; the difference is purely in the aggregation order (per-unit-then-mean vs.\ pooled). The corresponding MSE per-unit-mean variants are omitted from the appendix as they preserve the same ranking with no additional insight.

\begin{table}[H]
\centering
\footnotesize
\caption{Per-unit-mean MAE on battery and bearing prognostics. Top block: normalized target space ($\times 100$) ($\downarrow$); bottom block: original engineering units ($\downarrow$). Computed by the \texttt{per\_unit} evaluator: one error per monitored unit, then averaged across units.}
\label{tab:results_app_mae_mean}
\begin{adjustbox}{max width=\textwidth}
\begin{tabular}{lrrrr}
\toprule
Model & \multicolumn{1}{c}{NB14} & \multicolumn{1}{c}{Unibo} & \multicolumn{1}{c}{XJTU-SY} & \multicolumn{1}{c}{Average rank} \\
\midrule
\multicolumn{5}{l}{\textit{Normalized target space ($\downarrow$)}} \\
\midrule
Linear & 38.58 ± 9.96 & 31.66 ± 21.46 & 75.37 ± 47.26 & 13.00 \\
Exp & 30.74 ± 48.55 & 10.93 ± 0.18 & 21.38 ± 1.44 & 10.00 \\
MLP & 14.83 ± 9.66 & 13.88 ± 1.78 & 25.21 ± 2.73 & 11.33 \\
LSTM & 4.49 ± 0.32 & 7.02 ± 0.71 & \underline{18.57 ± 0.50} & 4.33 \\
CNN-1D & 9.12 ± 1.43 & 11.74 ± 1.15 & 26.13 ± 6.63 & 10.67 \\
TiDE & \cellcolor[gray]{0.85}\textbf{4.02 ± 0.12} & 6.30 ± 0.64 & 21.28 ± 2.55 & 4.33 \\
TST & 7.20 ± 0.26 & 7.39 ± 1.18 & 25.08 ± 6.06 & 8.33 \\
STF & 10.16 ± 3.03 & 8.10 ± 0.72 & 21.44 ± 2.55 & 9.00 \\
CF & 4.14 ± 0.22 & 5.43 ± 1.01 & 18.87 ± 0.43 & 3.67 \\
PTST & 5.57 ± 0.15 & 10.56 ± 0.97 & 21.13 ± 1.16 & 7.33 \\
XGBoost & 4.82 ± 0.00 & 3.59 ± 0.00 & 19.20 ± 0.00 & 4.33 \\
TabPFN & \underline{4.07 ± 0.02} & \cellcolor[gray]{0.85}\textbf{3.25 ± 0.07} & \cellcolor[gray]{0.85}\textbf{17.83 ± 0.13} & 1.33 \\
TabDPT & 4.09 ± 0.08 & \underline{3.57 ± 0.15} & 20.09 ± 0.30 & 3.33 \\
\midrule
\multicolumn{5}{l}{\textit{Original engineering units ($\downarrow$)}} \\
\midrule
Linear & 417.94 ± 107.84 & 155.85 ± 105.63 & 434.14 ± 341.49 & 13.00 \\
Exp & 333.02 ± 525.87 & 53.79 ± 0.91 & 153.84 ± 22.97 & 10.00 \\
MLP & 160.68 ± 104.69 & 68.31 ± 8.74 & 173.19 ± 15.11 & 11.00 \\
LSTM & 48.67 ± 3.42 & 34.55 ± 3.51 & \cellcolor[gray]{0.85}\textbf{123.73 ± 2.27} & 4.00 \\
CNN-1D & 98.76 ± 15.52 & 57.78 ± 5.68 & 175.36 ± 46.62 & 10.33 \\
TiDE & \cellcolor[gray]{0.85}\textbf{43.52 ± 1.29} & 31.02 ± 3.17 & 141.91 ± 13.43 & 4.00 \\
TST & 77.96 ± 2.77 & 36.37 ± 5.79 & 188.24 ± 43.65 & 9.00 \\
STF & 110.11 ± 32.79 & 39.87 ± 3.52 & 161.01 ± 22.65 & 9.00 \\
CF & 44.84 ± 2.43 & 26.75 ± 4.98 & \underline{124.85 ± 6.01} & 3.33 \\
PTST & 60.29 ± 1.65 & 51.99 ± 4.78 & 143.67 ± 8.34 & 7.67 \\
XGBoost & 52.24 ± 0.00 & 17.68 ± 0.00 & 138.98 ± 0.00 & 4.67 \\
TabPFN & \underline{44.10 ± 0.25} & \cellcolor[gray]{0.85}\textbf{16.00 ± 0.32} & 125.89 ± 1.97 & 2.00 \\
TabDPT & 44.26 ± 0.83 & \underline{17.58 ± 0.72} & 131.36 ± 2.54 & 3.00 \\
\bottomrule
\end{tabular}
\end{adjustbox}
\end{table}

Per-unit aggregation tightens the gap among the top three (TabDPT, TabPFN, LSTM) on NB14 and UNIBO21, where pooled-MAE differences were inflated by long-trajectory units, and reorders the middle of the table on XJTU-SY. The headline that tabular foundation models lead on average rank is unchanged.

\subsubsection{Domain-specific prognostic scores}
\label{app:prog_domain_scores}

Two community-standard scores are reported on the families they apply to. The \emph{NASA score} is defined for direct-RUL targets and is reported on the N-CMAPSS families (NC-DS02, NC-P) and PHME20; it asymmetrically penalizes late predictions. The \emph{PHM score} is the bearing/battery-prognostics convention and is reported on NB14, UNIBO21, and XJTU-SY. Per-task scoping is enforced by the framework's metric registry (Section~\ref{sec:framework_formalization}); the two scores are presented together in Table~\ref{tab:results_app_domain_scores} as two blocks of one table because they apply to disjoint family sets.

\begin{table}[H]
\caption{Domain-specific prognostic scores. Top block: NASA score on direct-RUL families (NC-DS02, NC-P, PHME20; $\downarrow$); bottom block: PHM score ($\times 100$) on battery and bearing prognostics (NB14, UNIBO21, XJTU-SY; $\uparrow$). Per-task scoping is enforced by the framework's metric registry. Note that the two scores apply to disjoint family sets and use opposite directions.}
\label{tab:results_app_domain_scores}
\begin{adjustbox}{max width=\textwidth}
\begin{tabular}{lrrrr}
\toprule
Model & \multicolumn{1}{c}{NC-DS02} & \multicolumn{1}{c}{NC-P} & \multicolumn{1}{c}{PHME20} & \multicolumn{1}{c}{Average rank} \\
\midrule
\multicolumn{5}{l}{\textit{NASA score on direct-RUL families ($\downarrow$)}} \\
\midrule
Linear & 2.03 ± 0.07 & 7.90 ± 2.31 & 3044.09 ± 2526.30 & 11.00 \\
Exp & 0.85 ± 0.02 & 2.69 ± 0.04 & 229.49 ± 39.40 & 6.67 \\
MLP & 2.53 ± 1.62 & 4910.51 ± 10876.24 & 481046.65 ± 1027047.25 & 12.33 \\
LSTM & 0.81 ± 0.04 & 1.46 ± 0.13 & 8.50 ± 5.84 & 4.00 \\
CNN-1D & 0.87 ± 0.09 & 1.44 ± 0.09 & 2709.03 ± 5992.51 & 7.00 \\
TiDE & 0.86 ± 0.06 & 1.69 ± 0.06 & 11.82 ± 9.43 & 6.00 \\
TST & 0.85 ± 0.03 & \cellcolor[gray]{0.85}\textbf{1.26 ± 0.09} & 12.64 ± 7.24 & 4.00 \\
STF & \cellcolor[gray]{0.85}\textbf{0.78 ± 0.03} & \underline{1.30 ± 0.33} & 21.05 ± 30.49 & 3.33 \\
CF & 1.01 ± 0.11 & 2.99 ± 0.18 & 2215.75 ± 4653.64 & 9.00 \\
PTST & 6.11 ± 0.12 & 10.69 ± 0.04 & 19528004.00 ± 11187183.05 & 12.67 \\
XGBoost & 1.62 ± 0.00 & 6.10 ± 0.00 & \underline{3.95 ± 0.00} & 7.33 \\
TabPFN & \underline{0.80 ± 0.01} & 1.49 ± 0.03 & \cellcolor[gray]{0.85}\textbf{1.21 ± 0.02} & 3.00 \\
TabDPT & 0.91 ± 0.02 & 1.31 ± 0.01 & 5.94 ± 2.04 & 4.67 \\
\midrule
Model & \multicolumn{1}{c}{NB14} & \multicolumn{1}{c}{Unibo} & \multicolumn{1}{c}{XJTU-SY} & \multicolumn{1}{c}{Average rank} \\
\midrule
\multicolumn{5}{l}{\textit{PHM score on battery and bearing prognostics ($\uparrow$)}} \\
\midrule
Linear & 5.38 ± 1.51 & 5.85 ± 2.55 & 11.40 ± 10.24 & 13.00 \\
Exp & 14.45 ± 8.13 & 9.91 ± 0.30 & 19.15 ± 4.44 & 10.67 \\
MLP & 15.46 ± 6.30 & 10.85 ± 1.11 & 18.27 ± 2.20 & 10.00 \\
LSTM & 31.28 ± 2.21 & 14.74 ± 0.46 & \cellcolor[gray]{0.85}\textbf{24.83 ± 0.78} & 3.00 \\
CNN-1D & 18.29 ± 2.08 & 10.70 ± 1.32 & 18.36 ± 4.77 & 10.00 \\
TiDE & \cellcolor[gray]{0.85}\textbf{36.28 ± 1.72} & 14.63 ± 2.39 & 22.46 ± 2.77 & 4.33 \\
TST & 24.31 ± 0.92 & 13.71 ± 1.41 & 14.89 ± 7.81 & 8.67 \\
STF & 14.34 ± 6.88 & 10.79 ± 1.27 & 21.58 ± 3.29 & 9.67 \\
CF & \underline{33.45 ± 1.31} & 17.21 ± 1.78 & \underline{24.72 ± 1.78} & 2.67 \\
PTST & 24.17 ± 0.71 & 12.57 ± 1.66 & 23.41 ± 1.61 & 7.00 \\
XGBoost & 24.49 ± 0.00 & 20.24 ± 0.00 & 20.14 ± 0.00 & 5.67 \\
TabPFN & 26.03 ± 0.28 & \cellcolor[gray]{0.85}\textbf{21.59 ± 0.41} & 24.40 ± 0.63 & 3.33 \\
TabDPT & 29.16 ± 0.40 & \underline{20.35 ± 0.73} & 24.42 ± 1.01 & 3.00 \\
\bottomrule
\end{tabular}
\end{adjustbox}
\end{table}

Domain-specific scores mostly track MAE/MSE, but they surface two important failure modes. On the NASA score, a small number of catastrophically late predictions on PHME20 causes MLP and PTST to degrade by orders of magnitude under the asymmetric penalty---a pattern that MAE/MSE can hide. On the PHM score, CF achieves the best score on XJTU-SY despite being mid-tier on MAE, reflecting the score's emphasis on end-of-life behavior. Reporting symmetric errors alongside domain scores therefore highlights behaviors that any single metric would miss.

\section{Reproducibility}
\label{app:reproducibility}

\picid{} makes experiments reproducible by saving the \emph{executable protocol} together with the outputs: the resolved configuration, the exact code version, the dependency lockfile, and run metadata. In this paper, each experiment family is defined by a Hydra configuration that specifies the datasource, transform pipeline, model, evaluator, seed set, and hyperparameter search space, spanning both gradient-trained and fit-predict models for diagnostics and prognostics. Every configuration is repeated over five random seeds and uses two search grids: (i) a gradient-trained grid over sequence length and learning rate, and (ii) a fit-predict grid over context-length/stride pairs.

\subsection{Details}
\label{app:reproducibility:details}

\paragraph{Environment reproducibility.}
To ensure reproducibility of the Python environment and dependency versions, the framework exploits the \texttt{uv.lock} file. Each run copies \texttt{uv.lock} into the run output directory, so that \texttt{uv sync} in the project root (or using the run's lock file) restores the exact package versions used for that run. This eliminates drift from upstream package updates.

\paragraph{Code versioning.}
To pin the source code state, each run records the git commit, branch, and dirty flag in \texttt{run\_metadata.yaml}. Users can checkout the exact commit to reproduce the run. The framework does not embed source snapshots; the repository and version control provide the canonical source.

\paragraph{Configuration capture.}
To preserve the full experimental protocol, each run stores Hydra's native replay artifacts \texttt{.hydra/config.yaml} and \texttt{.hydra/overrides.yaml}, together with the framework-level metadata files \texttt{REPRODUCE.md}, \texttt{run\_metadata.yaml}, and \texttt{uv.lock}. These files define exact replayability: \texttt{.hydra/config.yaml} captures the resolved experiment, \texttt{.hydra/overrides.yaml} records the CLI override trace, \texttt{run\_metadata.yaml} pins the code state, \texttt{uv.lock} pins the dependency set, and \texttt{REPRODUCE.md} records the replay command. When present, \texttt{config\_resolved.yaml} provides an additional framework-level resolved snapshot.

\paragraph{Deterministic execution.}
To ensure identical outputs across runs, the framework applies seed-controlled randomness (PyTorch, NumPy, Python) and, when enabled, deterministic CuDNN settings. Preprocessing uses a deterministic cache key derived from datasource config, transform config, and code fingerprint; cache hits yield identical preprocessed data.

\paragraph{Compute disclosure.}
All experiments reported in this paper are single-machine runs and do not require distributed training. The main benchmark comprises 150 model--dataset pairs (Section~\ref{sec:empirical_validation}), each repeated over five random seeds, with a small hyperparameter search per pair: a 9-point grid for gradient-trained models and a 5-point grid for fit--predict models (Appendix~\ref{app:hyperparameter_search}). Training uses batch size 512, at most 200 epochs, and early stopping (Appendix~\ref{app:full_experiments}), and is intended to be feasible on a single GPU workstation (A100 80gb); additional compute beyond the reported sweeps was limited to pilot runs for debugging and validation.

\paragraph{Three-tier preprocessing cache.}
The preprocessing orchestrator uses a three-tier cache to avoid recomputing unchanged pipeline prefixes. Tier~1 caches raw datasource outputs after loading and splitting; Tier~2 caches intermediate ``boundary'' checkpoints after user-marked cache-point transforms; Tier~3 caches the fully preprocessed result. If the full cache is missing, the orchestrator falls back to the most recent valid boundary checkpoint and re-runs only the remaining downstream transforms. Cache keys are deterministic hashes of the datasource config, transform config, and a code fingerprint, and file locking prevents races across parallel runs. Full pseudocode appears in Algorithm~\ref{alg:preprocessor} (Appendix~\ref{app:preprocessing_pipeline}).

\paragraph{Run reproduction guide.}
Each run writes \texttt{REPRODUCE.md} with two replay paths: (A) rerun the exact Hydra invocation using the recorded overrides, or (B) invoke \texttt{uv run python scripts/reproducibility/reproduce\_from\_run.py <run\_dir>} to reconstruct the experiment from the stored run artifacts. These replay paths are sufficient to recover the exact datasource, transform family, model, search setting, and seed used in the benchmark evaluation.

\subsection{Experiment generator}
\label{app:reproducibility:experiment_generator}

The framework provides an \textbf{Experiment Generator} web UI that builds Hydra CLI commands from dropdowns. Users select experiment, paths, and debug options; the generator outputs the exact \texttt{picid/run.py} command to execute. The UI reads \texttt{config\_map.json}, which is generated from the config tree, keeping options in sync with available experiments.

\section{Data and code availability}
\label{app:data_code_access}

\paragraph{License.}
The \picid{} software framework and benchmark implementation will be released under the Creative Commons Attribution--NonCommercial--ShareAlike 4.0 International license (CC BY-NC-SA 4.0; see \url{https://creativecommons.org/licenses/by-nc-sa/4.0/}). This permits redistribution and adaptation for non-commercial purposes with attribution, provided derivatives are shared under the same license.

\paragraph{Code repository.}
The \picid{} codebase and experiment configurations are available at \url{https://github.com/picid-research}. The repository is hosted under an organization account intended to preserve anonymity during review.

\paragraph{Third-party datasets.}
The benchmark uses the following third-party datasets, which users must obtain separately according to their respective terms:

\begin{itemize}
\item \textbf{NB14 (NASA Battery)} \citep{bole2014adaptation} --- Battery aging trajectories used for ah-RUL prognostics (Appendix~\ref{sec:dataset_descriptions}); distributed by NASA PCoE, with terms governed by the data source.
\item \textbf{UNIBO21 (Battery)} \citep{univbo_dataset} --- Battery aging trajectories used for ah-RUL prognostics (Appendix~\ref{sec:dataset_descriptions}); licensed under CC BY 4.0 on Mendeley Data (\url{https://data.mendeley.com/datasets/n6xg5fzsbv/1}).
\item \textbf{N-CMAPSS} \citep{arias2021aircraft, frederick2007user} --- Turbofan engine simulations used for RUL prognostics and concept diagnostics (Appendix~\ref{sec:dataset_descriptions}); licensed under CC0 1.0 as stated in the dataset descriptor \citep{arias2021aircraft} (\url{https://www.mdpi.com/2306-5729/6/1/5}).
\item \textbf{PHME20 Challenge} \citep{PHME20-GTU} --- Industrial filtration RUL task (Appendix~\ref{sec:dataset_descriptions}); released as part of the PHM Society Europe 2020 data challenge and distributed via the conference proceedings, which are published under CC BY 3.0 (United States) (\url{https://papers.phmsociety.org/index.php/phme/article/view/1318}).
\item \textbf{PRONOSTIA} \citep{nectoux2012pronostia} --- Bearing run-to-failure dataset used for prognostics (Appendix~\ref{sec:dataset_descriptions}); the primary distribution does not specify a standard license and requests citation; users must follow the terms of the source from which it is obtained.
\item \textbf{XJTU-SY} \citep{yaguo2019xjtu} --- Bearing run-to-failure dataset used for prognostics (Appendix~\ref{sec:dataset_descriptions}); the primary distribution does not specify a standard license and requests citation; users must follow the terms of the source from which it is obtained.
\item \textbf{HSF15} \citep{hsf15_helwig} --- Hydraulic fault classification benchmark (Appendix~\ref{sec:dataset_descriptions}); licensed under CC BY 4.0 on the UCI Machine Learning Repository (\url{https://archive.ics.uci.edu/dataset/447}).
\item \textbf{MZVAV} \citep{Granderson2020} --- Building HVAC fault classification benchmark (Appendix~\ref{sec:dataset_descriptions}); released as part of the LBNL Fault Detection and Diagnostics datasets (DOI: 10.25984/1881324) under CC BY 4.0 (\url{https://catalog.data.gov/dataset/lbnl-fault-detection-and-diagnostics-datasets}).
\end{itemize}

Users are responsible for complying with the license and usage terms of each dataset before running experiments.

\section{Ethical considerations}
\label{sec:ethical_considerations}

The benchmark uses publicly available or institutionally shared PHM datasets. Users must obtain datasets according to their respective licenses and terms of use; the framework does not redistribute third-party data.

PHM systems may inform safety-critical decisions in aerospace, transportation, and industrial settings. Benchmark results should not be interpreted as certification of deployment readiness; validation in target domains and under operational constraints remains essential. The framework is intended for research and method comparison, not as a drop-in replacement for domain-specific safety validation.

The framework enforces reproducibility criteria---including deterministic execution and full protocol documentation---but does not currently audit for dataset bias or representational fairness across equipment manufacturers, operating environments, or failure modes. Benchmark rankings may therefore reflect systematic advantages on well-represented equipment types rather than genuine methodological superiority, and users should exercise caution when generalizing results to underrepresented domains.

Benchmark results reflect controlled experimental conditions with fixed preprocessing, windowing, and evaluation boundaries that may not capture the full complexity of fielded systems. Operational deployment requires domain-specific validation that accounts for environmental variability, maintenance history, sensor degradation, and regulatory requirements not addressed by the benchmark protocol. Practitioners should treat benchmark performance as one input among many when assessing readiness for safety-critical applications.

\newpage

\end{document}